\documentclass{article}

\usepackage[eandd, final]{neurips_2026}

\usepackage[utf8]{inputenc} 
\usepackage[T1]{fontenc}    
\usepackage{hyperref}       
\usepackage{url}            
\usepackage{booktabs}       
\usepackage{amsfonts}       
\usepackage{nicefrac}       
\usepackage{microtype}      
\usepackage{xcolor}         

\usepackage[ruled,linesnumbered]{algorithm2e}
\usepackage{amsmath}
\usepackage{amssymb}
\usepackage{booktabs}     
\usepackage{etoolbox}
\usepackage{graphicx}
\usepackage{siunitx}
\usepackage{subcaption}
\usepackage{wrapfig}
\usepackage{tabularx}
\usepackage{multirow}
\usepackage{placeins}

\usepackage{cleveref}

\crefname{appsec}{Appendix}{Appendices}
\Crefname{appsec}{Appendix}{Appendices}

\setcitestyle{numbers,square,comma}
\setcitestyle{comma}
\setcitestyle{square}

\pretocmd{\appendix}{%
  \crefalias{section}{appsec}%
  \crefalias{subsection}{appsec}%
  \crefalias{subsubsection}{appsec}%
}{}{}

\usepackage[table]{xcolor}
\makeatletter

\newcommand{\Rmnum}[1]{\expandafter\@slowromancap\romannumeral #1@}
\makeatother

\usepackage{pifont}
\newcommand{\cmark}{{\color{green!70!black}\checkmark}} 
\newcommand{\xmark}{{\color{red!80!black}\ding{55}}}     
\newcommand{\dimcolor}[1]{%
  \ifx#1 2D {\color{blue}2D}%
  \else\ifx#1 3D {\color{violet}3D}%
  \else {\color{cyan}#1}%
  \fi\fi
}

\title{ALETHEIA: A Multi-Frequency Eddy Current Pulsed Thermography Dataset for Neural Operator Learning in Nondestructive Testing}

\author{%
  Changbin Sun \thanks{Equal contribution.}\\
  College of Computer Science \\
  Sichuan University
  \And
  Jie Xiao \footnotemark[1]\\
  School of Mechanical Engineering \\
  Sichuan University
  \AND
  Xiaotian Chen \thanks{Corresponding authors:
  Xiaotian Chen and Yuankai Wu.} \\
  School of Mechanical Engineering \\
  Sichuan University
  \And
  Yuankai Wu \footnotemark[2]\\
  College of Computer Science \\
  Sichuan University
}

\begin{document}

\maketitle

\begin{abstract}
Learning neural solvers for spatiotemporal partial differential equations (PDEs) under real-world constraints remains a key challenge in scientific machine learning, especially for inverse tasks with sparse and noisy boundary observations. We present the \textbf{Aletheia} dataset, the first 3D benchmark for learning data-driven solvers in the context of \textbf{nondestructive testing (NDT)}. The dataset simulates eddy-current-induced heating in conductive solids and models the resulting transient heat propagation governed by the heat equation. Aletheia contains over 4,700 high-resolution samples across 10 excitation frequencies (1-100\,kHz), each providing volumetric heat source and temperature fields over time. It supports both forward prediction of temperature evolution and inverse reconstruction of internal heat sources or defects from surface infrared measurements. Real infrared thermography data from cracked rail specimens are included for calibration and comparison. We define four canonical tasks on both regular and irregular grids and benchmark them using various neural operators. Aletheia establishes a unified platform for evaluating neural PDE solvers under realistic NDT conditions, enabling progress in reliable, data-driven inverse modeling. The dataset and code are publicly available at \url{https://github.com/sherlock-0327/ALETHEIA}.
\end{abstract}

\section{Introduction}
\label{section: Introduction}

Neural operator methods, such as the Fourier Neural Operator (FNO)~\citep{FNO, Geo-FNO, FFNO, xiao2024amortized} and Transformer-based solvers~\citep{li2023scalable, Transolver, lee2024inducing}, have emerged as a transformative approach for learning solution operators of partial differential equations (PDEs) directly from data. Unlike traditional methods, these architectures bypass mesh-dependent discretizations, enabling robust generalization across parameterized PDE families. However, they are typically evaluated on academic datasets (e.g., Darcy flow, Navier–Stokes) with fully observed fields and simplified geometries, which fail to capture the complexities of real-world inverse problems. In applications like nondestructive testing (NDT)~\citep{gupta2022advances, xiong2023detection, yuan2021investigation, gong2022nondestructive, tuschl2021inductive}, challenges such as sparse or noisy boundary observations, unknown source terms, and heterogeneous domains demand more robust benchmarks~\citep{molinaro2023neural, azizzadenesheli2024neural}.

\begin{figure}[t]
  \centering
  \includegraphics[width=1\textwidth]{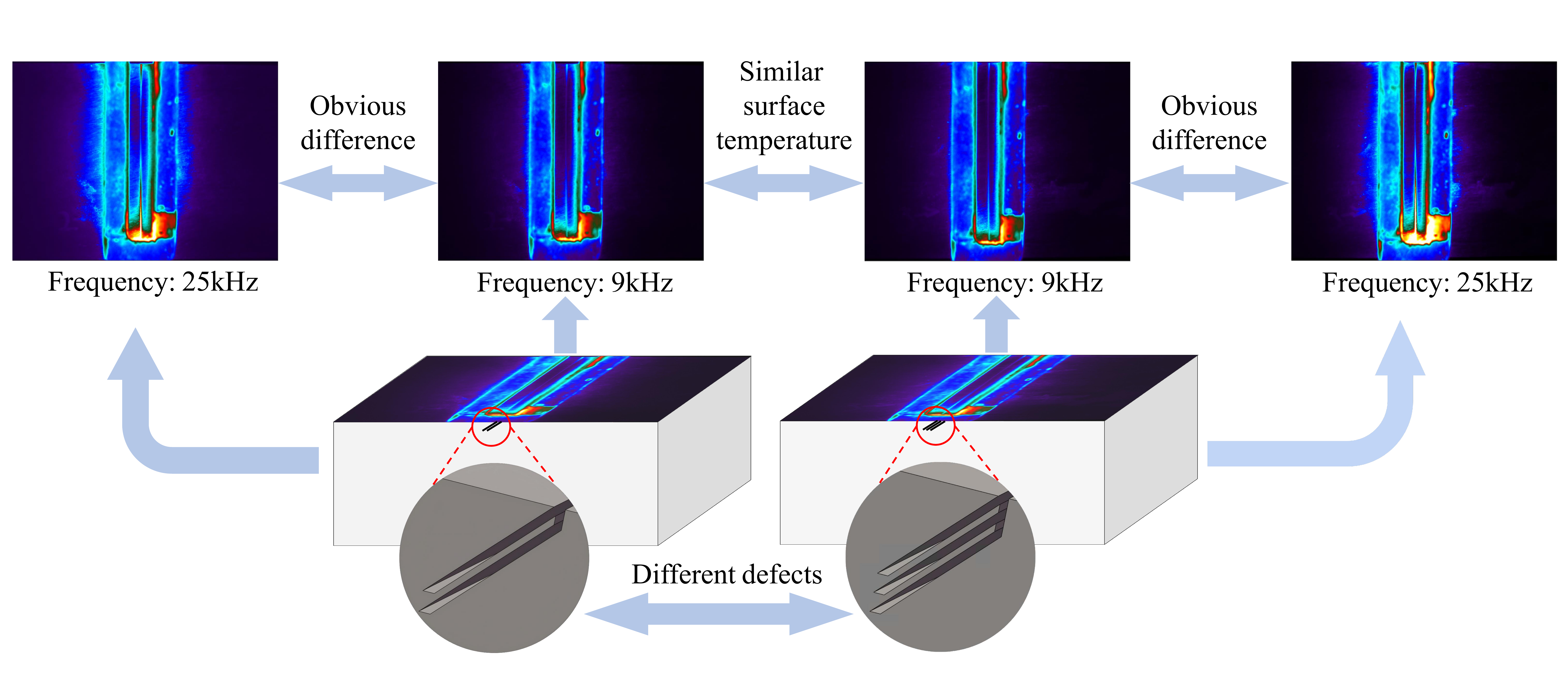}
  \caption{\textbf{Multi-frequency stimulation in the Aletheia dataset.} Different defects may produce similar surface temperatures under the same frequency (9 kHz), but show clear differences at another frequency (25 kHz). Using multiple frequencies helps distinguish hidden defects by capturing varied thermal responses at different depths.}
  \label{fig:differentfreqs}
\end{figure}

In NDT, reconstructing hidden defects~\citep{lin2023study, zhao2022end, tao2022unsupervised, wu2021reinforcement} from surface temperature measurements, as in inverse heat conduction problems~\citep{silva2023review}, is inherently ill-posed: distinct subsurface defects or excitation conditions can produce nearly identical surface temperature patterns~\citep{woodbury2023inverse}, as illustrated in \Cref{fig:differentfreqs}. To address this, we employ multi-frequency pulsed induction heating, where different excitation frequencies probe the material at varying depths\textemdash lower frequencies penetrate deeper to capture internal defect responses, while higher frequencies reveal surface-level thermal behavior~\citep{liang2024thermal}. As shown in \Cref{fig:differentfreqs}, while some frequencies (e.g., 9 kHz) may yield similar surface temperatures for different defects, others (e.g., 25 kHz) reveal distinct patterns. This frequency-dependent response diversity breaks single-frequency ambiguity, enhancing defect discriminability.

\begin{figure}[t]
  \centering
  \includegraphics[width=1\textwidth]{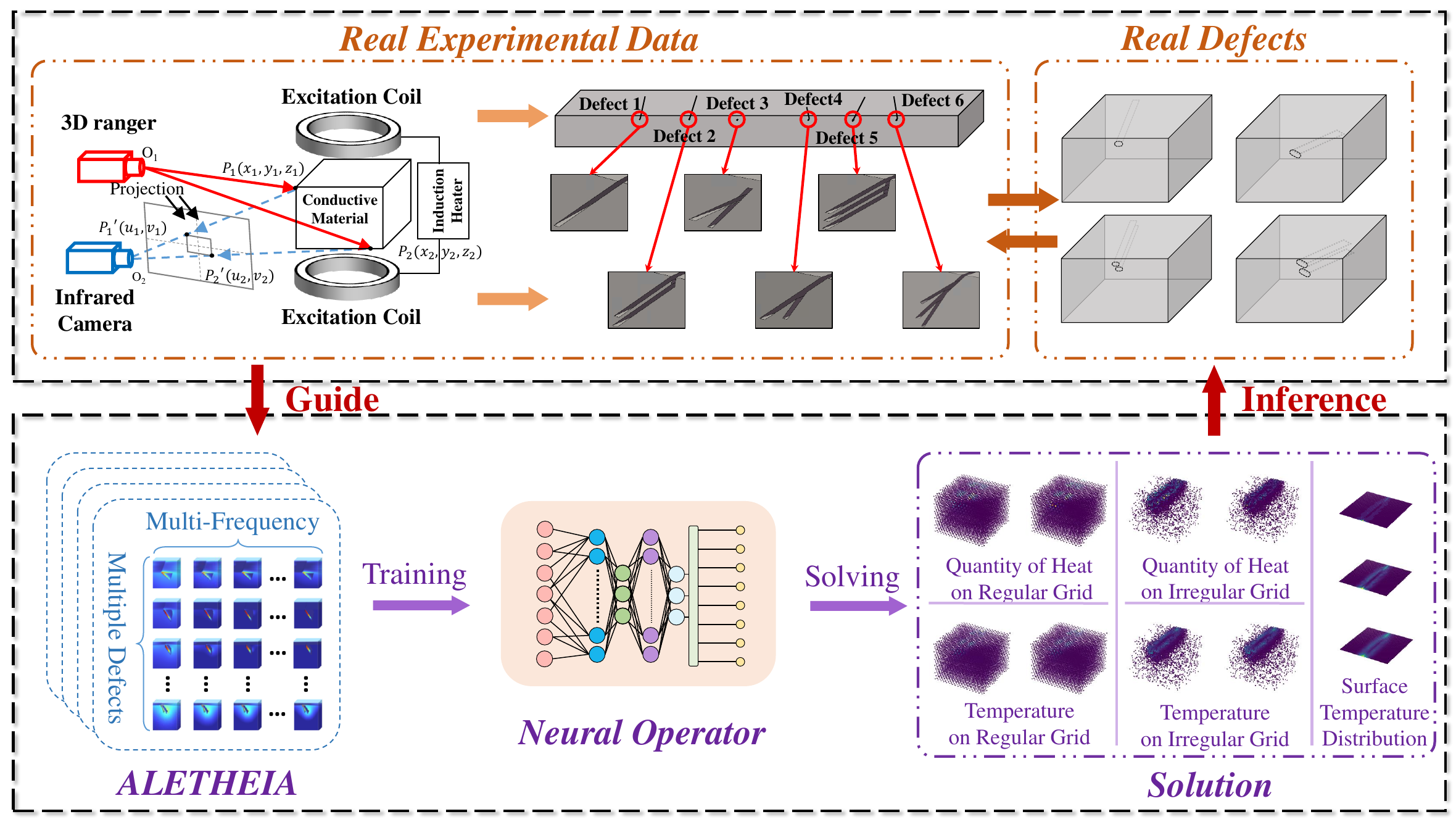}
  \caption{\textbf{Overview of the Aletheia dataset.} \emph{Top}: real infrared thermography sequences from rail specimens with fatigue cracks. \emph{Bottom}: synthetic data pipeline generating multi‑frequency (1\textemdash 100\,kHz) volumetric heat sources and corresponding transient temperature fields for 4,700+ defect cases, on both regular and irregular grids. This simulation benchmark calibrated against real measurements supports forward prediction, inverse reconstruction, and cross‑frequency evaluation in NDT.}
  \label{fig:aletheiaoverview}
\end{figure}

Existing PDE-learning benchmarks lack realistic thermal-boundary coupling and 3D scenarios tailored to heat-source inversion or volumetric temperature prediction in NDT. To bridge this gap, we introduce the \textbf{Aletheia} dataset (\Cref{fig:aletheiaoverview}), a comprehensive 3D benchmark that integrates high-fidelity simulations with real infrared measurements from rail specimens with internal fatigue cracks. The dataset includes over 4,700 defect cases across 10 excitation frequencies (1–100 kHz), providing volumetric heat-source maps (eddy-current-induced Joule heating) and time-resolved temperature fields on both regular and irregular grids. Calibrated thermography captures transient surface temperature sequences, while multi-frequency conditions supply depth-sensitive signals to mitigate the ill-posedness of surface-only observations.

Using Aletheia, we address three evaluation objectives in eddy current thermography and PDE benchmarking: (1) \textbf{forward thermal prediction} of full 3D temperature evolution from known sources; (2) \textbf{inverse source reconstruction} of latent heat distributions or defect geometries from sparse surface data; and (3) \textbf{out-of-distribution (OOD) generalization} to unseen frequencies and defect shapes. Overall, our contributions are summarized as follows:
\begin{itemize}

\item We present the first publicly available simulation dataset \textbf{Aletheia} in the context of electromagnetic-thermal coupling, enabling data-driven modeling of eddy current thermography.

\item Aletheia provides a multi-frequency dimension: data covering a range of excitation frequencies from low to high such as 1\textemdash 100 kHz, capturing the effect of frequency on the depth and effectiveness of the heat.

\item Aletheia contains three-dimensional, temporally-evolving data, such as the evolution of the entire temperature field after pulse heating, not just steady-state or two-dimensional observations

\item Aletheia combines high-fidelity simulations and experimental measurements. Simulation data provide comprehensive information on field distributions and “true value” defects, while experimental data introduce real noise and variability and verify the reliability of the simulation.

\item Built around the real engineering application of rail crack detection, Aletheia covers a wide range of defect types and sizes with direct engineering relevance.

\end{itemize}

\section{Background}
\label{section: Background}

Neural operators have revolutionized data-driven PDE modeling by learning mapping between function spaces ~\citep{DeepONet, FNO, NeuralOperator2}, but existing benchmarks remain narrowly focused on fully observed, forward-only problems and lack realistic inverse or measurement-sparse scenarios. Eddy-current thermography(ECT) offers a rich real-world setting in NDT, yet no public 3D dataset exists for learning heat-field inversion under varying excitation. We therefore position our dataset at the intersection of these gaps, enabling the study of both forward prediction and inverse source reconstruction for transient heat conduction from surface measurements.

\subsection{Challenges and applications of neural PDE solvers}

Neural operator methods (e.g., DeepONet ~\citep{DeepONet}, FNO ~\citep{FNO}) learn PDE solution operators directly from data, achieving mesh- and resolution-invariance and outperforming classical surrogates on benchmarks like Darcy flow and NavierStokes ~\citep{PDEBench}. Despite their promise, these models struggle to capture high-frequency components due to spectral truncation in Fourier layers, exhibit poor extrapolation to OOD parameters, and lack robustness under noisy or incomplete boundary observations common in inverse problems. Moreover, most evaluations assume full-field availability, whereas many applications demand reconstructing latent sources or fields from sparse measurements.

\subsection{Thermal holography for NDT}

ECT~\citep{gao2024quantification,zou2022nondestructive,zu2023stress} combines electromagnetic induction heating with infrared imaging to detect subsurface defects by capturing surface temperature anomalies. A common implementation, pulsed ECT (ECPT)~\citep{zhang2024surface,chen2021tomographic}, uses short bursts of alternating current to induce volumetric Joule heating in conductive materials, enabling high-sensitivity, non-contact inspection of internal structures~\citep{yin2021quantification,ma2024characterisation,zhang2021detection}. This process gives rise to the concept of thermal holography~\citep{utadiya2023digital}, where internal heat sources perturbed by defects such as cracks are inferred from transient surface temperature fields. {Mathematically, this is governed by the heat equation with a volumetric heat source:
\begin{equation}
\frac{\partial u}{\partial t} = \alpha \nabla^2 u + q,
\end{equation}
where \( u(\mathbf{x}, t) \) is the temperature field at position \( \mathbf{x} = (x, y, z) \) and time \( t \), \( \alpha \) is the thermal diffusivity, and \( q(\mathbf{x}, t) \) represents the internal heat sources induced by eddy currents. The inverse problem in thermal holography seeks to reconstruct the spatial distribution of the heat source \( q(x, y, z) \) from sparse and noisy surface temperature measurements \( u(x, y, z_s, t) \), where \( z_s \) denotes the surface of the domain \( \partial \Omega \), and the observation set is defined over coordinates \( (x, y, z_s, t) \).} In many practical settings, the primary goal is to reconstruct the volumetric distribution of these internal heat~\citep{hong2023reconstructed,wang2021study} sources rather than directly visualize the defects themselves. This inverse heat conduction task is inherently ill-posed and only sparse, noisy boundary data are typically available. Despite its critical role in nondestructive testing across domains such as rail, aerospace~\citep{gebrehiwet2023review,gholizadeh2022impact,jacob2022data}, and pipeline inspection~\citep{cheng2021review,wang2024basic}, there exists no standardized benchmark dataset for learning data-driven solvers that can tackle this class of inverse thermal problems under realistic conditions.

\subsection{Limitation of existing PDE benchmarks}
As summarized in \Cref{tab:benchmark_comparison}, most existing PDE benchmark datasets exhibit limited diversity in terms of task settings and data characteristics. Specifically, widely used datasets such as \textit{Darcy Flow}, \textit{NavierStokes}, \textit{Burgers' Equation}, and \textit{Shallow Water} primarily focus on forward and inverse problems in regular 2D geometries with full observations and lack support for more complex learning scenarios. Among commonly used benchmarks ~\citep{PDEBench, herde2024poseidon, DynaBench} we surveyed none of these datasets simultaneously offer partial observability and multi-task evaluation capabilities. Although the \textit{FWI-F/L/FL} ~\citep{FWI-F} suite introduces inverse modeling and partial observations, it remains confined to 2D domains with relatively simple geometries and lacks support for multitask learning. Similarly, \textit{BubbleML} ~\citep{bubbleml} is an excellent multiphase, multiphysics dataset, yet it still lacks testing capabilities for inverse problems and OOD problems. Furthermore, mainstream and widely used benchmarks predominantly focus on low spatial dimensions(1D or 2D), which falls short of the complexity found in real-world scientific and engineering problems that often involve high-dimensional, irregular domains with heterogeneous observability and spatiotemporal dynamics. 

\begin{table}[ht]
    \centering
    \small
    \setlength{\tabcolsep}{4.5pt}
    \renewcommand{\arraystretch}{1.0}
    \caption{Comparison of PDE benchmark datasets. Each checkmark (\cmark) indicates the presence of a specific feature in the dataset. \emph{Spatial Dim.} denotes the predominant dimensionality used in common benchmarks, not a limitation of the PDE itself.}
    \begin{tabular}{lccccccc}
    \toprule
    \textbf{Benchmark Dataset} & \textbf{Spatial Dim.} & \textbf{Inverse} & \textbf{Partial Obs.} & \textbf{Irregular Geo.} & \textbf{Multi-task} & \textbf{OOD} \\
    \midrule
    Advection          & \dimcolor{1D}           & \cmark & \xmark & \xmark & \xmark & \cmark \\
    Darcy Flow         & \dimcolor{2D}           & \cmark & \xmark & \xmark & \xmark & \cmark \\
    NavierStokes       & \dimcolor{1D/2D/3D}     & \cmark & \xmark & \xmark & \xmark & \cmark \\
    Burgers' Equation  & \dimcolor{1D/2D}        & \cmark & \xmark & \xmark & \xmark & \cmark \\
    Airfoil Flow       & \dimcolor{2D}           & \cmark & \xmark & \cmark & \xmark & \xmark \\
    Diffusion Reaction & \dimcolor{1D/2D}        & \cmark & \xmark & \xmark & \xmark & \cmark \\
    Shallow Water      & \dimcolor{2D}           & \xmark & \xmark & \xmark & \xmark & \xmark \\
    Plasticity / Elasticity   & \dimcolor{2D/3D} & \xmark & \xmark & \cmark & \xmark & \xmark \\
    BubbleML           & \dimcolor{2D/3D}        & \xmark & \xmark & \cmark & \cmark & \xmark \\
    FWI-F / L / FL     & \dimcolor{2D}           & \cmark & \cmark & \xmark & \xmark & \cmark \\
    \rowcolor{gray!10}
    \textbf{Aletheia (Ours)} & \dimcolor{3D} & \cmark & \cmark & \cmark & \cmark & \cmark \\
    \bottomrule
    \end{tabular}
    \label{tab:benchmark_comparison}
\end{table}

These limitations hinder comprehensive evaluation of model generalization, robustness, and versatility. In contrast, our proposed dataset, \textbf{Aletheia}, is designed to fill this gap by incorporating inverse tasks, partial and sparse observations, irregular 3D geometries with temporal dynamics, multi-task learning, and OOD generalization, providing a more realistic and challenging benchmark for PDE learning.

\section{Dataset construction and details}
\label{section: Dataset construction and details}

\subsection {Overview of the dataset}
Aletheia is derived from two complementary sources. The first source consists of high-fidelity multi-physics simulations, which generate synthetic data by numerically solving a fully coupled electric–magnetic–thermal transient process. The second source comprises experimental measurements, where actual data are collected via frequency-swept pulsed eddy current thermography experiments performed on steel rail samples containing artificial defects. Due to the scarcity and complexity of real defect specimens, experimental data primarily serve to calibrate and validate the simulation model. Conversely, simulated data are extensively utilized for model training as they offer complete three-dimensional temperature field evolutions, internal heat source distributions, and precise defect geometries\textemdash details which real ECPT experiments cannot provide, being limited to surface temperature time series measurements. Recognizing that the real experimental environment closely mimics authentic engineering inspection scenarios, we meticulously calibrated our simulation models. Specifically, we fine-tuned material properties and excitation parameters so that simulated surface temperature curves align closely with experimental measurements, achieving an accuracy within ±1°C. This calibration supports agreement with the measured surface-temperature curves under the tested conditions.
\subsection {Simulation data acquisition}

We used COMSOL Multiphysics\footnote{\url{https://www.comsol.com}} to develop a three-dimensional finite element simulation pipeline modeling the coupled process of electromagnetic induction heating and thermal diffusion. For each simulated sample, we first constructed a 3D rail model containing a defect and set material properties such as electrical conductivity $\sigma$, thermal conductivity $k$, and relative permeability $\mu_r$. We then specified the defect geometry and input parameters, including the excitation frequency $f$ and coil current. After the model was built, an adaptive mesh was generated with finer elements in the defect region due to the defect’s small size to ensure computational accuracy. The simulation employed a 600\,ms pulsed heating process. For more detailed simulation parameter settings, please refer to \Cref{appendix: Simulation experiment details}. When an alternating current pulse at the specified frequency was applied to an excitation coil fixed just above the rail, it induced eddy currents in the metal specimens according to Faraday's law. Meanwhile, according to Ampere's Law, the magnetic field generated by induced eddy currents in turn affects the original field. The Maxwell equations can describe this electromagnetic process:
\begin{equation}
    \nabla \cdot \mathbf{D} = \rho, \quad
\nabla \cdot \mathbf{B} = 0, \quad
\nabla \times \mathbf{E} = -\frac{\partial \mathbf{B}}{\partial t}, \quad
\nabla \times \mathbf{H} = \mathbf{J} + \frac{\partial \mathbf{D}}{\partial t}.
\end{equation}
where $\mathbf{D}$ denotes the electric displacement vector, $\rho$ is the free charge density; $\mathbf{B} = \mu \mathbf{H}$ represents the magnetic induction intensity; $\mathbf{H}$ is the magnetic field intensity; $\mathbf{E}$ is the electric field intensity; $\mathbf{J} = \sigma \mathbf{E}$ indicates the eddy current density. Eddy current inside the conductor generated Joule heat due to resistance, acting as an internal heat source:
\(
q(x, y, z) = \frac{|\mathbf{J}|^2}{\sigma}.
\)
After converting this electromagnetic energy into thermal energy, it diffused through thermal conduction in the material.  

The depth of eddy current penetration, known as the skin depth, is frequency-dependent and given by:
\(
\delta = \sqrt{\frac{1}{\pi f \mu \sigma}},
\)
where $\mu = \mu_r \mu_0$ is the magnetic permeability ($\mu_0$ is the permeability of free space). Higher frequencies reduce $\delta$, concentrating eddy currents and heat near the surface, while lower frequencies allow deeper penetration, enabling differentiation of subsurface defect responses. To overcome the inherent ill-posedness of mapping surface temperature back to defect characteristics, we employ multiple excitation frequencies in the pulsed heating process—while some frequencies may yield indistinguishable temperature profiles for certain defect types, others reveal distinct thermal distributions, enabling us to accurately distinguish different types of defects. The electromagnetic and thermal phenomena within our simulations were fully coupled, mutually influencing one another throughout the transient process. Upon completion of each simulation, we systematically recorded the temporal evolution of the temperature field $u(x, y, z, t)$ across the specimen's surface and internal volume, as well as the spatial distribution of the internal heat sources $q(x, y, z)$ generated by induced eddy currents.

\begin{figure*}[ht]
    \centering
    \begin{subfigure}{0.495\textwidth}
        \centering
         \includegraphics[width=\textwidth]{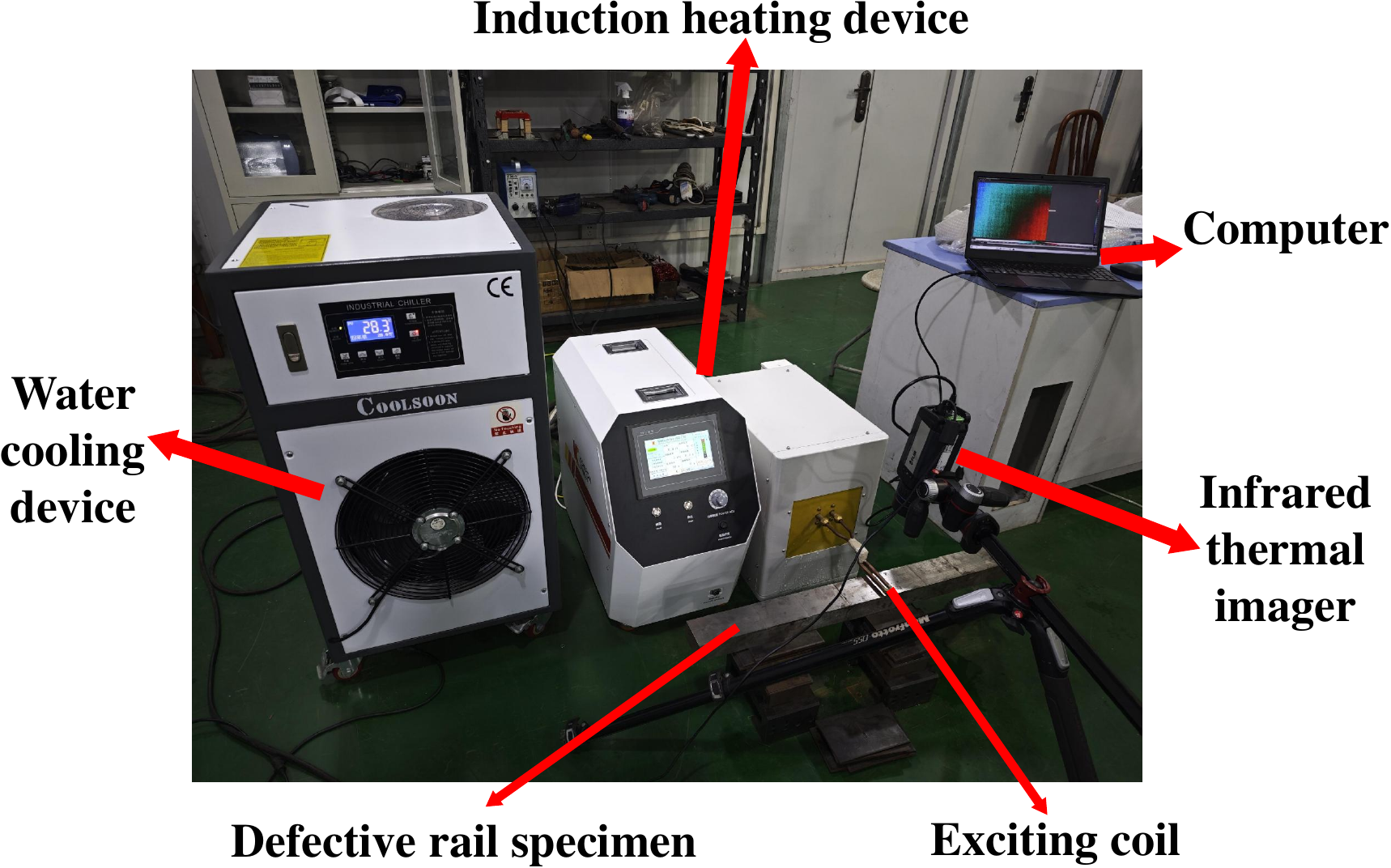} 
        \caption{Experimental setup for pulse eddy current thermal imaging.}
        \label{fig:device}
    \end{subfigure}
    \hfill 
    \begin{subfigure}{0.495\textwidth}
        \centering
         \includegraphics[width=\textwidth]{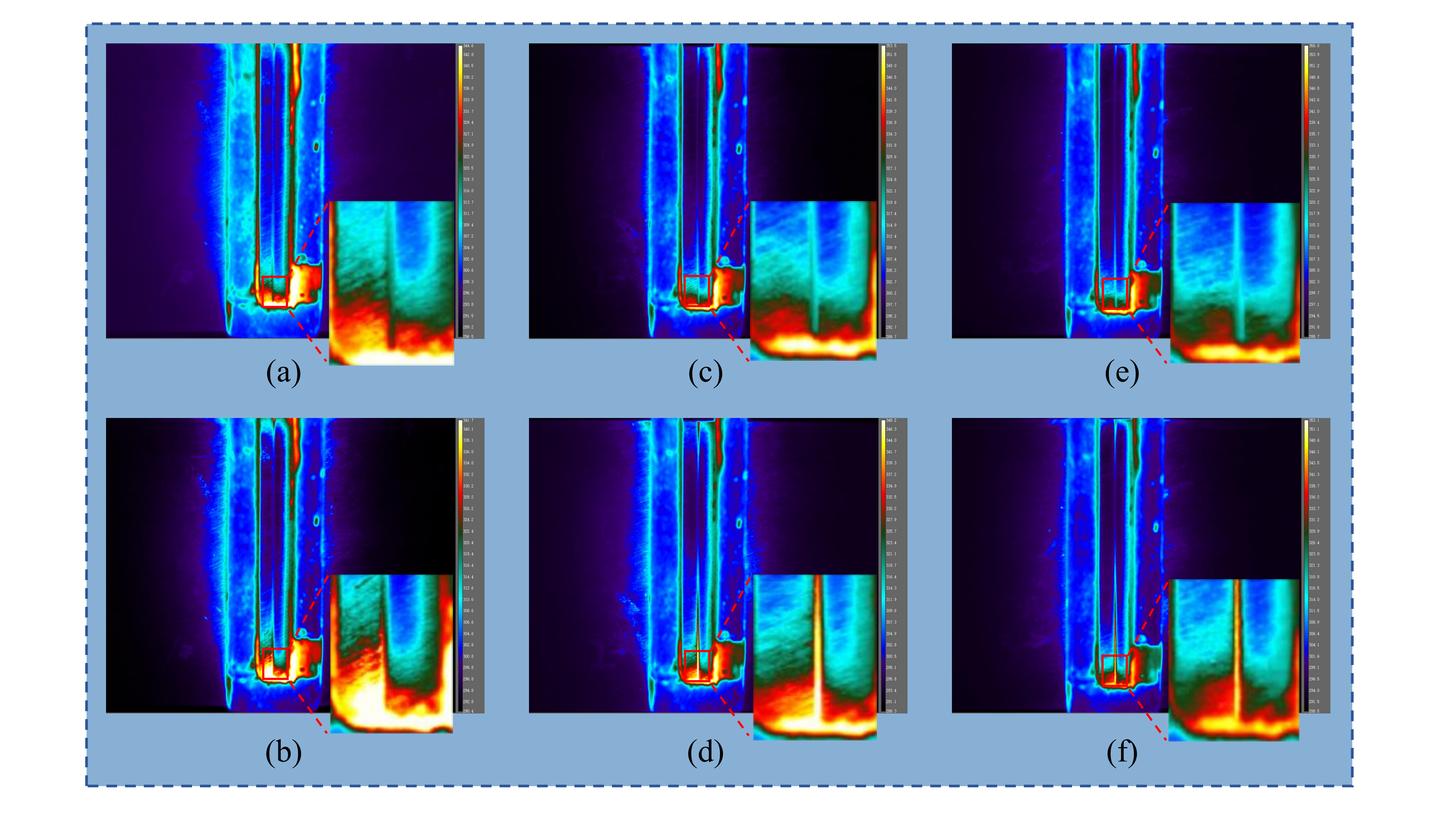} 
        \caption{Thermal imaging results for different defect types and excitation frequencies.}
        \label{fig:exp_data}
    \end{subfigure}
    \caption{Overview of the pulse eddy current thermal imaging experimental platform and selected thermal imaging results.}
    \label{fig:combined}
\end{figure*}

\subsection {Experimental data acquisition}
We additionally performed pulsed eddy current thermography experiments to acquire authentic measurement data. The experimental apparatus comprised a custom-built XZ-series DSP-controlled inverter power supply paired with a specially designed excitation coil featuring a central slot, as depicted in Figure~\ref{fig:combined}. During each experimental run, a sinusoidal pulse current with a duration of 600 ms was applied at selected frequencies within the range of 1\textemdash 100 kHz. The excitation coil was maintained at a constant lift-off distance of 5 mm above the steel rail sample. A high-resolution FLIR SC6550A infrared thermal imager was employed to capture the evolution of surface temperatures, providing imagery at a spatial resolution of 640×480 pixels, a frame rate of 50 frames per second, and an intensity depth of 14 bits. This allowed detailed recording of the temporal temperature distribution throughout both the heating and cooling phases. 

The collected experimental data served to introduce realistic ambient noise into our dataset and was used to validate the accuracy and reliability of our simulation model.

\begin{table}[htbp]
  \caption{Number of different types of defects. There are a total of 6 types of defects, please refer to \Cref{appendix: Defect Details} for more detailed information on each type of defect.}
  \label{table: Number of different types of defects}
  \centering
  \begin{tabular}{ccccccc}
    \toprule
        \multirow{2}{*}{Single layer} & Type I & Type II & Type III & Type I & Type II &  \multirow{2}{*}{Total} \\ 
  & double-layer & double-layer & double-layer & multi-layer & multi-layer \\ 
    \midrule
    88 & 1050 & 69 & 2285 & 90 & 1200 & 4782 \\ 
    \bottomrule
    \end{tabular}
    \label{table:defect}
\end{table}

\subsection {Dataset structure}
Leveraging our calibrated simulation framework, we systematically generated a comprehensive set of defect samples using automated scripts and parallelized computations. After extensive simulations and rigorous calibration processes, the final dataset comprises a total of 4,782 unique defect instances, categorized into 2,407 open-crack and 2,375 closed-crack samples. Detailed statistics for each defect type are summarized in \Cref{table:defect}. Each defect instance includes simulation data captured across ten discrete excitation frequencies\textemdash specifically, 1 kHz, 4 kHz, 9 kHz, 16 kHz, 25 kHz, 36 kHz, 49 kHz, 64 kHz, 81 kHz, and 100 kHz\textemdash to reflect varied depth sensitivities and thermal responses. For versatility across different modeling approaches, the thermal and temperature fields within each defect instance were sampled using two distinct strategies: regular grid sampling and irregular point sampling. Regular grid sampling aligns directly with the structured grid employed by experimental infrared imagery, providing consistent surface temperature time-series data for model input. Conversely, the irregular sampling strategy mimics the actual positioning of defects relative to the infrared camera used in experimental setups. These irregularly sampled points follow a multivariate Gaussian distribution concentrated around the central surface line of the specimen, accurately reflecting defect locations encountered in practical inspections. Points near defect regions are sampled densely to ensure higher resolution and precise reconstruction of defect contours, whereas areas less relevant to defect detection are sampled more sparsely, thus reducing computational overhead. This carefully designed sampling method optimally balances computational efficiency with reconstruction accuracy.

\section{Benchmark}
\label{section: Benchmark}

To systematically assess the modeling capabilities of neural operators in 3D spatiotemporal heat transfer problems, we construct a comprehensive benchmark suite based on the Aletheia dataset. This benchmark evaluates four operator-learning tasks, encompassing both \textit{Same-Frequency} and \textit{Out-of-Distribution} scenarios. Each task is evaluated on two types of spatial grids: regular and irregular to rigorously test the generalization and robustness of neural operator models across diverse spatial discretizations. We benchmark several representative neural operator architectures, including FNO and Transolver, under consistent training and evaluation protocols. This unified framework offers a strong baseline for future research and provides critical insight into the strengths and limitations of current neural PDE solvers in realistic NDT contexts. 

\begin{figure}[htbp]
    \centering
    \includegraphics[width=0.9\textwidth]{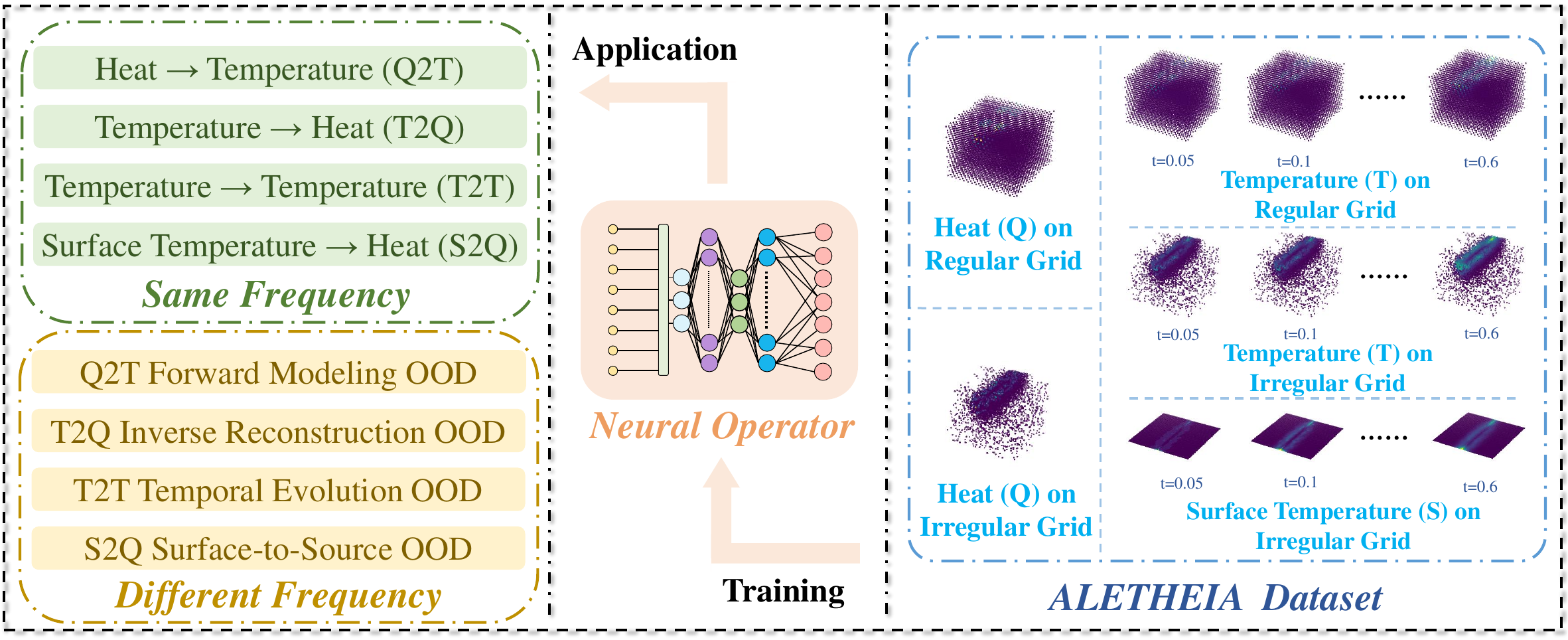}
    \caption{Our benchmark covers source-to-temperature prediction (Q2T), temperature-to-source reconstruction (T2Q), temporal forecasting (T2T), and surface-to-source reconstruction (S2Q), with evaluation under frequency and geometric distribution shifts.} 
    \label{fig: task-overview}
\end{figure}

\subsection{Task setup}
We organize the tasks into two categories according to the alignment between training and testing distributions:

\paragraph{1. Full-frequency tasks (in-distribution / closed-set)} 
These tasks evaluate the accuracy of the model when the training and testing data are sampled from the same excitation frequencies and boundary conditions:
\begin{itemize}
    \item \textbf{Forward Modeling (Q2T):} Learn the mapping \( q(\mathbf{x}, t) \mapsto u(\mathbf{x}, t) \), i.e., predict the transient temperature field given the internal heat source.
    \item \textbf{Inverse Source Reconstruction (T2Q):} Learn the inverse mapping \( u(\mathbf{x}, t) \mapsto q(\mathbf{x}, t) \), recovering the latent volumetric heat source from the transient temperature field \( u(x, y, z, t) \).
    \item \textbf{Temporal Evolution Prediction (T2T):} Learn \( \{u(\mathbf{x}, t_i)\}_{i=0}^{10} \mapsto \{u(\mathbf{x}, t_j)\}_{j=11}^{12} \), forecasting future temperature fields based on historical states. 
    \item \textbf{Surface-to-Source Reconstruction (S2Q):} Learn \( s(\mathbf{x}_{\partial}, t) \mapsto q(\mathbf{x}, t) \), Reconstructing the implicit volumetric heat source from the top surface temperature trace  \( s(x, y, z, t) \), when the values of \(z\) are 0.
\end{itemize}

\paragraph{2. Out-of-distribution (OOD) generalization tasks} 
These tasks introduce controlled distributional shifts to test generalization under unseen physical conditions. In ALETHEIA we instantiate three physically motivated OOD regimes, each applied to the forward, inverse, temporal, and surface-to-source tasks:

\begin{itemize}
    \item \textbf{Frequency-band OOD (Low/Mid/High):} The 10 excitation frequencies are partitioned into low, mid, and high bands (e.g., \(1/4\) kHz, \(25/36\) kHz, \(81/100\) kHz). For each band, it trains on the remaining frequencies and tests only on the held-out band, probing the ability of models to extrapolate across qualitatively different penetration depths.
    
    \item \textbf{Single-frequency OOD (SFO):} For a given defect and grid type, the model is trained on a \emph{single} excitation frequency and evaluated on the remaining frequencies. This setting isolates how much multi-frequency supervision is needed for stable inverse reconstruction, and quantifies the degradation when moving from multi-frequency to single-frequency training. In these protocols, excitation frequencies define training and evaluation conditions; the baselines do not explicitly fuse measurements from multiple excitation frequencies in a single inference.
    
    \item \textbf{Geometric OOD (Geom-OOD):} The model is trained on samples from five defect types and tested on the held-out sixth type (e.g., Type II multi-layer defect). This evaluates extrapolation to unseen defect geometries while keeping the excitation protocol fixed.
\end{itemize}

To further enhance diversity and realism, each task is evaluated on both \textbf{regular} and \textbf{irregular} spatial grids. This enables us to systematically investigate the robustness of neural operators under varying spatial discretization schemes, mimicking real-world sensing constraints.

\subsection{Experimental Setup}
\label{subsection: experimental setup}

We benchmark all models on both regular and irregular 3D geometries using simulations from all defect types. Each sample was downsampled to 8000 points to achieve 2D-3D mapping alignment between surface temperatures and internal heat sources. Each training set comprised 600 samples from multiple simulation instances. Prior to training, all input data were normalized using global statistics computed from the training set; the same statistics were applied to validation and test inputs.

To handle the dataset, we implement a VTU parser that supports streaming and normalization, and define one in-distribution and three OOD partitioning schemes. In the normal setting, all 10 excitation frequencies are randomly shuffled at the sample level. For frequency-band OOD, we hold out one band (Low: 1/4 kHz, Mid: 25/36 kHz, High: 81/100 kHz) for testing and train on the remaining frequencies. For single-frequency OOD (SFO), models are trained on a single frequency and evaluated on the other frequencies data. For geometric OOD (Geom-OOD), models are trained on several kinds of defect types and tested on the other unseen type. These splits assess generalization to unseen excitation conditions and defect geometries beyond random train–test shuffling.

We compare a range of baseline and state-of-the-art neural operator models. These include Fourier-based models such as FNO~\citep{FNO}, FFNO~\citep{FFNO}, FCNO~\citep{li2024component}, and GeoFNO~\citep{Geo-FNO}, as well as attention-based models like LNO~\citep{LNO} and Transolver~\citep{Transolver}. Additionally, we have introduced LaMO~\citep{lamo} which is based on Mamba~\citep{mamba, mamba2}, as well as including DeepONet and a simple MLP as baseline models to provide a fair and comprehensive comparison.

All models were trained under the same configurations, please refer to \Cref{appendix: Experimental Setup Details} for more details.

\subsection{Experimental Results}

We report representative operator-learning performance under both in-distribution and OOD settings using diverse metrics (e.g., MSE, RMSE, SSIM; see \Cref{appendix: Metrics} for details). \Cref{fig:geom-ood-radar} reports a subset of Geom-OOD results, while \Cref{tab:summary} provides a concise overview of T2Q and S2Q performance across normal and OOD regimes on both irregular and regular grids. More detailed quantitative results and visualizations can be found in \Cref{appendix: More experiments}. 

For the representative geometric-OOD split, \Cref{tab:test6_mean_std} reports mean and sample standard deviation over three training runs for T2Q in panel (a) and S2Q in panel (b), each evaluated on regular and irregular grids.

\begin{figure}[htbp]
    \centering
    \includegraphics[width=1\textwidth]{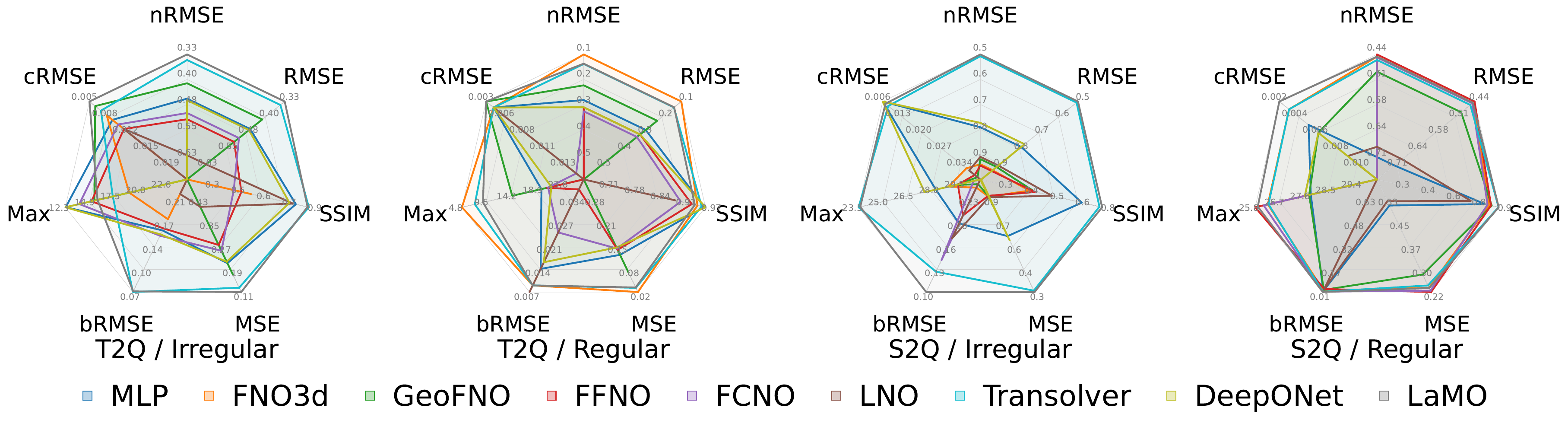}
    \caption{Comparison of nine models on S2Q and T2Q using seven metrics on different grids. The results were derived from geometric OOD, testing on the held-out Type II multi-layer defect.}
    \label{fig:geom-ood-radar}
\end{figure}

\begin{table}[htbp]
    \scriptsize
    \setlength{\tabcolsep}{3pt}
    \renewcommand{\arraystretch}{1.4}
    \centering
    \caption{Concise summary of T2Q and S2Q results under the Normal setting and three OOD regimes (MSE): OOD (averaged over High/Mid/Low frequency bands), SFO (averaged over 9/36/81 kHz single-frequency training), and Geom-OOD (geometric OOD, the equal-weight macro-average over six splits, each holding out one defect type), evaluated on irregular and regular grids.}
    \label{tab:summary}
    \begin{tabular}{l | cc | cc | cc | cc | cc | cc | cc | cc}
        \toprule
        & \multicolumn{8}{c|}{Irregular Data}
        & \multicolumn{8}{c}{Regular Data} \\
        \cmidrule{2-17}
        \multicolumn{1}{c|}{Model}
        & \multicolumn{2}{c|}{Normal}
        & \multicolumn{2}{c|}{OOD}
        & \multicolumn{2}{c|}{SFO}
        & \multicolumn{2}{c|}{Geom-OOD}
        & \multicolumn{2}{c|}{Normal}
        & \multicolumn{2}{c|}{OOD}
        & \multicolumn{2}{c|}{SFO}
        & \multicolumn{2}{c}{Geom-OOD} \\
        \cmidrule{2-17}
        & T2Q & S2Q & T2Q & S2Q & T2Q & S2Q & T2Q & S2Q
        & T2Q & S2Q & T2Q & S2Q & T2Q & S2Q & T2Q & S2Q \\
        \midrule
        MLP        & 0.218 & 0.510 & 0.248 & 0.649 & 0.263 & 0.629 & 0.241 & 0.614 & 0.058 & 0.484 & 0.067 & 0.518 & 0.119 & 0.537 & 0.090 & 0.501 \\
        FNO        & 0.175 & 0.292 & 0.232 & 0.424 & 0.255 & 0.716 & 0.390 & 0.773 & 0.029 & 0.092 & 0.038 & 0.119 & 0.119 & 0.302 & 0.066 & 0.221 \\
        GeoFNO     & 0.217 & 0.309 & 0.257 & 0.352 & 0.287 & 0.396 & 0.249 & 0.566 & 0.061 & 0.093 & 0.075 & 0.147 & 0.164 & 0.164 & 0.089 & 0.222 \\
        FFNO       & 0.158 & 0.479 & 0.206 & 0.561 & 0.196 & 0.614 & 0.278 & 0.778 & 0.032 & 0.097 & 0.040 & 0.124 & 0.096 & 0.223 & 0.077 & 0.207 \\
        FCNO       & 0.178 & 0.653 & 0.218 & 0.732 & 0.224 & 0.766 & 0.267 & 0.859 & 0.035 & 0.112 & 0.044 & 0.138 & 0.104 & 0.247 & 0.080 & 0.223 \\
        LNO        & 0.324 & 0.258 & 0.365 & 0.369 & 0.416 & 0.393 & 0.387 & 0.541 & 0.128 & 0.194 & 0.204 & 0.239 & 0.333 & 0.309 & 0.270 & 0.342 \\
        Transolver & 0.112 & 0.160 & 0.135 & 0.245 & 0.175 & 0.347 & 0.175 & 0.396 & 0.041 & 0.098 & 0.051 & 0.147 & 0.137 & 0.324 & 0.070 & 0.245 \\
        DeepONet   & 0.199 & 0.471 & 0.231 & 0.635 & 0.244 & 0.616 & 0.233 & 0.615 & 0.043 & 0.571 & 0.052 & 0.652 & 0.113 & 0.683 & 0.084 & 0.599 \\
        LaMO       & 0.077 & 0.169 & 0.125 & 0.247 & 0.177 & 0.363 & 0.175 & 0.381 & 0.039 & 0.098 & 0.050 & 0.163 & 0.126 & 0.315 & 0.069 & 0.240 \\
        \bottomrule
    \end{tabular}
\end{table}

Across the four T2Q configurations, FFNO provides consistently strong recovery of volumetric sources from full-field temperature data on both irregular and regular meshes. Its global Fourier layers smooth frequency-structured thermodynamics and support stable extrapolation when excitation frequencies shift, so its MSE typically remains at a low level under Normal, OOD(high/mid/low), and SFO settings. Although it is not always the top performer (FNO has the lowest regular-grid T2Q MSE under Normal, frequency-band OOD, and Geom-OOD in \Cref{tab:summary}), especially in the hardest geometric OOD regime, FFNO generally offers a tight upper envelope on core error metrics, indicating that spectral operators capture large-scale thermal conduction with limited aliasing while preserving global accuracy more effectively than most alternatives. To simulate real-world conditions, we designed a noise injection experiment, as detailed in \Cref{appendix: noise analysis}. Performance on the most challenging held-out defect types is further analyzed in \Cref{appendix: worst family analysis}.

Transolver demonstrates superior performance in S2Q, where only the surface temperature is accessible and internal states remain unobservable. Across regular and irregular layouts it remains competitive, with performance depending on the grid in the geometric OOD regime, which is closest to real deployment where crack geometries differ from those seen during training: in this setting LaMO attains the lowest irregular-grid macro-average S2Q MSE (0.381), while FFNO leads on regular grids (0.207); Transolver obtains 0.396 and 0.245, respectively. On irregular grids, Transolver attains the lowest S2Q MSE under the Normal, OOD and SFO settings. Its Physics-Attention softly assigns mesh points to learnable slices and attends among the resulting slice tokens, giving an adaptive receptive field. 

\begin{figure}[htbp]
    \centering
    \includegraphics[width=1\textwidth]{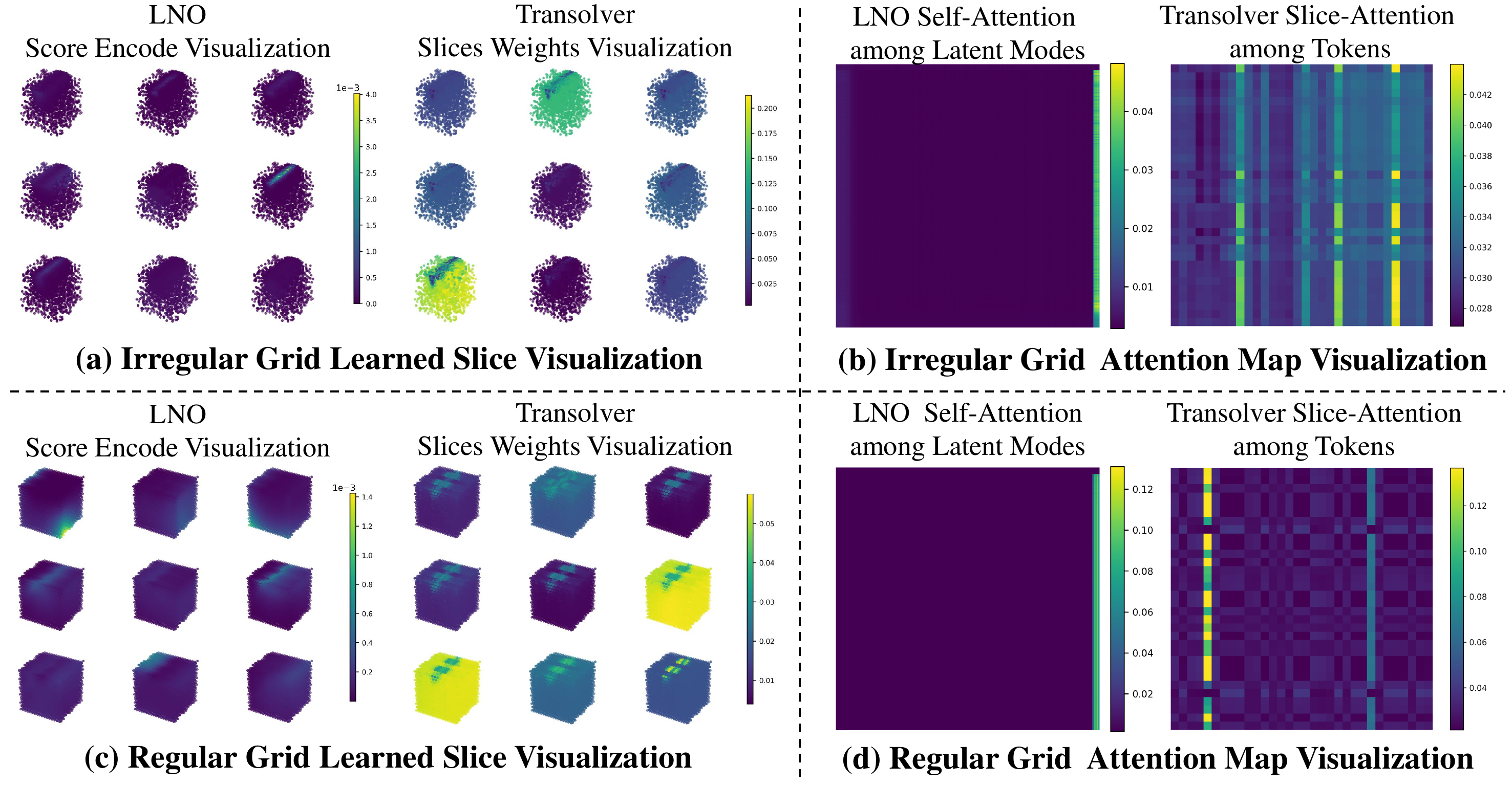}
    \caption{Attention visualization in the S2Q task for Type I double-layer defect with different grid: (a) slice weights in the last layer of Transolver and LNO for irregular grid, (b) attention maps of the last layer in Transolver and LNO for irregular grid, (c) slice weights in the last layer of Transolver and LNO for regular grid, (d) attention maps of the last layer in Transolver and LNO for regular grid.}
    \label{fig:attention_vis}
\end{figure}

\Cref{fig:attention_vis} visualizes the last-layer slice weights and attention maps of Transolver and LNO on Type~I double-layer defects. On both irregular and regular grids, Transolver's slices form compact clusters around the source layers, crack fronts and the informative top surface, whereas LNO's encoder scores are diffuse or nearly block-uniform. Transolver's slice attention also shows high-contrast block structure that routes information from surface and boundary slices to nearby interior sources, while LNO's latent attention is near-uniform. Adaptive slices thus segment the domain into physically coherent regions and concentrate capacity on thin boundary layers, consistent with Transolver's lower S2Q MSE in many (though not all) configurations. Further analysis of attention can be found in \Cref{appendix: physics attention analysis}.

The experimental results of LaMO are comparable to those of Transolver. Given that LaMO utilizes the same slicing technique as Transolver, with the sole modification being the substitution of the core attention layer with a Mamba layer, the underlying principles of the two models are analogous. Consequently, further detailed analysis is omitted herein.

It should be noted that, compared to T2Q with full‐field observations, S2Q only accesses top surface temperatures and cannot directly observe the interior, substantially reducing identifiability and exacerbating ill posedness. Empirically, almost all models degrade in S2Q, with the drop most pronounced in irregular meshes and under OOD excitations. This is consistent with the interpretation that limited observations and sparse/heterogeneous sampling further weaken the identifiability of the inverse problem and promote error accumulation across space and frequency. This mirrors the core challenge of NDT: inferring internal heat sources/defects from boundary‐only measurements is inherently information-limited.

\section{Conclusion}
\label{section: conclusion}
We have developed and released the first large-scale, multi-frequency coupled 3D Pulsed Eddy Current Thermography benchmark dataset, aimed at advancing the application of machine learning methods in internal crack detection for NDT. This dataset is derived from high fidelity electromagnetic thermal multiphysics simulation data and calibrated with real infrared thermal imaging experimental data, covering various defect types and frequency response scenarios. It supports a range of forward and inverse modeling tasks. Benchmark results show that neural operator models can effectively learn the dynamic process of heat diffusion and reconstruct the primary internal heat source distribution from surface temperature sequences. However, challenges remain in accurately reconstructing fine-grained three-dimensional morphologies of complex defects.

This dataset is the first to encompass the electromagnetic-thermal coupling response of rail materials under different excitation frequencies, providing a standardized testing platform for research into neural network surrogate models, operator learning methods, and defect inversion. Looking ahead, we plan to extend this framework to additional materials, such as composites and pipeline structures, as well as to a broader range of NDT scenarios, further promoting the widespread application of data-driven modeling in industrial inspections. We hope that the Aletheia benchmark will serve as a foundational resource for multi-physics modeling and model generalization research. 

\section{Acknowledgements \& Disclosures}
\label{section: ack}
This work was supported in part by the National Natural Science Foundation of China under Grant Nos. 62406206 and 62203312, the Fundamental Research Funds for the Central Universities, and the Natural Science Foundation of Sichuan Province of China, under Grant No. 2024NSFSC1484. The authors declare no conflict of interest.

\bibliography{Aletheia_ref}
\bibliographystyle{plain}


\newpage
\appendix

\section{Simulation Experiment Details}
\label{appendix: Simulation experiment details}
In COMSOL Multiphysics, the simulation domain consisted of a steel rail geometry with dimensions set to 100 $\times {70} \times {45}$ which is embedded artificial defects of configurable size, orientation, and depth, surrounded by air domain; The rail material used in the physical experiments was U71Mn steel, and the material parameters used in the simulation were modified from U71Mn properties based on calibration against experimental surface temperature measurements. The constant material properties for steel, air, and excitation coil were assigned as listed in \Cref{Material properties-table}. All external surfaces of the rail were thermally insulated except for the top surface, which had a convective cooling boundary condition to ambient air, with a heat transfer coefficient set to 5 W/($\mathrm{m^{2}}$$\cdot$K); the excitation coil was driven via an applied current boundary condition to simulate inductive heating. The initial temperature of the entire model was uniformly set to 20 °C. A free tetrahedral mesh was employed, featuring fine elements around defect regions and within the electromagnetic skin depth, while coarser elements were used elsewhere; an adaptive refinement scheme further adjusted the mesh based on local conductivity gradients. The simulation was run using a time-dependent solver with a backward differentiation formula (BDF) scheme, covering a time span from 0 to 0.6 s with a fixed time step of 0.05 s.

\begin{table}[ht]
  \caption{Material property parameters. In the ECPT experiment, due to the extremely short heating process and relatively small temperature rise changes, the impact on the various properties of the material can be ignored. Therefore, the material's property parameters are set to a fixed constant. $\sim$ represents the physical quantity of the material that is not involved in actual calculations.}
  \label{Material properties-table}
  \setlength{\tabcolsep}{4.5pt}
  \centering
  \begin{tabular}{ccccccc}
    \toprule
    \multirow{3}{*}{Material} & \multirow{2}{*}{Conductivity} & \multirow{2}{*}{Relative} & Relative  & Thermal  & \multirow{2}{*}{Density} & Constant pressure  \\
    & \multirow{2}{*}{(S/m)} & \multirow{2}{*}{permeability} & dielectric & conductivity & \multirow{2}{*}{(kg/m$^{3}$)} & heat capacity \\
    & &  & constant & (W/m$\cdot$K) &  & (J/kg$\cdot$K) \\
    \midrule
    Steel & \num{1.3e7} & 200 & 2 & 48 & 8000 & 450 \\ 
    Air & 0 & 1 & 1 & 0.0257 & 1.205 & 1005 \\ 
    Copper & \num{5.998e7} & 1 & 1 & $\sim$ & $\sim$ & $\sim$ \\ 
    \bottomrule
  \end{tabular}
\end{table}

\section{Data Generation and Details}
\label{appendix: Data generation and details}
The diversity of defect samples in our dataset is achieved through systematic variations in defect orientation, depth, and length. For single-layer defects, the angle between the defect extension direction and the upper surface differs from sample to sample. In double-layer and multi-layer defects, not only do the individual cracks vary in their inclination to the surface, but the angles between multiple cracks also change. These controlled variations result in a wide range of geometric configurations. Detailed statistical parameters of defect angles are presented in \Cref{angle-table}.

\begin{table}[htbp]
  \caption{Angle parameters for different types of defects. Among them, angle 1 represents the angle formed between the defect direction and the upper surface, while angles 2 and 3 represent the angles between multiple cracks. All units are based on angle system.}
  \label{angle-table}
  \centering
  \begin{tabular}{cccc}
    \toprule
    \multirow{2}{*}{\textbf{Defect type}} & \multicolumn{3}{c}{\textbf{Range of different angles}} \\ 
        \cmidrule{2-4} 
         & angle I & angle II & angle III \\
         \midrule
        Single layer defect & 3$\sim$90 & - & - \\ 
        Type \Rmnum{1} double-layer defect & 24.47$\sim$75.47 & 5.72$\sim$81.27 & - \\ 
        Type \Rmnum{2} double-layer defect & 3$\sim$71 & - & - \\
        Type \Rmnum{3} double-layer defect & 10.06$\sim$80.06 & 0.83$\sim$24.03 & - \\ 
        Type \Rmnum{1} multi-layer defect & 0$\sim$89 & - & - \\ 
        Type \Rmnum{2} multi-layer defect & 30$\sim$90 & 12.09$\sim$31.1 & 12.12$\sim$31.12 \\ 
        \bottomrule
    \end{tabular}
\end{table}

The dataset contains a total of 4,782 samples, with an overall size of approximately 1.89 terabytes. These samples are organized into six main folders, each corresponding to a specific defect type and named accordingly. Each sample folder is labeled by its defect type followed by one or more angle values that indicate the orientation of the defect. For example, names may include a single angle for single-layer defects or two to three angles for more complex configurations. Within each sample folder, there are three subfolders storing different types of data. The first subfolder contains surface temperature data sampled on a structured grid. The second provides full three-dimensional temperature and heat field data, also on a structured grid with a uniform spacing of one unit. The third subfolder includes similar 3D data, but sampled on an unstructured grid. The unstructured sampling strategy, which distributes 50,000 points per sample, is detailed in the main text. \Cref{Sample collection point statistics table} summarizes the number of sampling points for each data type. 

\begin{table}[htbp]
  \caption{Sampling point statistics for different types of data. Here, $\mathrm{T_{surf}}$ denotes surface temperature measurements, \( \{q(\mathbf{x}, t), u(\mathbf{x}, t)\}_{\text{structured}} \) represents paired heat source and temperature field data sampled on regular grids, and \( \{q(\mathbf{x}, t), u(\mathbf{x}, t)\}_{\text{unstructured}} \) denotes the same sampled irregularly.}
  \label{Sample collection point statistics table}
  \centering
  \begin{tabular}{ccccc}
    \toprule
        Data type & x & y & z & Sampling points \\ 
        \midrule
        $\mathrm{T_{surf}}$ & 150 & 105 & - & 15750 \\ 
        \( \{q, u\}_{\text{structured}} \) & 50 & 30 & 30 & 45000 \\ 
        \( \{q, u\}_{\text{unstructured}} \) & - & - & - & 50000 \\ 
        \bottomrule
    \end{tabular}
\end{table}

Each sample is simulated under ten excitation frequencies ranging from 1 to 100 kHz, with each frequency corresponding to one VTU format file. Both surface and volumetric temperature data are time-resolved from 0 to 0.6 seconds, with a time step of 0.05 seconds, resulting in 13 time-series data stored in each VTU file. This structure ensures that the dataset supports diverse tasks involving spatial, temporal, and multi-frequency analysis.

\section{Defect Details}
\label{appendix: Defect Details}
\begin{figure}[htbp]
    \centering
    \includegraphics[width=1\textwidth]{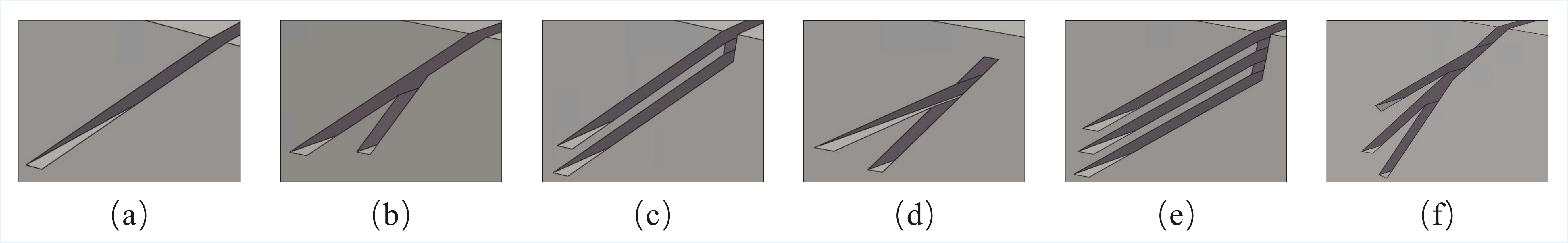}
    \caption{6 types of defects: (a)Single layer defect (b)Type \Rmnum{1} double-layer defect (c)Type \Rmnum{2} double-layer defect (d)Type \Rmnum{3} double-layer defect (e)Type \Rmnum{1} multi-layer defect (f)Type \Rmnum{2} multi-layer defect} 
    \label{fig: defect}
\end{figure}

Our dataset encompasses six distinct internal crack defect types, spanning single-layer, double-layer, and multi-layer configurations and including both open and closed crack cases. Six types of defects are shown in \Cref{fig: defect} and the detailed parameters of each defect are shown in \Cref{table: defect detailed parameters}.

\begin{table}[htbp]
    \caption{Detailed parameters of 6 types of defects. The units of depth, width, and length are all in millimeters. In double-layer and multi-layer defects, the length of the defect refers to the length of the longest crack.}
    \label{table: defect detailed parameters}
    \centering
    \begin{tabular}{ccccccc}
        \toprule
        Defect type & Sample size & Depth & Width & Length & Open/Closed \\
        \midrule
        Single layer defect & 88 & 0.2$\sim$3.82 & 0.1$\sim$0.2 & 3.82 & Open  \\ 
        Type \Rmnum{1} double-layer defect & 1050 & 1.17$\sim$4.1 & 0.2 & 4.53 & Open \\ 
        Type \Rmnum{2} double-layer defect & 69 & 0.24$\sim$4.02 & 0.1$\sim$0.2 & 4.22 & Open \\ 
        Type \Rmnum{3} double-layer defect & 2285 & 1.40$\sim$3.75 & 0.2 & 3.29 & Closed \\ 
        Type \Rmnum{1} multi-layer defect & 90 & 0.81$\sim$2.41 & 0.2 & 2.4 & Closed \\ 
        Type \Rmnum{2} multi-layer defect & 1200 & 1.89$\sim$3.12 & 0.15 & 3.12 & Open \\ 
        \bottomrule
    \end{tabular}
\end{table}

\section{Defect Reconstruction from Heat Field}
\label{appendix: Defect reconstruction}
We conducted a simple test using a basic 3D CNN to reconstruct the heat source field \(Q(x,y,z)\). The results demonstrate that \(Q\) can effectively capture defect morphology: high-Q-value dense zones align with defect surfaces, Q-value decay indicates defect propagation direction, and low-Q-value regions correspond to internal cavities. Leveraging a 3D CNN with a large receptive field, we extract global spatial features from the reconstructed Q-field and use a regression network to predict key defect parameters such as angle, length, and depth. Experimental results (\Cref{Regression evaluation indicators}) show reliable predictions for these parameters. The width parameter shows low variance in the dataset, resulting in negligible predictive signal rather than a model limitation. Therefore, as long as our neural operator can reconstruct \(Q\) accurately, it can serve as a reliable basis for predicting multidimensional defect characteristics.

\begin{table}[htbp]
    \caption{Regression evaluation indicators for crack parameters}
    \label{Regression evaluation indicators}
    \centering
    \begin{tabular}{c | ccccc}
        \toprule
        Attributes & Model & MSE & RMSE & MAE  \\ 
        \midrule
        \multirow{3}{*}{Crack angle}    & MLP        & 0.23 & 0.48 & 0.38  \\ 
                                        & FNO        & 0.17 & 0.41 & 0.35  \\ 
                                        & Transolver & 0.20 & 0.44 & 0.30  \\ 
        \midrule
        \multirow{3}{*}{Crack length}   & MLP        & 0.0020 & 0.044 & 0.034  \\ 
                                        & FNO        & 0.0012 & 0.035 & 0.028  \\ 
                                        & Transolver & 0.0011 & 0.033 & 0.020  \\ 
        \midrule
        \multirow{3}{*}{Crack depth}    & MLP        & 1.24  & 1.11  & 0.91  \\ 
                                        & FNO        & 0.85 & 0.92 & 0.80  \\ 
                                        & Transolver & 1.02  & 1.01  & 0.70  \\ 
        \midrule
        \multirow{3}{*}{Crack width}    & MLP        & \num{1.00e-14} & \num{1.00e-7} & \num{7.86e-8}  \\ 
                                        & FNO        & \num{1.26e-14} & \num{1.12e-7} & \num{1.10e-7}  \\ 
                                        & Transolver & \num{1.43e-14} & \num{1.19e-7} & \num{1.16e-7}  \\ 
        
        \bottomrule
    \end{tabular}
\end{table}

\section{Metrics}
\label{appendix: Metrics}
Standard methods for calculating the root mean square error (RMSE) of test data fail to capture important optimization criteria in scientific machine learning. It is not enough to fit (usually sparse) data well if the physical laws of the underlying problem are seriously violated. Therefore, they must be evaluated using appropriate metrics. Furthermore, a single evaluation metric is not sufficient to compare differences in the ability of different methods to infer unseen time steps and parameters, which are important but not yet fully explored evaluation criteria for machine learning alternative models. We used the PDEBench evaluation metrics and the SSIM evaluation metrics, as shown in \Cref{tab:metrics}

\begin{table}[htbp]
    \centering
    \setlength{\tabcolsep}{4.5pt}
    \renewcommand{\arraystretch}{1.2}
    \caption{Evaluation indicators provided by PDEBench}
    \label{tab:metrics}
    \begin{tabular}{lll}
        \toprule
        Scope & Acronym & Metric \\ 
        \midrule
        \multirow{4}{*}{Data view} & RMSE & root-mean-squared-error \\ 
        \cmidrule(r){2-3}
         & nRMSE & normalized RMSE (ensuring scale independence) \\ 
        \cmidrule(r){2-3}
         & \multirow{2}{*}{max error} & maximum error \\ 
         & &  (local worst case; also proxy for stability of time-stepping) \\ 
        \midrule 
        \multirow{8}{*}{Physics view} &  \multirow{2}{*}{cRMSE} & RMSE of conserved value  \\ 
         & &  (deviation from conserved physical quantity) \\ 
         \cmidrule(r){2-3}
         &  \multirow{2}{*}{bRMSE} & RMSE on boundary  \\ 
         & &  (whether boundary condition can be learned) \\ 
         \cmidrule(r){2-3}
         &  \multirow{2}{*}{fRMSE low} & RMSE in Fourier space, low frequency regime  \\ 
         & &  (wavelength dependence) \\ 
         \cmidrule(r){2-3}
         & fRMSE mid & RMSE in Fourier space, medium frequency regime \\ 
         \cmidrule(r){2-3}
         & fRMSE high & RMSE in Fourier space, high frequency regime \\ 
        \bottomrule 
    \end{tabular}
\end{table}

The normalized RMSE is a variant of the RMSE to provide scale-independent information defined as:
\begin{equation}
    \mathrm{nRMSE} \equiv \frac{\| u_{\mathrm{pred}} - u_{\mathrm{true}} \|_2}{\| u_{\mathrm{true}} \|_2},
\end{equation}
where $\| u \|_2$ is the $L_2$-norm of a (vector-valued) variable $u$, and $u_{\mathrm{true}}, u_{\mathrm{pred}}$ are true and predicted values, respectively. The maximum error measures the model’s worst prediction, which quantifies both local performance and models’ stability of their prediction.

cRMSE is defined as
\begin{equation}
\mathrm{cRMSE} \equiv \frac{\left\| \sum u_{\mathrm{pred}} - \sum u_{\mathrm{true}} \right\|_2}{N},
\end{equation}
which measures the deviation of the prediction from some physically conserved value.

bRMSE measures the error at the boundary, indicating if the model understands the boundary condition properly.

Finally, fRMSE measures the error in low/middle/high-frequency ranges defined as:
\begin{equation}
    \sqrt{
        \frac{ \sum\limits_{k_{\min}}^{k_{\max}} \left| \mathcal{F}(u_{\mathrm{pred}}) - \mathcal{F}(u_{\mathrm{true}}) \right|^2 }
        { k_{\max} - k_{\min} + 1 }
    },
\end{equation}
where $\mathcal{F}$ is a discrete Fourier transformation, and $k_{\min}, k_{\max}$ are the minimum and maximum indices in Fourier coordinates. 

In PDEBench paper, the low/middle/high-frequency regions are defined as:
\begin{itemize}
    \item Low: $k_{\min} = 0$, $k_{\max} = 4$
    \item Middle: $k_{\min} = 5$, $k_{\max} = 12$
    \item High: $k_{\min} = 13$, $k_{\max} = \infty$
\end{itemize}

This allows a quantitative discussion of the model performance’s dependence on the wavelength. In the multidimensional cases, the $|\mathcal{F}(u_{\mathrm{pred}} - u_{\mathrm{true}})(k)|^2$ in the angular coordinate direction is first integrated and summed along the $k$ coordinate.

\section{Experimental Setup Details}
\label{appendix: Experimental Setup Details}
To ensure a fair comparison, this section details all hyperparameters and training configurations employed across the models. For the FNO family of models, given their similar parameter sets, these hyperparameters are presented in \Cref{tab:hyperparameters}.

\begin{table}[htbp]
    \setlength{\tabcolsep}{4.5pt}
    \caption{Shared architectural settings for the FNO family baselines. All models use the same spectral bandwidth and channel widths to ensure fairness. $(m_x,m_y,m_z)$ denotes spectral modes, $Width$ denotes the linear transformation applied on the spatial domain, $Grid / scale$ denotes the resolution of the grid.}
    \label{tab:hyperparameters}
    \centering
    \begin{tabular}{l c c c c c }
    \toprule
    Model  & $(m_x,m_y,m_z)$ & $Width$  & $Grid / scale$ & Model Variant \\
    \midrule
    FNO    & $(12,12,8)$ & 32 & $(20,20,20)$ & Original Fourier Neural Operator  \\
    GeoFNO & $(12,12,8)$ & 32 & $(20,20,20)$ & Geometry-Aware FNO \\
    FFNO   & $(12,12,8)$ & 32 & $(20,20,20)$ & Factorized FNO  \\
    FCNO   & $(12,12,8)$ & 32 & $(20,20,20)$ & Factorized Cosine Neural Operator  \\
    \bottomrule
    \end{tabular}
\end{table}

\textbf{GeoFNO.}
We adapt its coordinate handling to three-dimensional meshes by using point coordinates for the sampling and query locations and making IPHI's final layer output three coordinate-wise deformation factors. The core Fourier operator architecture remains unchanged.

\textbf{Transolver.}
We instantiate Transolver with hidden size $n_{hidden}=256$, depth $n_{layers}=8$, and spatial dimension $space_{dim}=3$. Multi-head attention uses $n_{head}=8$ with an MLP expansion ratio of $2$. We further use $slice_{num}=32$ and disable positional unification. The same configuration is used for all tasks.

\textbf{LaMO.}
We instantiate LaMO with hidden size $n_{hidden}=256$, depth $n_{layers}=8$, and spatial dimension $space_{dim}=3$. Mamba modules uses $n_{head}=8$ with an MLP expansion ratio of $2$. We further use $slice_{num}=32$ and disable positional unification. The same configuration is used for all tasks.

\textbf{LNO.}
The LNO is consisting of $n_{block}=4$ operator blocks with spectral resolution $n_{mode}=256$ and hidden width $n_{\mathrm{dim}}=128$. Attention uses $n_{\mathrm{head}}=8$ and $n_{layer}=2$ transformer layers with \texttt{GELU} activations and the vanilla attention kernel; temporal modeling is disabled.

\textbf{MLP.}
We employ a point-wise multilayer perceptron with hidden width $32$, and $4$ layers in total, serving as a lightweight regression baseline.

\textbf{DeepONet.}
The DeepONet baseline adopts a branch--trunk decomposition with width $32$ in both sub-networks. The branch network has depth $2$; the trunk network has depth $3$, receiving the functional input of dimension $13$ and producing a scalar output of dimension $1$. All other settings follow the original implementation.

All models were trained under the same configuration: batch size of 4, learning rate of 0.0001, and 200 training epochs. Training employed the Adam optimizer alongside the OneCycleLR learning rate scheduler. All experiments were conducted on a single NVIDIA RTX 4090 GPU with 24GB of memory. The models were trained using mean squared error (MSE) loss, and evaluated using SSIM along with several PDEBench metrics. Regarding the part of evaluation indicators, please refer to \Cref{appendix: Metrics} for detailed definitions.

With this configuration, GeoFNO consumed the most VRAM (approximately 22 GB) and took the longest to complete (90 minutes), and Transolver consumed 6GB VRAM and 30 minutes training time, all experiments were conducted in a single GPU environment. We do not discuss the consumption of computational resources in further detail.

Detailed configurations reference \Cref{tab:train_configuration}. Apart from DeepONet, each model permits customisation of input and output channels. Due to DeepONet's constraint that its output channel must be limited to $1$, DeepONet participated solely in the T2Q and S2Q tasks.

\begin{table}[htbp]
  \centering
  \scriptsize
  \setlength{\tabcolsep}{4.5pt}
  \caption{Training configuration across tasks. Points denote the number of samples. \emph{In $\to$ Out ch.} denotes input and output channels, \emph{Sample Points} denotes the number of downsampling points for each data, \emph{Epochs} denotes total training epochs, \emph{LR} denotes initial learning rate, \emph{Sample Data} denotes the number of simulated data used for training, \emph{Optimizer} denotes the optimiser employed for training, \emph{Batch} denotes the batch size used for training, \emph{Training Times} denotes the number of repeated training sessions for each task}
  \label{tab:train_configuration}
  \begin{tabular}{l|cccccccc}
    \toprule
    Task & In $\to$ Out ch. & Sample Points & Epochs & LR & Sample Data  & Optimizer & Batch & Training Times \\
    \midrule
    T2Q  & $13 \to 1$  & $8\,000$ (vol.)  & \multirow{4}{*}{200} & \multirow{4}{*}{$1\times 10^{-4}$} & \multirow{4}{*}{600} & \multirow{4}{*}{Adam}  & \multirow{4}{*}{4} & \multirow{4}{*}{3}  \\
    S2Q  & $13 \to 1$  & $8\,000$ (vol.) $+$ $8\,000$ (surf.) &  &  &  &  &  &   \\
    T2T  & $11 \to 2$  & $8\,000$ (vol.)                      &  &  &  &  &  &   \\
    Q2T  & $1 \to 13$  & $8\,000$ (vol.)                      &  &  &  &  &  &   \\
    \bottomrule
  \end{tabular}
\end{table}


\section{Noise Analysis}
\label{appendix: noise analysis}
\begin{table}[htbp]
    \scriptsize
    \setlength{\tabcolsep}{2.5pt}
    \renewcommand{\arraystretch}{1.4}
    \centering
    \caption{S2Q Surface-to-Source Reconstruction errors (MSE / RMSE) for geometric OOD (Geom-OOD, train on defect types 1--5 and test on type~6) under different noise levels (Normal, Low-Noise, Mid-Noise, High-Noise), evaluated on irregular and regular grids.}
    \label{tab:s2q_fsplit_noise}
    \begin{tabular}{l | cc | cc | cc | cc | cc | cc | cc | cc}
        \toprule
         & 
        \multicolumn{8}{c|}{Irregular Data} &
        \multicolumn{8}{c}{Regular Data} \\
        \cmidrule{2-17}
         \multicolumn{1}{c|}{Model}  &
         \multicolumn{2}{c|}{Normal} &
          \multicolumn{2}{c|}{Low-Noise} &
          \multicolumn{2}{c|}{Mid-Noise} &
          \multicolumn{2}{c|}{High-Noise} &
          \multicolumn{2}{c|}{Normal} &
          \multicolumn{2}{c|}{Low-Noise} &
          \multicolumn{2}{c|}{Mid-Noise} &
          \multicolumn{2}{c}{High-Noise} \\
        \cmidrule{2-17}
        & MSE & RMSE & MSE & RMSE & MSE & RMSE & MSE & RMSE & MSE & RMSE & MSE & RMSE & MSE & RMSE & MSE & RMSE \\
        \midrule
        MLP        & 0.634 & 0.795 & 0.652 & 0.805 & 0.696 & 0.832 & 1.019 & 1.007 & 0.519 & 0.719 & 0.565 & 0.772 & 0.570 & 0.781 & 0.647 & 0.837 \\
        FNO        & 0.886 & 0.939 & 0.912 & 0.967 & 0.921 & 0.973 & 0.942 & 0.995 & 0.232 & 0.445 & 0.246 & 0.459 & 0.248 & 0.462 & 0.255 & 0.471 \\
        GeoFNO     & 0.891 & 0.919 & 0.899 & 0.937 & 0.927 & 0.961 & 1.002 & 1.065 & 0.278 & 0.486 & 0.292 & 0.513 & 0.421 & 0.636 & 0.658 & 0.814 \\
        FFNO       & 0.892 & 0.944 & 1.018 & 1.007 & 1.034 & 1.015 & 1.112 & 1.051 & 0.217 & 0.439 & 0.242 & 0.463 & 0.256 & 0.492 & 0.296 & 0.548 \\
        FCNO       & 0.997 & 0.998 & 1.013 & 1.006 & 1.013 & 1.009 & 1.018 & 1.009 & 0.222 & 0.447 & 0.233 & 0.460 & 0.253 & 0.496 & 0.342 & 0.576 \\
        LNO        & 0.883 & 0.910 & 0.890 & 0.915 & 0.934 & 1.028 & 1.579 & 1.652 & 0.534 & 0.692 & 0.610 & 0.766 & 0.641 & 0.767 & 0.710 & 0.823 \\
        Transolver & 0.285 & 0.519 & 0.313 & 0.518 & 0.321 & 0.526 & 0.473 & 0.697 & 0.240 & 0.454 & 0.261 & 0.443 & 0.276 & 0.465 & 0.312 & 0.518 \\
        DeepONet   & 0.609 & 0.778 & 0.611 & 0.775 & 0.610 & 0.778 & 0.623 & 0.737 & 0.610 & 0.780 & 0.660 & 0.838 & 0.661 & 0.845 & 0.672 & 0.863 \\
        LaMO       & 0.276 & 0.512 & 0.343 & 0.571 & 0.392 & 0.631 & 0.453 & 0.663 & 0.231 & 0.445 & 0.257 & 0.402 & 0.316 & 0.534 & 0.348 & 0.527 \\
        \bottomrule
    \end{tabular}
\end{table}

In real engineering scenarios, structured disturbances can affect surface-temperature measurements. For example, the infrared instrument also records thermal emission from the excitation coil, introducing a disturbance into the measured surface-temperature signal. We separately acquired this coil-temperature component and injected it into the simulated surface observations at 20\%, 50\%, and 100\% of its recorded amplitude, corresponding to Low-, Mid-, and High-Noise, respectively. Noise is added only to the test inputs to simulate unexpected measurement disturbances under practical operating conditions; the training data and volumetric heat-source targets remain unchanged. Normal denotes testing without the added component. This is a controlled measurement-derived perturbation test, not a general sensor-noise model or a sim-to-real evaluation.

As shown in \Cref{tab:s2q_fsplit_noise}, we further investigate the robustness of the S2Q task under different noise levels, in a setting that more closely reflects real industrial NDT conditions. We conduct a noise-robustness study in the geometric OOD regime: all models are trained on defect types 1–5 and evaluated on the unseen Type II multi-layer defect under multiple noise settings on both irregular and regular grids. Overall, injecting noise into the surface temperature measurements leads to a gradual degradation of performance rather than catastrophic failure. On the irregular grid, most baselines exhibit noticeable increases in mean squared error (MSE) from the Normal to the High-Noise regime; for example, the purely data-driven MLP rises from 0.634 to 1.019 ($\approx +61\%$), and Transolver increases from 0.285 to 0.473 ($\approx +66\%$). On the regular grid, where the forward diffusion operator is better resolved, the same models show substantially smaller relative changes; MLP grows only from 0.519 to 0.647 ($\approx +24\%$), and Transolver from 0.240 to 0.312 ($\approx +30\%$). FNO exhibits modest MSE increases on both grids, whereas the response of other frequency-domain baselines depends on the grid and perturbation level.

FNO and Transolver display two distinct robustness patterns, consistent with their architectural biases. On the irregular grid, FNO’s MSE increases only slightly from 0.886 (Normal) to 0.942 (High-Noise), a relative change of about $6\%$; Transolver shows mild degradation under Low-Noise and Mid-Noise conditions, and although its performance drop becomes larger in the High-Noise setting, it remains competitive in absolute accuracy. This suggests that its physics-aware tokenization and attention mechanism can retain competitive reconstruction accuracy under the injected coil-temperature perturbation. In the most challenging geometric OOD scenario, degradation across neural operators is therefore expected but far from catastrophic. On the regular grid with high noise, FNO achieves the lowest absolute errors overall, with MSE increasing only from 0.232 to 0.255 ($\approx +10\%$), which aligns with its spectral bias toward smooth diffusion on structured meshes. Transolver’s absolute accuracy on the regular grid is slightly worse than that of FNO but remains competitive and significantly better than simpler baselines such as MLP and DeepONet.

Taken together, in a configuration that more faithfully approximates practical ECPT measurement conditions, all neural operators experience some performance loss under noise, but the degradation follows a structured and interpretable pattern. Regular grids are slightly more favorable than irregular point clouds, indicating that geometric and sampling uncertainty compounds the impact of noise; FNO exhibits particularly strong robustness and accuracy on structured meshes, whereas Transolver maintains strong robustness and low error under complex geometric OOD and varying noise levels. Benchmarking with ALETHEIA and analyzing how different neural-operator architectures trade off accuracy and robustness under realistic noise provides an evaluation setting that is much closer to engineering practice, and can guide the design of future neural operators with explicit uncertainty awareness and noise-suppression capabilities.

\FloatBarrier

\section{Worst-Family Analysis under Geometric OOD}
\label{appendix: worst family analysis}

The six geometric-OOD splits evaluate generalization to unseen defect types, with each type held out in turn and the remaining five used for training. To complement the macro-average MSE in \Cref{tab:summary}, we report the highest test MSE across the six held-out defect types.

\begin{table}[htbp]
    \centering
    \caption{Macro-average and maximum MSE across the six held-out defect types under geometric OOD. \textit{Mean} gives each defect type equal weight; \textit{Worst} denotes the maximum MSE. Lower values are better, and the lowest displayed values in each column are shown in bold.}
    \label{tab:geom_ood_worst}
    \begin{tabular}{l | cc | cc | cc | cc}
        \toprule
        & \multicolumn{4}{c|}{Irregular Data} & \multicolumn{4}{c}{Regular Data} \\
        \cmidrule(lr){2-5}
        \cmidrule(lr){6-9}
        Model & \multicolumn{2}{c|}{T2Q} & \multicolumn{2}{c|}{S2Q} & \multicolumn{2}{c|}{T2Q} & \multicolumn{2}{c}{S2Q} \\
        \cmidrule(lr){2-3}          
        \cmidrule(lr){4-5}
        \cmidrule(lr){6-7}
        \cmidrule(lr){8-9}
        & Mean & Worst & Mean & Worst & Mean & Worst & Mean & Worst \\
        \midrule
        MLP        & 0.241          & 0.416          & 0.614          & 0.720          & 0.090          & 0.223          & 0.501          & 0.679           \\
        FNO        & 0.390          & 0.503          & 0.773          & 0.923          & \textbf{0.066} & 0.202          & 0.221          & 0.466           \\
        GeoFNO     & 0.249          & 0.400          & 0.566          & 0.891          & 0.089          & 0.226          & 0.222          & 0.424           \\
        FFNO       & 0.278          & 0.389          & 0.778          & 0.916          & 0.077          & \textbf{0.177} & \textbf{0.207} & 0.428           \\
        FCNO       & 0.267          & 0.401          & 0.859          & 0.997          & 0.080          & 0.187          & 0.223          & \textbf{0.413} \\
        LNO        & 0.387          & 0.639          & 0.541          & 0.883          & 0.270          & 0.479          & 0.342          & 0.573           \\
        Transolver & \textbf{0.175} & \textbf{0.341} & 0.396          & 0.694          & 0.070          & 0.211          & 0.245          & 0.446           \\
        DeepONet   & 0.233          & 0.402          & 0.615          & 0.725          & 0.084          & 0.199          & 0.599          & 0.736           \\
        LaMO       & \textbf{0.175} & 0.360          & \textbf{0.381} & \textbf{0.669} & 0.069          & 0.201          & 0.240          & 0.435           \\
        \bottomrule
    \end{tabular}
\end{table}

On irregular grids, Transolver obtains the lowest maximum MSE across defect types for T2Q (0.341), followed by LaMO (0.360). Although their macro-average MSEs are nearly identical, Transolver has a lower error on its most challenging held-out defect type. For S2Q, LaMO achieves both the lowest macro-average MSE (0.381) and the lowest maximum MSE across defect types (0.669), followed by Transolver.

On regular grids, the rankings differ between the two summaries. FNO has the lowest T2Q macro-average MSE (0.066), whereas FFNO has the lowest maximum MSE across defect types (0.177). Similarly, FFNO achieves the lowest S2Q macro-average MSE (0.207), while FCNO obtains the lowest maximum MSE (0.413).

These results show that strong average performance does not necessarily imply the smallest error on the most challenging unseen defect type.  Reporting both summaries provides a fuller assessment of geometric-OOD performance.

\section{Physics Attention Analysis}
\label{appendix: physics attention analysis}

The S2Q task is governed by a smooth diffusive background with sharply localized anomalies at material interfaces and crack tips: the volumetric heat source is piecewise smooth in the interior but exhibits strong gradients near the upper surface and along the defect boundaries. Successful inversion therefore requires the model to selectively attend to these “physically active’’ regions, rather than treating all interior voxels uniformly. Transolver is explicitly designed for this purpose through Physics-Attention, which first projects each mesh point feature into a low-entropy slice weight over learnable slices and then aggregates points within each slice into physics-aware tokens, followed by attention among tokens.  This bottom-up decomposition pushes points with similar geometry and temperature patterns into the same slice, and the Softmax over the slice dimension sharpens the assignment so that slices specialize to distinct physical states.

To verify that this mechanism also emerges on Aletheia, we visualize in \Cref{fig:attention_vis}(a, c), the learned slice weights in the last Transolver layer and compare them with the encoder scores of LNO on both irregular and regular grids for Type I double-layer defect. On the irregular grids, Transolver’s slices form several compact 3D clusters that hug the vicinity of the defect source and the crack front, while assigning low weights to homogeneous interior regions. In contrast, the LNO score-encode fields are much more diffuse: the high-response regions spread over large portions of the volume and show weak alignment with the true source interfaces. On regular grids, this gap is also quite obvious. Transolver learns slice volumes that sharply align with the two latent source layers and their vertical transition zones, producing box-like clusters concentrated near the interfaces and near the top surface where the measured temperature is most informative for inversion. LNO again yields almost block-uniform scores inside each cube, indicating that its latent representation does not clearly separate boundary and background states.

We further compare the attention maps in \Cref{fig:attention_vis}(b, d). For Transolver, attention is computed among physics-aware tokens, each summarizing one slice. The resulting slice–attention matrices exhibit structured, high-contrast patterns: a small subset of “boundary’’ slices receive high attention, and form blocks of strong interactions with their neighboring interior slices. This reflects a physically meaningful message-passing pattern: information is routed from the surface and boundary slices into nearby interior slices where the heat source lies. By contrast, LNO encodes point features into latent modes, applies attention in the latent space, and decodes the resulting features back to mesh points. The corresponding 2D attention maps are markedly closer to a degenerate pattern: most rows are near-uniform with a few narrow bright columns, and there is no clear block structure aligned with the defect geometry. This is consistent with the observation in the Transolver paper that attention over raw mesh points tends to be much flatter and closer to a uniform distribution than Physics-Attention over tokens.

Overall, these visualizations support the hypothesis that the adaptive, slice-based receptive field is well suited to S2Q. The learned slices provide an intermediate representation that (i) automatically segments the domain into physically coherent regions, (ii) concentrates model capacity on the thin boundary layers that encode most of the information about subsurface heat sources, and (iii) enables sharp, interpretable token-level attention patterns that capture long-range coupling between surface measurements and interior sources. In contrast, LNO’s point-wise attention struggles to disentangle boundary anomalies from smooth background diffusion, leading to more diffuse score fields and nearly uniform attention. These qualitative differences in physical attention align with the quantitative results in \Cref{tab:s2q_fsplit_test_1,tab:s2q_fsplit_test_2,tab:s2q_fsplit_test_3,tab:s2q_fsplit_test_4,tab:s2q_fsplit_test_5,tab:s2q_fsplit_test_6,tab:s2q_fsplit_noise,tab:s2q_I_double_ood,tab:s2q_I_double_sfo,tab:s2q_I_multi_ood,tab:s2q_II_double_ood,tab:s2q_II_multi_ood,tab:s2q_II_multi_sfo,tab:s2q_III_double_ood,tab:s2q_III_double_sfo,tab:s2q_multi_type_123456,tab:s2q_multi_type_123456_sfo,tab:s2q_single_ood}, where Transolver achieves lower S2Q MSE than LNO in many configurations, with exceptions depending on the grid and held-out defect family.

\section{Limitation}
Although the dataset provides a valuable benchmark for rail nondestructive testing, several boundary conditions exist. First, the data domain is primarily constrained to U71Mn steel specimens with predefined artificial crack morphologies. Geometrically, the narrow variance in the defect width limits the network's ability to resolve this specific spatial feature. Second, restricted by the nature of experimental volumetric sensing, real-world observations only capture surface temperature time-series, rendering the 3D internal states reliant on simulated approximations. Additionally, material properties are treated as constants; while physically sound for a 600 ms pulsed heating process, this restricts the capture of temperature-dependent material variations. Given these factors, the dataset is intended as an algorithmic benchmark rather than a proxy for safety-critical deployment, and its generalization to other material domains requires further calibration.

\section{Broader Impacts}
\label{appendix: Broader Impacts}
This study advances the application of neural operators in non-destructive testing (NDT) using the Aletheia dataset, with the potential to enhance the safety and service life of critical infrastructure such as high-speed rail and aviation. However, when faced with out-of-distribution (OOD) defects or complex noise, model performance degradation may lead to catastrophic false negatives. Therefore, prior to practical deployment, rigorous uncertainty quantification and physical consistency verification must be implemented, and human experts must retain control over decision-making. Furthermore, current benchmarks are primarily calibrated for specific steels; blindly transferring these models to other materials (such as composites) may introduce systematic biases, making it imperative for the research community to establish more rigorous cross-domain evaluation standards in the future.

\section{More Experimental Results}
\label{appendix: More experiments}

\paragraph{Variability across training runs.}
To assess training variability, \Cref{tab:test6_mean_std} reports mean and sample standard deviation over three training runs for the representative test6 geometric-OOD split. Panels (a) and (b) present T2Q and S2Q, respectively, each on regular and irregular grids. These statistics concern a single held-out defect family rather than the six-split macro-average.

\begin{table}[htbp]
  \centering
  \footnotesize
  \setlength{\tabcolsep}{2pt}
  \renewcommand{\arraystretch}{1.15}
  \caption{Performance on the sixth geometric-OOD split (test6), with the Type II multi-layer defect family held out for testing and the other five families used for training. Panels (a) and (b) report T2Q and S2Q, respectively. Entries are mean $\pm$ sample standard deviation over three training runs. MSE and RMSE are lower-is-better; SSIM is higher-is-better.}
  \label{tab:test6_mean_std}
  \begin{tabular*}{\linewidth}{@{\extracolsep{\fill}}lcccccc@{}}
    \toprule
    \multicolumn{7}{c}{\textbf{(a) T2Q}} \\
    \midrule
    Model & \multicolumn{3}{c}{Irregular grid} & \multicolumn{3}{c}{Regular grid} \\
    \cmidrule(lr){2-4}\cmidrule(lr){5-7}
    & MSE $\downarrow$ & RMSE $\downarrow$ & SSIM $\uparrow$ & MSE $\downarrow$ & RMSE $\downarrow$ & SSIM $\uparrow$ \\
    \midrule
    MLP        & 0.213 $\pm$ 0.002 & 0.460 $\pm$ 0.002 & 0.796 $\pm$ 0.008 & 0.126 $\pm$ 0.004 & 0.299 $\pm$ 0.009 & 0.966 $\pm$ 0.006 \\
    FNO        & 0.503 $\pm$ 0.028 & 0.707 $\pm$ 0.019 & 0.555 $\pm$ 0.030 & 0.020 $\pm$ 0.021 & 0.141 $\pm$ 0.022 & 0.948 $\pm$ 0.032 \\
    GeoFNO     & 0.175 $\pm$ 0.004 & 0.413 $\pm$ 0.003 & 0.211 $\pm$ 0.006 & 0.076 $\pm$ 0.019 & 0.248 $\pm$ 0.029 & 0.649 $\pm$ 0.077 \\
    FFNO       & 0.276 $\pm$ 0.004 & 0.523 $\pm$ 0.004 & 0.503 $\pm$ 0.009 & 0.140 $\pm$ 0.004 & 0.324 $\pm$ 0.004 & 0.933 $\pm$ 0.003 \\
    FCNO       & 0.256 $\pm$ 0.002 & 0.504 $\pm$ 0.002 & 0.462 $\pm$ 0.026 & 0.144 $\pm$ 0.002 & 0.338 $\pm$ 0.001 & 0.904 $\pm$ 0.005 \\
    LNO        & 0.408 $\pm$ 0.036 & 0.633 $\pm$ 0.030 & 0.781 $\pm$ 0.005 & 0.341 $\pm$ 0.012 & 0.574 $\pm$ 0.012 & 0.890 $\pm$ 0.002 \\
    Transolver & 0.128 $\pm$ 0.016 & 0.342 $\pm$ 0.021 & 0.866 $\pm$ 0.004 & 0.032 $\pm$ 0.017 & 0.174 $\pm$ 0.021 & 0.959 $\pm$ 0.006 \\
    DeepONet   & 0.218 $\pm$ 0.004 & 0.465 $\pm$ 0.005 & 0.762 $\pm$ 0.008 & 0.147 $\pm$ 0.005 & 0.324 $\pm$ 0.011 & 0.969 $\pm$ 0.018 \\
    LaMO       & 0.113 $\pm$ 0.010 & 0.325 $\pm$ 0.013 & 0.859 $\pm$ 0.010 & 0.033 $\pm$ 0.009 & 0.173 $\pm$ 0.008 & 0.941 $\pm$ 0.002 \\
    \midrule
    \multicolumn{7}{c}{} \\
    \multicolumn{7}{c}{\textbf{(b) S2Q}} \\
    \midrule
    Model & \multicolumn{3}{c}{Irregular grid} & \multicolumn{3}{c}{Regular grid} \\
    \cmidrule(lr){2-4}\cmidrule(lr){5-7}
    & MSE $\downarrow$ & RMSE $\downarrow$ & SSIM $\uparrow$ & MSE $\downarrow$ & RMSE $\downarrow$ & SSIM $\uparrow$ \\
    \midrule
    MLP        & 0.634 $\pm$ 0.001 & 0.795 $\pm$ 0.001 & 0.667 $\pm$ 0.010 & 0.519 $\pm$ 0.008 & 0.719 $\pm$ 0.006 & 0.838 $\pm$ 0.001 \\
    FNO        & 0.886 $\pm$ 0.027 & 0.939 $\pm$ 0.014 & 0.409 $\pm$ 0.042 & 0.232 $\pm$ 0.001 & 0.445 $\pm$ 0.001 & 0.868 $\pm$ 0.026 \\
    GeoFNO     & 0.891 $\pm$ 0.011 & 0.919 $\pm$ 0.009 & 0.437 $\pm$ 0.017 & 0.278 $\pm$ 0.001 & 0.486 $\pm$ 0.001 & 0.885 $\pm$ 0.141 \\
    FFNO       & 0.892 $\pm$ 0.004 & 0.944 $\pm$ 0.002 & 0.438 $\pm$ 0.012 & 0.217 $\pm$ 0.004 & 0.439 $\pm$ 0.005 & 0.880 $\pm$ 0.003 \\
    FCNO       & 0.997 $\pm$ 0.006 & 0.998 $\pm$ 0.003 & 0.170 $\pm$ 0.010 & 0.222 $\pm$ 0.001 & 0.447 $\pm$ 0.000 & 0.856 $\pm$ 0.002 \\
    LNO        & 0.883 $\pm$ 0.059 & 0.910 $\pm$ 0.061 & 0.521 $\pm$ 0.064 & 0.534 $\pm$ 0.008 & 0.692 $\pm$ 0.014 & 0.742 $\pm$ 0.021 \\
    Transolver & 0.285 $\pm$ 0.020 & 0.519 $\pm$ 0.021 & 0.753 $\pm$ 0.009 & 0.240 $\pm$ 0.002 & 0.454 $\pm$ 0.001 & 0.932 $\pm$ 0.006 \\
    DeepONet   & 0.609 $\pm$ 0.022 & 0.778 $\pm$ 0.015 & 0.174 $\pm$ 0.017 & 0.610 $\pm$ 0.020 & 0.780 $\pm$ 0.013 & 0.096 $\pm$ 0.016 \\
    LaMO       & 0.276 $\pm$ 0.017 & 0.512 $\pm$ 0.016 & 0.766 $\pm$ 0.009 & 0.231 $\pm$ 0.004 & 0.445 $\pm$ 0.003 & 0.933 $\pm$ 0.008 \\
    \bottomrule
  \end{tabular*}
\end{table}

On irregular grids, LaMO and Transolver achieve lower mean reconstruction errors than the other baselines in both tasks. Their S2Q MSEs are close relative to the observed run-to-run variation, so the results do not establish a reliable ranking between them. On regular grids, FNO achieves the lowest mean T2Q MSE, whereas FFNO achieves the lowest mean S2Q MSE. Variability differs across models and metrics.

The following tables provide detailed comparisons across Q2T, T2Q, T2T, and S2Q on regular and irregular grids under in-distribution and OOD settings. For in-distribution, frequency-band OOD, and single-frequency OOD experiments, we report MSE and RMSE. For geometric OOD, the detailed per-family tables additionally report SSIM, nRMSE, cRMSE, maximum error, and bRMSE. Higher SSIM indicates better performance, whereas lower values are preferred for the other metrics. Entries denoted by “-” correspond to cases where repeated trainings exhibited highly unstable behavior or consistently failed to produce meaningful results.

\makeatletter
\setlength{\@fptop}{25pt}
\setlength{\@fpsep}{25pt}
\setlength{\@fpbot}{0pt plus 1fil}
\makeatother


\begin{table}[p]
    \scriptsize
    \setlength{\tabcolsep}{3pt}
    \renewcommand{\arraystretch}{1.4}
    \centering
    \caption{Surface-to-Source reconstruction under geometric OOD with the single-layer defect (type 1) held out. Models are trained on the other five defect types and tested on the held-out type, using irregular and regular grids.}
    \label{tab:s2q_fsplit_test_1}
\begin{tabular}{l | ccccccc | ccccccc}
        \toprule
        \multirow{2}{*}[-0.6ex]{Model}
        & \multicolumn{7}{c|}{Irregular Data} & \multicolumn{7}{c}{Regular Data} \\
        \cmidrule{2-15}
        & MSE & SSIM & RMSE & nRMSE & cRMSE & Max & bRMSE
        & MSE & SSIM & RMSE & nRMSE & cRMSE & Max & bRMSE \\
        \midrule
        MLP        & 0.546 & 0.720 & 0.736 & 0.736 & 0.004 & 32.009 & 0.030 & 0.415 & 0.817 & 0.642 & 0.642 & 0.003 & 21.362 & 0.033 \\
        FNO        & 0.469 & 0.605 & 0.678 & 0.678 & 0.016 & 23.876 & 0.027 & 0.069 & 0.918 & 0.239 & 0.239 & 0.005 & 13.647 & 0.014 \\
        GeoFNO     & 0.388 & 0.191 & 0.615 & 0.615 & 0.006 & 20.238 & 0.074 & 0.076 & 0.652 & 0.317 & 0.317 & 0.002 & 11.491 & 0.027 \\
        FFNO       & 0.375 & 0.545 & 0.610 & 0.610 & 0.016 & 23.593 & 0.033 & 0.065 & 0.912 & 0.238 & 0.238 & 0.012 & 12.259 & 0.016 \\
        FCNO       & 0.439 & 0.233 & 0.662 & 0.662 & 0.029 & 24.518 & 0.055 & 0.073 & 0.885 & 0.256 & 0.256 & 0.011 & 12.836 & 0.014 \\
        LNO        & 0.317 & 0.672 & 0.552 & 0.552 & 0.020 & 25.224 & 0.035 & 0.126 & 0.873 & 0.347 & 0.347 & 0.005 & 13.908 & 0.031 \\
        Transolver & 0.255 & 0.770 & 0.484 & 0.484 & 0.007 & 17.889 & 0.024 & 0.110 & 0.951 & 0.303 & 0.303 & 0.003 & 14.516 & 0.012 \\
        DeepONet   & 0.551 & 0.172 & 0.738 & 0.738 & 0.002 & 32.718 & 0.128 & 0.521 & 0.116 & 0.719 & 0.719 & 0.002 & 21.642 & 1.014 \\
        LaMO       & 0.243 & 0.797 & 0.473 & 0.473 & 0.008 & 19.172 & 0.013 & 0.127 & 0.945 & 0.329 & 0.329 & 0.003 & 16.657 & 0.009 \\
        \bottomrule
    \end{tabular}
\end{table}

\begin{table}[p]
    \scriptsize
    \setlength{\tabcolsep}{3pt}
    \renewcommand{\arraystretch}{1.4}
    \centering
    \caption{Surface-to-Source reconstruction under geometric OOD with the Type I double-layer defect (type 2) held out. Models are trained on the other five defect types and tested on the held-out type, using irregular and regular grids.}
    \label{tab:s2q_fsplit_test_2}
    \begin{tabular}{l | ccccccc | ccccccc}
        \toprule
        \multirow{2}{*}[-0.6ex]{Model}
        & \multicolumn{7}{c|}{Irregular Data} & \multicolumn{7}{c}{Regular Data} \\
        \cmidrule{2-15}
        & MSE & SSIM & RMSE & nRMSE & cRMSE & Max & bRMSE
        & MSE & SSIM & RMSE & nRMSE & cRMSE & Max & bRMSE \\
        \midrule
        MLP        & 0.720 & 0.640 & 0.845 & 0.845 & 0.006 & 34.390 & 0.060 & 0.679 & 0.832 & 0.818 & 0.818 & 0.004 & 46.161 & 0.042 \\
        FNO        & 0.669 & 0.597 & 0.806 & 0.806 & 0.007 & 32.483 & 0.055 & 0.466 & 0.890 & 0.655 & 0.655 & 0.006 & 35.192 & 0.015 \\
        GeoFNO     & 0.635 & 0.235 & 0.785 & 0.785 & 0.006 & 32.184 & 0.053 & 0.424 & 0.344 & 0.630 & 0.630 & 0.004 & 35.516 & 0.055 \\
        FFNO       & 0.713 & 0.510 & 0.837 & 0.837 & 0.008 & 32.941 & 0.071 & 0.428 & 0.878 & 0.638 & 0.638 & 0.016 & 37.223 & 0.020 \\
        FCNO       & 0.767 & 0.219 & 0.870 & 0.870 & 0.025 & 34.413 & 0.093 & 0.413 & 0.868 & 0.627 & 0.627 & 0.016 & 36.044 & 0.022 \\
        LNO        & 0.793 & 0.628 & 0.879 & 0.879 & 0.005 & 35.524 & 0.073 & 0.573 & 0.892 & 0.741 & 0.741 & 0.007 & 40.241 & 0.024 \\
        Transolver & 0.694 & 0.703 & 0.819 & 0.819 & 0.004 & 33.344 & 0.054 & 0.446 & 0.921 & 0.648 & 0.648 & 0.003 & 35.502 & 0.011 \\
        DeepONet   & 0.725 & 0.168 & 0.848 & 0.848 & 0.002 & 34.443 & 0.148 & 0.736 & 0.113 & 0.854 & 0.854 & 0.002 & 45.513 & 0.854 \\
        LaMO       & 0.669 & 0.710 & 0.803 & 0.803 & 0.004 & 32.815 & 0.037 & 0.435 & 0.916 & 0.642 & 0.642 & 0.003 & 35.539 & 0.014 \\
        \bottomrule
    \end{tabular}
\end{table}

\begin{table}[p]
    \scriptsize
    \setlength{\tabcolsep}{3pt}
    \renewcommand{\arraystretch}{1.4}
    \centering
    \caption{Surface-to-Source reconstruction under geometric OOD with the Type II double-layer defect (type 3) held out. Models are trained on the other five defect types and tested on the held-out type, using irregular and regular grids.}
    \label{tab:s2q_fsplit_test_3}
    \begin{tabular}{l | ccccccc | ccccccc}
        \toprule
        \multirow{2}{*}[-0.6ex]{Model}
        & \multicolumn{7}{c|}{Irregular Data} & \multicolumn{7}{c}{Regular Data} \\
        \cmidrule{2-15}
        & MSE & SSIM & RMSE & nRMSE & cRMSE & Max & bRMSE
        & MSE & SSIM & RMSE & nRMSE & cRMSE & Max & bRMSE \\
        \midrule
        MLP        & 0.611 & 0.714 & 0.779 & 0.779 & 0.007 & 29.687 & 0.270 & 0.427 & 0.815 & 0.652 & 0.652 & 0.006 & 20.437 & 0.039 \\
        FNO        & 0.809 & 0.419 & 0.897 & 0.897 & 0.026 & 29.815 & 0.297 & 0.118 & 0.908 & 0.311 & 0.311 & 0.004 & 14.576 & 0.029 \\
        GeoFNO     & 0.436 & 0.104 & 0.654 & 0.654 & 0.013 & 25.259 & 0.252 & 0.145 & 0.479 & 0.392 & 0.392 & 0.003 & 14.281 & 0.038 \\
        FFNO       & 0.879 & 0.435 & 0.937 & 0.937 & 0.036 & 31.617 & 0.190 & 0.129 & 0.902 & 0.327 & 0.327 & 0.012 & 14.975 & 0.023 \\
        FCNO       & 0.983 & 0.203 & 0.991 & 0.991 & 0.032 & 32.534 & 0.210 & 0.162 & 0.860 & 0.367 & 0.367 & 0.011 & 15.826 & 0.012 \\
        LNO        & 0.322 & 0.751 & 0.551 & 0.551 & 0.013 & 24.316 & 0.109 & 0.190 & 0.906 & 0.428 & 0.428 & 0.006 & 15.186 & 0.013 \\
        Transolver & 0.293 & 0.810 & 0.511 & 0.511 & 0.007 & 23.298 & 0.079 & 0.152 & 0.942 & 0.354 & 0.354 & 0.003 & 14.726 & 0.012 \\
        DeepONet   & 0.603 & 0.149 & 0.773 & 0.773 & 0.005 & 29.179 & 0.324 & 0.541 & 0.111 & 0.733 & 0.733 & 0.005 & 21.350 & 0.916 \\
        LaMO       & 0.280 & 0.816 & 0.501 & 0.501 & 0.007 & 23.879 & 0.109 & 0.162 & 0.942 & 0.369 & 0.369 & 0.004 & 15.491 & 0.022 \\
        \bottomrule
    \end{tabular}
\end{table}

\begin{table}[p]
    \scriptsize
    \setlength{\tabcolsep}{3pt}
    \renewcommand{\arraystretch}{1.4}
    \centering
    \caption{Surface-to-Source reconstruction under geometric OOD with the Type III double-layer defect (type 4) held out. Models are trained on the other five defect types and tested on the held-out type, using irregular and regular grids.}
    \label{tab:s2q_fsplit_test_4}
    \begin{tabular}{l | ccccccc | ccccccc}
        \toprule
        \multirow{2}{*}[-0.6ex]{Model}
        & \multicolumn{7}{c|}{Irregular Data} & \multicolumn{7}{c}{Regular Data} \\
        \cmidrule{2-15}
        & MSE & SSIM & RMSE & nRMSE & cRMSE & Max & bRMSE
        & MSE & SSIM & RMSE & nRMSE & cRMSE & Max & bRMSE \\
        \midrule
        MLP        & 0.570 & 0.672 & 0.752 & 0.752 & 0.006 & 33.520 & 0.213 & 0.471 & 0.843 & 0.680 & 0.680 & 0.007 & 32.253 & 0.038 \\
        FNO        & 0.880 & 0.358 & 0.934 & 0.934 & 0.021 & 33.318 & 0.287 & 0.229 & 0.879 & 0.465 & 0.465 & 0.005 & 21.281 & 0.014 \\
        GeoFNO     & 0.564 & 0.099 & 0.746 & 0.746 & 0.018 & 30.183 & 0.315 & 0.216 & 0.356 & 0.449 & 0.449 & 0.004 & 22.292 & 0.055 \\
        FFNO       & 0.916 & 0.403 & 0.957 & 0.957 & 0.039 & 35.588 & 0.332 & 0.205 & 0.882 & 0.442 & 0.442 & 0.015 & 21.008 & 0.039 \\
        FCNO       & 0.989 & 0.182 & 0.995 & 0.995 & 0.033 & 36.495 & 0.359 & 0.252 & 0.866 & 0.490 & 0.490 & 0.017 & 25.182 & 0.015 \\
        LNO        & 0.526 & 0.641 & 0.715 & 0.715 & 0.012 & 30.830 & 0.282 & 0.257 & 0.914 & 0.496 & 0.496 & 0.006 & 24.425 & 0.013 \\
        Transolver & 0.470 & 0.679 & 0.673 & 0.673 & 0.008 & 26.708 & 0.256 & 0.284 & 0.933 & 0.509 & 0.509 & 0.006 & 25.494 & 0.011 \\
        DeepONet   & 0.604 & 0.166 & 0.774 & 0.774 & 0.005 & 33.244 & 0.237 & 0.595 & 0.119 & 0.766 & 0.766 & 0.005 & 33.322 & 0.898 \\
        LaMO       & 0.463 & 0.665 & 0.669 & 0.669 & 0.007 & 27.055 & 0.262 & 0.271 & 0.931 & 0.502 & 0.502 & 0.004 & 24.345 & 0.014 \\
        \bottomrule
    \end{tabular}
\end{table}

\begin{table}[p]
    \scriptsize
    \setlength{\tabcolsep}{3pt}
    \renewcommand{\arraystretch}{1.4}
    \centering
    \caption{Surface-to-Source reconstruction under geometric OOD with the Type I multi-layer defect (type 5) held out. Models are trained on the other five defect types and tested on the held-out type, using irregular and regular grids.}
    \label{tab:s2q_fsplit_test_5}
    \begin{tabular}{l | ccccccc | ccccccc}
        \toprule
        \multirow{2}{*}[-0.6ex]{Model}
        & \multicolumn{7}{c|}{Irregular Data} & \multicolumn{7}{c}{Regular Data} \\
        \cmidrule{2-15}
        & MSE & SSIM & RMSE & nRMSE & cRMSE & Max & bRMSE
        & MSE & SSIM & RMSE & nRMSE & cRMSE & Max & bRMSE \\
        \midrule
        MLP        & 0.600 & 0.703 & 0.771 & 0.771 & 0.006 & 31.321 & 0.224 & 0.496 & 0.816 & 0.701 & 0.701 & 0.006 & 23.783 & 0.034 \\
        FNO        & 0.923 & 0.387 & 0.954 & 0.954 & 0.018 & 31.957 & 0.246 & 0.213 & 0.853 & 0.431 & 0.431 & 0.006 & 18.226 & 0.029 \\
        GeoFNO     & 0.483 & 0.091 & 0.687 & 0.687 & 0.011 & 26.892 & 0.257 & 0.190 & 0.234 & 0.411 & 0.411 & 0.010 & 17.033 & 0.084 \\
        FFNO       & 0.892 & 0.415 & 0.943 & 0.943 & 0.036 & 34.986 & 0.329 & 0.197 & 0.876 & 0.422 & 0.422 & 0.011 & 17.387 & 0.033 \\
        FCNO       & 0.978 & 0.198 & 0.989 & 0.989 & 0.033 & 36.036 & 0.253 & 0.217 & 0.832 & 0.448 & 0.448 & 0.010 & 17.300 & 0.018 \\
        LNO        & 0.407 & 0.669 & 0.619 & 0.619 & 0.017 & 25.462 & 0.185 & 0.369 & 0.875 & 0.576 & 0.576 & 0.009 & 20.651 & 0.034 \\
        Transolver & 0.376 & 0.707 & 0.567 & 0.567 & 0.009 & 26.039 & 0.129 & 0.235 & 0.920 & 0.446 & 0.446 & 0.004 & 17.845 & 0.017 \\
        DeepONet   & 0.597 & 0.144 & 0.768 & 0.768 & 0.006 & 30.776 & 0.247 & 0.589 & 0.106 & 0.765 & 0.765 & 0.004 & 24.585 & 1.052 \\
        LaMO       & 0.355 & 0.714 & 0.546 & 0.546 & 0.005 & 25.759 & 0.152 & 0.215 & 0.923 & 0.432 & 0.432 & 0.004 & 17.134 & 0.017 \\
        \bottomrule
    \end{tabular}
\end{table}

\begin{table}[p]
    \scriptsize
    \setlength{\tabcolsep}{3pt}
    \renewcommand{\arraystretch}{1.4}
    \centering
    \caption{Surface-to-Source reconstruction under geometric OOD with the Type II multi-layer defect (type 6) held out. Models are trained on the other five defect types and tested on the held-out type, using irregular and regular grids.}
    \label{tab:s2q_fsplit_test_6}
    \begin{tabular}{l | ccccccc | ccccccc}
        \toprule
        \multirow{2}{*}[-0.6ex]{Model}
        & \multicolumn{7}{c|}{Irregular Data} & \multicolumn{7}{c}{Regular Data} \\
        \cmidrule{2-15}
        & MSE & SSIM & RMSE & nRMSE & cRMSE & Max & bRMSE
        & MSE & SSIM & RMSE & nRMSE & cRMSE & Max & bRMSE \\
        \midrule
        MLP        & 0.634 & 0.667 & 0.795 & 0.795 & 0.006 & 28.223 & 0.197 & 0.519 & 0.838 & 0.719 & 0.719 & 0.005 & 27.901 & 0.021 \\
        FNO        & 0.886 & 0.409 & 0.939 & 0.939 & 0.036 & 29.004 & 0.250 & 0.232 & 0.868 & 0.445 & 0.445 & 0.003 & 26.311 & 0.024 \\
        GeoFNO     & 0.891 & 0.437 & 0.919 & 0.919 & 0.040 & 29.464 & 0.255 & 0.278 & 0.885 & 0.486 & 0.486 & 0.006 & 27.841 & 0.029 \\
        FFNO       & 0.892 & 0.438 & 0.944 & 0.944 & 0.039 & 29.581 & 0.209 & 0.217 & 0.880 & 0.439 & 0.439 & 0.012 & 25.842 & 0.035 \\
        FCNO       & 0.997 & 0.170 & 0.998 & 0.998 & 0.041 & 30.558 & 0.143 & 0.222 & 0.856 & 0.447 & 0.447 & 0.012 & 26.155 & 0.017 \\
        LNO        & 0.883 & 0.521 & 0.910 & 0.910 & 0.037 & 30.969 & 0.175 & 0.534 & 0.742 & 0.692 & 0.692 & 0.009 & 30.351 & 0.032 \\
        Transolver & 0.285 & 0.753 & 0.519 & 0.519 & 0.008 & 23.460 & 0.126 & 0.240 & 0.932 & 0.454 & 0.454 & 0.003 & 26.334 & 0.016 \\
        DeepONet   & 0.609 & 0.174 & 0.778 & 0.778 & 0.006 & 27.534 & 0.262 & 0.610 & 0.096 & 0.780 & 0.780 & 0.006 & 27.682 & 0.786 \\
        LaMO       & 0.276 & 0.766 & 0.512 & 0.512 & 0.007 & 23.563 & 0.096 & 0.231 & 0.933 & 0.445 & 0.445 & 0.002 & 25.937 & 0.014 \\
        \bottomrule
    \end{tabular}
\end{table}

\begin{table}[p]
    \scriptsize
    \setlength{\tabcolsep}{2.5pt}
    \renewcommand{\arraystretch}{1.4}
    \centering
    \caption{Surface-to-Source reconstruction performance (MSE / RMSE) for the all six defect types defect under normal training and three frequency-band OOD regimes, on irregular and regular grids.}
    \label{tab:s2q_multi_type_123456}
    \begin{tabular}{l | cc | cc | cc | cc | cc | cc | cc | cc}
        \toprule
         &
        \multicolumn{8}{c|}{Irregular Data} &
        \multicolumn{8}{c}{Regular Data} \\
        \cmidrule{2-17}
        \multicolumn{1}{c|}{Model}  &
        \multicolumn{2}{c|}{Normal} &
        \multicolumn{2}{c|}{High-OOD} &
        \multicolumn{2}{c|}{Mid-OOD} &
        \multicolumn{2}{c|}{Low-OOD} &
        \multicolumn{2}{c|}{Normal} &
        \multicolumn{2}{c|}{High-OOD} &
        \multicolumn{2}{c|}{Mid-OOD} &
        \multicolumn{2}{c}{Low-OOD} \\
        \cmidrule{2-17}
        & MSE & RMSE & MSE & RMSE & MSE & RMSE & MSE & RMSE & MSE & RMSE & MSE & RMSE & MSE & RMSE & MSE & RMSE \\
        \midrule
        MLP        & 0.510 & 0.719 & 0.702 & 0.835 & 0.636 & 0.795 & 0.610 & 0.774 & 0.484 & 0.691 & 0.578 & 0.754 & 0.489 & 0.698 & 0.486 & 0.703 \\
        FNO        & 0.292 & 0.478 & 0.441 & 0.605 & 0.360 & 0.544 & 0.471 & 0.656 & 0.092 & 0.270 & 0.123 & 0.310 & 0.105 & 0.308 & 0.129 & 0.336 \\
        GeoFNO     & 0.309 & 0.530 & 0.383 & 0.575 & 0.316 & 0.515 & 0.356 & 0.583 & 0.093 & 0.277 & 0.110 & 0.291 & 0.163 & 0.415 & 0.167 & 0.443 \\
        FFNO       & 0.479 & 0.620 & 0.553 & 0.712 & 0.516 & 0.682 & 0.615 & 0.764 & 0.097 & 0.284 & 0.140 & 0.348 & 0.108 & 0.315 & 0.123 & 0.343 \\
        FCNO       & 0.653 & 0.733 & 0.727 & 0.826 & 0.698 & 0.800 & 0.771 & 0.862 & 0.112 & 0.308 & 0.160 & 0.377 & 0.115 & 0.310 & 0.140 & 0.368 \\
\bottomrule
    \end{tabular}
\end{table}

\begin{table}[p]
    \scriptsize
    \setlength{\tabcolsep}{2.2pt}
    \renewcommand{\arraystretch}{1.4}
    \centering
    \caption{Surface-to-Source reconstruction performance (MSE / RMSE) for the all six defect types defect under single-frequency OOD (SFO). Models are trained on all frequencies (Normal) or on a single excitation frequency, evaluated on irregular and regular grids.}
    \label{tab:s2q_multi_type_123456_sfo}
    \begin{tabular}{l | cc | cc | cc | cc | cc | cc | cc | cc}
        \toprule
         &
        \multicolumn{8}{c|}{Irregular Data} &
        \multicolumn{8}{c}{Regular Data} \\
        \cmidrule{2-17}
        \multicolumn{1}{c|}{Model}  &
        \multicolumn{2}{c|}{Normal} &
        \multicolumn{2}{c|}{SFO-9kHz} &
        \multicolumn{2}{c|}{SFO-36kHz} &
        \multicolumn{2}{c|}{SFO-81kHz} &
        \multicolumn{2}{c|}{Normal} &
        \multicolumn{2}{c|}{SFO-9kHz} &
        \multicolumn{2}{c|}{SFO-36kHz} &
        \multicolumn{2}{c}{SFO-81kHz} \\
        \cmidrule{2-17}
        & MSE & RMSE & MSE & RMSE & MSE & RMSE & MSE & RMSE & MSE & RMSE & MSE & RMSE & MSE & RMSE & MSE & RMSE \\
        \midrule
        MLP        & 0.510 & 0.719 & 0.612 & 0.779 & 0.623 & 0.787 & 0.652 & 0.807 & 0.484 & 0.691 & 0.510 & 0.708 & 0.525 & 0.721 & 0.577 & 0.758 \\
        FNO        & 0.292 & 0.478 & 0.869 & 0.890 & 0.595 & 0.736 & 0.684 & 0.782 & 0.092 & 0.270 & 0.368 & 0.558 & 0.234 & 0.445 & 0.305 & 0.505 \\
        GeoFNO     & 0.309 & 0.530 & 0.385 & 0.606 & 0.423 & 0.627 & 0.380 & 0.580 & 0.093 & 0.277 & 0.149 & 0.351 & 0.216 & 0.447 & 0.126 & 0.313 \\
        FFNO       & 0.479 & 0.620 & 0.563 & 0.743 & 0.649 & 0.770 & 0.631 & 0.785 & 0.097 & 0.284 & 0.229 & 0.449 & 0.198 & 0.421 & 0.242 & 0.468 \\
        FCNO       & 0.653 & 0.733 & 0.759 & 0.863 & 0.746 & 0.853 & 0.794 & 0.883 & 0.112 & 0.308 & 0.230 & 0.457 & 0.214 & 0.441 & 0.298 & 0.523 \\
\bottomrule
    \end{tabular}
\end{table}

\begin{table}[p]
    \scriptsize
    \setlength{\tabcolsep}{2.5pt}
    \renewcommand{\arraystretch}{1.4}
    \centering
    \caption{Surface-to-Source reconstruction performance (MSE / RMSE) for the single layer defect under normal training and three frequency-band OOD regimes, on irregular and regular grids.}
    \label{tab:s2q_single_ood}
    \begin{tabular}{l | cc | cc | cc | cc | cc | cc | cc | cc}
        \toprule
         &
        \multicolumn{8}{c|}{Irregular Data} &
        \multicolumn{8}{c}{Regular Data} \\
        \cmidrule{2-17}
        \multicolumn{1}{c|}{Model}  &
        \multicolumn{2}{c|}{Normal} &
        \multicolumn{2}{c|}{High-OOD} &
        \multicolumn{2}{c|}{Mid-OOD} &
        \multicolumn{2}{c|}{Low-OOD} &
        \multicolumn{2}{c|}{Normal} &
        \multicolumn{2}{c|}{High-OOD} &
        \multicolumn{2}{c|}{Mid-OOD} &
        \multicolumn{2}{c}{Low-OOD} \\
        \cmidrule{2-17}
        & MSE & RMSE & MSE & RMSE & MSE & RMSE & MSE & RMSE & MSE & RMSE & MSE & RMSE & MSE & RMSE & MSE & RMSE \\
        \midrule
        MLP        & 0.502 & 0.705 & 0.625 & 0.786 & 0.552 & 0.741 & 0.568 & 0.754 & 0.368 & 0.604 & 0.455 & 0.671 & 0.366 & 0.603 & 0.465 & 0.681 \\
        FNO        & 0.063 & 0.241 & 0.096 & 0.294 & 0.064 & 0.247 & 0.166 & 0.387 & 0.024 & 0.149 & 0.055 & 0.210 & 0.026 & 0.152 & 0.075 & 0.260 \\
        GeoFNO     & 0.062 & 0.240 & 0.094 & 0.295 & 0.065 & 0.252 & 0.178 & 0.392 & 0.030 & 0.150 & 0.056 & 0.212 & 0.029 & 0.159 & 0.060 & 0.209 \\
        FFNO       & 0.072 & 0.261 & 0.105 & 0.315 & 0.073 & 0.267 & 0.111 & 0.327 & 0.021 & 0.139 & 0.065 & 0.233 & 0.023 & 0.145 & 0.080 & 0.274 \\
        FCNO       & 0.078 & 0.272 & 0.115 & 0.330 & 0.078 & 0.274 & 0.138 & 0.366 & 0.019 & 0.143 & 0.060 & 0.229 & 0.025 & 0.152 & 0.083 & 0.280 \\
\bottomrule
    \end{tabular}
\end{table}

\begin{table}[p]
    \scriptsize
    \setlength{\tabcolsep}{2.5pt}
    \renewcommand{\arraystretch}{1.4}
    \centering
    \caption{Surface-to-Source reconstruction performance (MSE / RMSE) for the Type I double-layer defect under normal training and three frequency-band OOD regimes, on irregular and regular grids.}
    \label{tab:s2q_I_double_ood}
    \begin{tabular}{l | cc | cc | cc | cc | cc | cc | cc | cc}
        \toprule
         &
        \multicolumn{8}{c|}{Irregular Data} &
        \multicolumn{8}{c}{Regular Data} \\
        \cmidrule{2-17}
        \multicolumn{1}{c|}{Model}  &
        \multicolumn{2}{c|}{Normal} &
        \multicolumn{2}{c|}{High-OOD} &
        \multicolumn{2}{c|}{Mid-OOD} &
        \multicolumn{2}{c|}{Low-OOD} &
        \multicolumn{2}{c|}{Normal} &
        \multicolumn{2}{c|}{High-OOD} &
        \multicolumn{2}{c|}{Mid-OOD} &
        \multicolumn{2}{c}{Low-OOD} \\
        \cmidrule{2-17}
        & MSE & RMSE & MSE & RMSE & MSE & RMSE & MSE & RMSE & MSE & RMSE & MSE & RMSE & MSE & RMSE & MSE & RMSE \\
        \midrule
        MLP        & 0.691 & 0.829 & 0.774 & 0.878 & 0.708 & 0.840 & 0.707 & 0.839 & 0.628 & 0.787 & 0.718 & 0.844 & 0.678 & 0.822 & 0.667 & 0.816 \\
        FNO        & 0.319 & 0.551 & 0.473 & 0.676 & 0.385 & 0.609 & 0.413 & 0.633 & 0.201 & 0.418 & 0.298 & 0.517 & 0.214 & 0.422 & 0.415 & 0.616 \\
        GeoFNO     & 0.299 & 0.535 & 0.453 & 0.662 & 0.442 & 0.674 & 0.406 & 0.647 & 0.188 & 0.402 & 0.285 & 0.500 & 0.269 & 0.493 & 0.283 & 0.515 \\
        FFNO       & 0.349 & 0.579 & 0.467 & 0.676 & 0.387 & 0.615 & 0.373 & 0.606 & 0.184 & 0.397 & 0.275 & 0.492 & 0.198 & 0.423 & 0.243 & 0.481 \\
        FCNO       & 0.367 & 0.595 & 0.486 & 0.690 & 0.393 & 0.620 & 0.427 & 0.647 & 0.189 & 0.405 & 0.282 & 0.509 & 0.191 & 0.416 & 0.257 & 0.496 \\
\bottomrule
    \end{tabular}
\end{table}

\begin{table}[p]
    \scriptsize
    \setlength{\tabcolsep}{2.2pt}
    \renewcommand{\arraystretch}{1.4}
    \centering
    \caption{Surface-to-Source reconstruction performance (MSE / RMSE) for the Type I double-layer defect under single-frequency OOD (SFO). Models are trained on all frequencies (Normal) or on a single excitation frequency, evaluated on irregular and regular grids.}
    \label{tab:s2q_I_double_sfo}
    \begin{tabular}{l | cc | cc | cc | cc | cc | cc | cc | cc}
        \toprule
         &
        \multicolumn{8}{c|}{Irregular Data} &
        \multicolumn{8}{c}{Regular Data} \\
        \cmidrule{2-17}
        \multicolumn{1}{c|}{Model}  &
        \multicolumn{2}{c|}{Normal} &
        \multicolumn{2}{c|}{SFO-9kHz} &
        \multicolumn{2}{c|}{SFO-36kHz} &
        \multicolumn{2}{c|}{SFO-81kHz} &
        \multicolumn{2}{c|}{Normal} &
        \multicolumn{2}{c|}{SFO-9kHz} &
        \multicolumn{2}{c|}{SFO-36kHz} &
        \multicolumn{2}{c}{SFO-81kHz} \\
        \cmidrule{2-17}
        & MSE & RMSE & MSE & RMSE & MSE & RMSE & MSE & RMSE & MSE & RMSE & MSE & RMSE & MSE & RMSE & MSE & RMSE \\
        \midrule
        MLP        & 0.691 & 0.829 & 0.688 & 0.826 & 0.708 & 0.841 & 0.756 & 0.869 & 0.628 & 0.787 & 0.671 & 0.817 & 0.702 & 0.837 & 0.732 & 0.855 \\
        FNO        & 0.319 & 0.551 & 0.850 & 0.891 & 0.882 & 0.909 & 0.942 & 0.944 & 0.201 & 0.418 & 0.838 & 0.849 & 0.577 & 0.712 & 0.655 & 0.766 \\
        GeoFNO     & 0.299 & 0.535 & 0.547 & 0.722 & 0.470 & 0.670 & 0.612 & 0.760 & 0.188 & 0.402 & 0.460 & 0.669 & 0.414 & 0.636 & 0.539 & 0.693 \\
        FFNO       & 0.349 & 0.579 & 0.485 & 0.686 & 0.496 & 0.694 & 0.577 & 0.749 & 0.184 & 0.397 & 0.427 & 0.619 & 0.384 & 0.588 & 0.375 & 0.589 \\
        FCNO       & 0.367 & 0.595 & 0.492 & 0.693 & 0.504 & 0.700 & 0.568 & 0.745 & 0.189 & 0.405 & 0.359 & 0.584 & 0.354 & 0.567 & 0.366 & 0.577 \\
\bottomrule
    \end{tabular}
\end{table}

\begin{table}[p]
    \scriptsize
    \setlength{\tabcolsep}{2.5pt}
    \renewcommand{\arraystretch}{1.4}
    \centering
    \caption{Surface-to-Source reconstruction performance (MSE / RMSE) for the Type II double-layer defect under normal training and three frequency-band OOD regimes, on irregular and regular grids.}
    \label{tab:s2q_II_double_ood}
    \begin{tabular}{l | cc | cc | cc | cc | cc | cc | cc | cc}
        \toprule
         &
        \multicolumn{8}{c|}{Irregular Data} &
        \multicolumn{8}{c}{Regular Data} \\
        \cmidrule{2-17}
        \multicolumn{1}{c|}{Model}  &
        \multicolumn{2}{c|}{Normal} &
        \multicolumn{2}{c|}{High-OOD} &
        \multicolumn{2}{c|}{Mid-OOD} &
        \multicolumn{2}{c|}{Low-OOD} &
        \multicolumn{2}{c|}{Normal} &
        \multicolumn{2}{c|}{High-OOD} &
        \multicolumn{2}{c|}{Mid-OOD} &
        \multicolumn{2}{c}{Low-OOD} \\
        \cmidrule{2-17}
        & MSE & RMSE & MSE & RMSE & MSE & RMSE & MSE & RMSE & MSE & RMSE & MSE & RMSE & MSE & RMSE & MSE & RMSE \\
        \midrule
        MLP        & 0.490 & 0.700 & 0.724 & 0.849 & 0.576 & 0.639 & 0.756 & 0.799 & 0.406 & 0.636 & 0.491 & 0.699 & 0.400 & 0.632 & 0.495 & 0.703 \\
        FNO        & 0.627 & 0.791 & 0.720 & 0.841 & 0.699 & 0.834 & 0.661 & 0.809 & 0.031 & 0.166 & 0.063 & 0.246 & 0.032 & 0.178 & 0.047 & 0.209 \\
        GeoFNO     & 0.660 & 0.794 & 0.671 & 0.783 & 0.794 & 0.825 & 0.709 & 0.751 & 0.032 & 0.166 & 0.063 & 0.248 & 0.032 & 0.179 & 0.033 & 0.175 \\
        FFNO       & 0.867 & 0.930 & 0.890 & 0.942 & 0.866 & 0.929 & 0.901 & 0.949 & 0.029 & 0.163 & 0.066 & 0.254 & 0.030 & 0.173 & 0.073 & 0.263 \\
        FCNO       & 0.976 & 0.988 & 0.981 & 0.990 & 0.970 & 0.984 & 0.985 & 0.992 & 0.034 & 0.180 & 0.078 & 0.276 & 0.037 & 0.191 & 0.103 & 0.314 \\
\bottomrule
    \end{tabular}
\end{table}

\begin{table}[p]
    \scriptsize
    \setlength{\tabcolsep}{2.5pt}
    \renewcommand{\arraystretch}{1.4}
    \centering
    \caption{Surface-to-Source reconstruction performance (MSE / RMSE) for the Type III double-layer defect under normal training and three frequency-band OOD regimes, on irregular and regular grids.}
    \label{tab:s2q_III_double_ood}
    \begin{tabular}{l | cc | cc | cc | cc | cc | cc | cc | cc}
        \toprule
         &
        \multicolumn{8}{c|}{Irregular Data} &
        \multicolumn{8}{c}{Regular Data} \\
        \cmidrule{2-17}
        \multicolumn{1}{c|}{Model}  &
        \multicolumn{2}{c|}{Normal} &
        \multicolumn{2}{c|}{High-OOD} &
        \multicolumn{2}{c|}{Mid-OOD} &
        \multicolumn{2}{c|}{Low-OOD} &
        \multicolumn{2}{c|}{Normal} &
        \multicolumn{2}{c|}{High-OOD} &
        \multicolumn{2}{c|}{Mid-OOD} &
        \multicolumn{2}{c}{Low-OOD} \\
        \cmidrule{2-17}
        & MSE & RMSE & MSE & RMSE & MSE & RMSE & MSE & RMSE & MSE & RMSE & MSE & RMSE & MSE & RMSE & MSE & RMSE \\
        \midrule
        MLP        & 0.463 & 0.680 & 0.641 & 0.796 & 0.556 & 0.743 & 0.554 & 0.740 & 0.490 & 0.696 & 0.584 & 0.759 & 0.575 & 0.787 & 0.517 & 0.717 \\
        FNO        & 0.384 & 0.614 & 0.421 & 0.638 & 0.392 & 0.618 & 0.471 & 0.681 & 0.111 & 0.324 & 0.145 & 0.369 & 0.114 & 0.332 & 0.406 & 0.596 \\
        GeoFNO     & 0.210 & 0.451 & 0.239 & 0.479 & 0.214 & 0.454 & 0.275 & 0.517 & 0.105 & 0.313 & 0.136 & 0.355 & 0.111 & 0.324 & 0.162 & 0.390 \\
        FFNO       & 0.648 & 0.802 & 0.653 & 0.803 & 0.672 & 0.823 & 0.744 & 0.862 & 0.091 & 0.291 & 0.119 & 0.334 & 0.094 & 0.297 & 0.184 & 0.416 \\
        FCNO       & 0.881 & 0.937 & 0.901 & 0.947 & 0.878 & 0.935 & 0.943 & 0.971 & 0.101 & 0.308 & 0.130 & 0.350 & 0.103 & 0.312 & 0.230 & 0.464 \\
\bottomrule
    \end{tabular}
\end{table}

\begin{table}[p]
    \scriptsize
    \setlength{\tabcolsep}{2.2pt}
    \renewcommand{\arraystretch}{1.4}
    \centering
    \caption{Surface-to-Source reconstruction performance (MSE / RMSE) for the Type III double-layer defect under single-frequency OOD (SFO). Models are trained on all frequencies (Normal) or on a single excitation frequency, evaluated on irregular and regular grids.}
    \label{tab:s2q_III_double_sfo}
    \begin{tabular}{l | cc | cc | cc | cc | cc | cc | cc | cc}
        \toprule
         &
        \multicolumn{8}{c|}{Irregular Data} &
        \multicolumn{8}{c}{Regular Data} \\
        \cmidrule{2-17}
        \multicolumn{1}{c|}{Model}  &
        \multicolumn{2}{c|}{Normal} &
        \multicolumn{2}{c|}{SFO-9kHz} &
        \multicolumn{2}{c|}{SFO-36kHz} &
        \multicolumn{2}{c|}{SFO-81kHz} &
        \multicolumn{2}{c|}{Normal} &
        \multicolumn{2}{c|}{SFO-9kHz} &
        \multicolumn{2}{c|}{SFO-36kHz} &
        \multicolumn{2}{c}{SFO-81kHz} \\
        \cmidrule{2-17}
        & MSE & RMSE & MSE & RMSE & MSE & RMSE & MSE & RMSE & MSE & RMSE & MSE & RMSE & MSE & RMSE & MSE & RMSE \\
        \midrule
        MLP        & 0.463 & 0.680 & 0.554 & 0.740 & 0.547 & 0.737 & 0.573 & 0.756 & 0.490 & 0.696 & 0.494 & 0.701 & 0.497 & 0.702 & 0.584 & 0.760 \\
        FNO        & 0.384 & 0.614 & 0.663 & 0.796 & 0.463 & 0.670 & 0.496 & 0.693 & 0.111 & 0.324 & 0.543 & 0.696 & 0.290 & 0.491 & 0.292 & 0.506 \\
        GeoFNO     & 0.210 & 0.451 & 0.681 & 0.823 & 0.467 & 0.672 & 0.415 & 0.643 & 0.105 & 0.313 & 0.232 & 0.470 & 0.185 & 0.409 & 0.246 & 0.459 \\
        FFNO       & 0.648 & 0.802 & 0.653 & 0.818 & 0.696 & 0.892 & 0.706 & 0.836 & 0.091 & 0.291 & 0.192 & 0.422 & 0.159 & 0.382 & 0.186 & 0.412 \\
        FCNO       & 0.881 & 0.937 & 0.901 & 0.947 & 0.894 & 0.945 & 0.910 & 0.952 & 0.101 & 0.308 & 0.217 & 0.450 & 0.183 & 0.406 & 0.214 & 0.438 \\
\bottomrule
    \end{tabular}
\end{table}

\begin{table}[p]
    \scriptsize
    \setlength{\tabcolsep}{2.5pt}
    \renewcommand{\arraystretch}{1.4}
    \centering
    \caption{Surface-to-Source reconstruction performance (MSE / RMSE) for the Type I multi-layer defect under normal training and three frequency-band OOD regimes, on irregular and regular grids.}
    \label{tab:s2q_I_multi_ood}
    \begin{tabular}{l | cc | cc | cc | cc | cc | cc | cc | cc}
        \toprule
         &
        \multicolumn{8}{c|}{Irregular Data} &
        \multicolumn{8}{c}{Regular Data} \\
        \cmidrule{2-17}
        \multicolumn{1}{c|}{Model}  &
        \multicolumn{2}{c|}{Normal} &
        \multicolumn{2}{c|}{High-OOD} &
        \multicolumn{2}{c|}{Mid-OOD} &
        \multicolumn{2}{c|}{Low-OOD} &
        \multicolumn{2}{c|}{Normal} &
        \multicolumn{2}{c|}{High-OOD} &
        \multicolumn{2}{c|}{Mid-OOD} &
        \multicolumn{2}{c}{Low-OOD} \\
        \cmidrule{2-17}
        & MSE & RMSE & MSE & RMSE & MSE & RMSE & MSE & RMSE & MSE & RMSE & MSE & RMSE & MSE & RMSE & MSE & RMSE \\
        \midrule
        MLP        & 0.475 & 0.688 & 0.552 & 0.739 & 0.709 & 0.842 & 0.623 & 0.789 & 0.428 & 0.650 & 0.559 & 0.746 & 0.435 & 0.658 & 0.425 & 0.652 \\
        FNO        & 0.725 & 0.850 & 0.791 & 0.888 & 0.755 & 0.868 & 0.652 & 0.806 & 0.057 & 0.233 & 0.111 & 0.304 & 0.192 & 0.419 & 0.115 & 0.325 \\
        GeoFNO     & 0.468 & 0.678 & 0.580 & 0.761 & 0.493 & 0.701 & 0.584 & 0.714 & 0.114 & 0.307 & 0.194 & 0.425 & 0.117 & 0.329 & 0.241 & 0.494 \\
        FFNO       & 0.915 & 0.956 & 0.967 & 0.983 & 0.918 & 0.957 & 0.911 & 0.954 & 0.087 & 0.273 & 0.156 & 0.377 & 0.089 & 0.284 & 0.092 & 0.298 \\
        FCNO       & 1.009 & 1.014 & 1.018 & 1.019 & 1.016 & 1.018 & 0.999 & 1.000 & 0.108 & 0.306 & 0.179 & 0.404 & 0.112 & 0.318 & 0.137 & 0.362 \\
\bottomrule
    \end{tabular}
\end{table}

\begin{table}[p]
    \scriptsize
    \setlength{\tabcolsep}{2.5pt}
    \renewcommand{\arraystretch}{1.4}
    \centering
    \caption{Surface-to-Source reconstruction performance (MSE / RMSE) for the Type II multi-layer defect under normal training and three frequency-band OOD regimes, on irregular and regular grids.}
    \label{tab:s2q_II_multi_ood}
    \begin{tabular}{l | cc | cc | cc | cc | cc | cc | cc | cc}
        \toprule
         &
        \multicolumn{8}{c|}{Irregular Data} &
        \multicolumn{8}{c}{Regular Data} \\
        \cmidrule{2-17}
        \multicolumn{1}{c|}{Model}  &
        \multicolumn{2}{c|}{Normal} &
        \multicolumn{2}{c|}{High-OOD} &
        \multicolumn{2}{c|}{Mid-OOD} &
        \multicolumn{2}{c|}{Low-OOD} &
        \multicolumn{2}{c|}{Normal} &
        \multicolumn{2}{c|}{High-OOD} &
        \multicolumn{2}{c|}{Mid-OOD} &
        \multicolumn{2}{c}{Low-OOD} \\
        \cmidrule{2-17}
        MSE & RMSE & MSE & RMSE & MSE & RMSE & MSE & RMSE & MSE & RMSE & MSE & RMSE & MSE & RMSE & MSE & RMSE \\
        \midrule
        MLP        & 0.456 & 0.675 & 0.530 & 0.725 & 0.614 & 0.782 & 0.551 & 0.741 & 0.456 & 0.674 & 0.582 & 0.761 & 0.454 & 0.673 & 0.485 & 0.696 \\
        FNO        & 0.188 & 0.423 & 0.218 & 0.456 & 0.195 & 0.430 & 0.243 & 0.485 & 0.021 & 0.132 & 0.050 & 0.214 & 0.029 & 0.154 & 0.061 & 0.234 \\
        GeoFNO     & 0.177 & 0.363 & 0.208 & 0.422 & 0.177 & 0.371 & 0.198 & 0.402 & 0.021 & 0.134 & 0.047 & 0.207 & 0.029 & 0.155 & 0.045 & 0.197 \\
        FFNO       & 0.460 & 0.672 & 0.478 & 0.685 & 0.510 & 0.708 & 0.635 & 0.795 & 0.026 & 0.149 & 0.061 & 0.238 & 0.035 & 0.174 & 0.065 & 0.246 \\
        FCNO       & 0.768 & 0.874 & 0.783 & 0.883 & 0.795 & 0.889 & 0.873 & 0.934 & 0.022 & 0.135 & 0.066 & 0.248 & 0.031 & 0.170 & 0.077 & 0.268 \\
\bottomrule
    \end{tabular}
\end{table}

\begin{table}[p]
    \scriptsize
    \setlength{\tabcolsep}{2.2pt}
    \renewcommand{\arraystretch}{1.4}
    \centering
    \caption{Surface-to-Source reconstruction performance (MSE / RMSE) for the Type II multi-layer defect under single-frequency OOD (SFO). Models are trained on all frequencies (Normal) or on a single excitation frequency, evaluated on irregular and regular grids.}
    \label{tab:s2q_II_multi_sfo}
    \begin{tabular}{l | cc | cc | cc | cc | cc | cc | cc | cc}
        \toprule
         &
        \multicolumn{8}{c|}{Irregular Data} &
        \multicolumn{8}{c}{Regular Data} \\
        \cmidrule{2-17}
        \multicolumn{1}{c|}{Model}  &
        \multicolumn{2}{c|}{Normal} &
        \multicolumn{2}{c|}{SFO-9kHz} &
        \multicolumn{2}{c|}{SFO-36kHz} &
        \multicolumn{2}{c|}{SFO-81kHz} &
        \multicolumn{2}{c|}{Normal} &
        \multicolumn{2}{c|}{SFO-9kHz} &
        \multicolumn{2}{c|}{SFO-36kHz} &
        \multicolumn{2}{c}{SFO-81kHz} \\
        \cmidrule{2-17}
        & MSE & RMSE & MSE & RMSE & MSE & RMSE & MSE & RMSE & MSE & RMSE & MSE & RMSE & MSE & RMSE & MSE & RMSE \\
        \midrule
        MLP        & 0.456 & 0.675 & 0.530 & 0.725 & 0.528 & 0.726 & 0.578 & 0.760 & 0.456 & 0.674 & 0.467 & 0.675 & 0.468 & 0.682 & 0.541 & 0.733 \\
        FNO        & 0.188 & 0.423 & 0.512 & 0.680 & 0.370 & 0.584 & 0.456 & 0.643 & 0.021 & 0.132 & 0.286 & 0.493 & 0.235 & 0.431 & 0.307 & 0.496 \\
        GeoFNO     & 0.177 & 0.363 & 0.405 & 0.531 & 0.469 & 0.689 & 0.340 & 0.558 & 0.021 & 0.134 & 0.172 & 0.384 & 0.219 & 0.407 & 0.273 & 0.467 \\
        FFNO       & 0.460 & 0.672 & 0.559 & 0.739 & 0.521 & 0.710 & 0.613 & 0.771 & 0.026 & 0.149 & 0.175 & 0.389 & 0.162 & 0.371 & 0.221 & 0.425 \\
        FCNO       & 0.768 & 0.874 & 0.834 & 0.911 & 0.796 & 0.889 & 0.827 & 0.906 & 0.022 & 0.135 & 0.185 & 0.405 & 0.166 & 0.382 & 0.223 & 0.427 \\
\bottomrule
    \end{tabular}
\end{table}

\clearpage


\begin{table}[p]
    \scriptsize
    \setlength{\tabcolsep}{3pt}
    \renewcommand{\arraystretch}{1.4}
    \centering
    \caption{Temperature-to-Source reconstruction under geometric OOD with the single-layer defect (type 1) held out. Models are trained on the other five defect types and tested on the held-out type, using irregular and regular grids.}
    \label{tab:t2q_fsplit_test_1}
    \begin{tabular}{l | ccccccc | ccccccc}
        \toprule
        \multirow{2}{*}[-0.6ex]{Model}
        & \multicolumn{7}{c|}{Irregular Data} & \multicolumn{7}{c}{Regular Data} \\
        \cmidrule{2-15}
        & MSE & SSIM & RMSE & nRMSE & cRMSE & Max & bRMSE
        & MSE & SSIM & RMSE & nRMSE & cRMSE & Max & bRMSE \\
        \midrule
        MLP        & 0.174 & 0.844 & 0.412 & 0.412 & 0.010 & 11.890 & 0.018 & 0.020 & 0.966 & 0.137 & 0.137 & 0.008 & 5.142 & 0.007 \\
        FNO        & 0.224 & 0.745 & 0.471 & 0.471 & 0.026 & 16.085 & 0.028 & 0.011 & 0.958 & 0.098 & 0.098 & 0.002 & 4.489 & 0.007 \\
        GeoFNO     & 0.121 & 0.396 & 0.344 & 0.344 & 0.004 & 10.996 & 0.023 & 0.012 & 0.769 & 0.107 & 0.107 & 0.002 & 4.981 & 0.009 \\
        FFNO       & 0.213 & 0.628 & 0.460 & 0.460 & 0.048 & 11.894 & 0.021 & 0.009 & 0.946 & 0.091 & 0.091 & 0.011 & 3.876 & 0.012 \\
        FCNO       & 0.192 & 0.531 & 0.436 & 0.436 & 0.044 & 11.655 & 0.026 & 0.010 & 0.926 & 0.095 & 0.095 & 0.010 & 3.948 & 0.013 \\
        LNO        & 0.217 & 0.784 & 0.458 & 0.458 & 0.004 & 19.311 & 0.024 & 0.110 & 0.866 & 0.325 & 0.325 & 0.009 & 12.725 & 0.026 \\
        Transolver & 0.076 & 0.890 & 0.262 & 0.262 & 0.005 & 12.080 & 0.009 & 0.011 & 0.982 & 0.097 & 0.097 & 0.001 & 4.444 & 0.003 \\
        DeepONet   & 0.154 & 0.812 & 0.388 & 0.388 & 0.009 & 10.799 & 0.015 & 0.015 & 0.954 & 0.117 & 0.117 & 0.003 & 4.520 & 0.008 \\
        LaMO       & 0.071 & 0.891 & 0.254 & 0.254 & 0.004 & 11.480 & 0.008 & 0.010 & 0.975 & 0.096 & 0.096 & 0.002 & 4.129 & 0.008 \\
        \bottomrule
    \end{tabular}
\end{table}

\begin{table}[p]
    \scriptsize
    \setlength{\tabcolsep}{3pt}
    \renewcommand{\arraystretch}{1.4}
    \centering
    \caption{Temperature-to-Source reconstruction under geometric OOD with the Type I double-layer defect (type 2) held out. Models are trained on the other five defect types and tested on the held-out type, using irregular and regular grids.}
    \label{tab:t2q_fsplit_test_2}
    \begin{tabular}{l | ccccccc | ccccccc}
        \toprule
        \multirow{2}{*}[-0.6ex]{Model}
        & \multicolumn{7}{c|}{Irregular Data} & \multicolumn{7}{c}{Regular Data} \\
        \cmidrule{2-15}
        & MSE & SSIM & RMSE & nRMSE & cRMSE & Max & bRMSE
        & MSE & SSIM & RMSE & nRMSE & cRMSE & Max & bRMSE \\
        \midrule
        MLP        & 0.416 & 0.729 & 0.637 & 0.637 & 0.012 & 23.081 & 0.035 & 0.223 & 0.926 & 0.427 & 0.427 & 0.017 & 22.349 & 0.011 \\
        FNO        & 0.451 & 0.686 & 0.658 & 0.658 & 0.008 & 26.187 & 0.039 & 0.202 & 0.913 & 0.395 & 0.395 & 0.006 & 22.316 & 0.015 \\
        GeoFNO     & 0.400 & 0.316 & 0.617 & 0.617 & 0.009 & 25.356 & 0.048 & 0.226 & 0.300 & 0.423 & 0.423 & 0.004 & 23.145 & 0.042 \\
        FFNO       & 0.389 & 0.572 & 0.617 & 0.617 & 0.018 & 23.058 & 0.043 & 0.177 & 0.881 & 0.364 & 0.364 & 0.022 & 20.369 & 0.021 \\
        FCNO       & 0.401 & 0.501 & 0.626 & 0.626 & 0.017 & 23.247 & 0.045 & 0.187 & 0.865 & 0.377 & 0.377 & 0.019 & 21.024 & 0.021 \\
        LNO        & 0.639 & 0.707 & 0.792 & 0.792 & 0.021 & 33.852 & 0.037 & 0.479 & 0.871 & 0.681 & 0.681 & 0.011 & 40.345 & 0.098 \\
        Transolver & 0.341 & 0.739 & 0.567 & 0.567 & 0.008 & 24.910 & 0.034 & 0.211 & 0.935 & 0.427 & 0.427 & 0.005 & 23.838 & 0.010 \\
        DeepONet   & 0.402 & 0.704 & 0.629 & 0.629 & 0.024 & 22.471 & 0.023 & 0.199 & 0.920 & 0.392 & 0.392 & 0.005 & 21.630 & 0.013 \\
        LaMO       & 0.360 & 0.747 & 0.584 & 0.584 & 0.011 & 24.267 & 0.028 & 0.201 & 0.942 & 0.413 & 0.413 & 0.005 & 23.154 & 0.008 \\
        \bottomrule
    \end{tabular}
\end{table}

\begin{table}[p]
    \scriptsize
    \setlength{\tabcolsep}{3pt}
    \renewcommand{\arraystretch}{1.4}
    \centering
    \caption{Temperature-to-Source reconstruction under geometric OOD with the Type II double-layer defect (type 3) held out. Models are trained on the other five defect types and tested on the held-out type, using irregular and regular grids.}
    \label{tab:t2q_fsplit_test_3}
    \begin{tabular}{l | ccccccc | ccccccc}
        \toprule
        \multirow{2}{*}[-0.6ex]{Model}
        & \multicolumn{7}{c|}{Irregular Data} & \multicolumn{7}{c}{Regular Data} \\
        \cmidrule{2-15}
        & MSE & SSIM & RMSE & nRMSE & cRMSE & Max & bRMSE
        & MSE & SSIM & RMSE & nRMSE & cRMSE & Max & bRMSE \\
        \midrule
        MLP        & 0.201 & 0.839 & 0.437 & 0.437 & 0.010 & 15.332 & 0.095 & 0.033 & 0.970 & 0.172 & 0.172 & 0.007 & 6.951 & 0.010 \\
        FNO        & 0.402 & 0.657 & 0.630 & 0.630 & 0.008 & 19.751 & 0.128 & 0.041 & 0.936 & 0.174 & 0.174 & 0.005 & 8.568 & 0.009 \\
        GeoFNO     & 0.187 & 0.208 & 0.422 & 0.422 & 0.007 & 15.607 & 0.156 & 0.057 & 0.531 & 0.199 & 0.199 & 0.003 & 9.764 & 0.045 \\
        FFNO       & 0.240 & 0.591 & 0.483 & 0.483 & 0.014 & 16.310 & 0.086 & 0.022 & 0.937 & 0.136 & 0.136 & 0.013 & 6.160 & 0.016 \\
        FCNO       & 0.227 & 0.540 & 0.467 & 0.467 & 0.013 & 15.565 & 0.139 & 0.022 & 0.911 & 0.137 & 0.137 & 0.011 & 6.180 & 0.024 \\
        LNO        & 0.326 & 0.786 & 0.554 & 0.554 & 0.008 & 24.485 & 0.123 & 0.145 & 0.882 & 0.369 & 0.369 & 0.011 & 14.693 & 0.185 \\
        Transolver & 0.141 & 0.886 & 0.344 & 0.344 & 0.004 & 16.562 & 0.078 & 0.022 & 0.979 & 0.128 & 0.128 & 0.002 & 6.470 & 0.006 \\
        DeepONet   & 0.184 & 0.806 & 0.417 & 0.417 & 0.012 & 14.387 & 0.102 & 0.024 & 0.960 & 0.143 & 0.143 & 0.002 & 6.235 & 0.007 \\
        LaMO       & 0.140 & 0.886 & 0.344 & 0.344 & 0.004 & 16.992 & 0.064 & 0.029 & 0.976 & 0.145 & 0.145 & 0.003 & 7.130 & 0.007 \\
        \bottomrule
    \end{tabular}
\end{table}

\begin{table}[p]
    \scriptsize
    \setlength{\tabcolsep}{3pt}
    \renewcommand{\arraystretch}{1.4}
    \centering
    \caption{Temperature-to-Source reconstruction under geometric OOD with the Type III double-layer defect (type 4) held out. Models are trained on the other five defect types and tested on the held-out type, using irregular and regular grids.}
    \label{tab:t2q_fsplit_test_4}
    \begin{tabular}{l | ccccccc | ccccccc}
        \toprule
        \multirow{2}{*}[-0.6ex]{Model}
        & \multicolumn{7}{c|}{Irregular Data} & \multicolumn{7}{c}{Regular Data} \\
        \cmidrule{2-15}
        & MSE & SSIM & RMSE & nRMSE & cRMSE & Max & bRMSE
        & MSE & SSIM & RMSE & nRMSE & cRMSE & Max & bRMSE \\
        \midrule
        MLP        & 0.287 & 0.772 & 0.530 & 0.530 & 0.010 & 17.339 & 0.232 & 0.094 & 0.946 & 0.303 & 0.303 & 0.012 & 13.430 & 0.008 \\
        FNO        & 0.439 & 0.619 & 0.658 & 0.658 & 0.026 & 19.476 & 0.331 & 0.074 & 0.917 & 0.267 & 0.267 & 0.003 & 11.876 & 0.010 \\
        GeoFNO     & 0.360 & 0.223 & 0.606 & 0.606 & 0.031 & 18.237 & 0.329 & 0.093 & 0.433 & 0.299 & 0.299 & 0.004 & 13.123 & 0.067 \\
        FFNO       & 0.343 & 0.557 & 0.581 & 0.581 & 0.012 & 18.606 & 0.212 & 0.071 & 0.919 & 0.262 & 0.262 & 0.013 & 11.768 & 0.016 \\
        FCNO       & 0.326 & 0.494 & 0.566 & 0.566 & 0.011 & 17.508 & 0.271 & 0.075 & 0.900 & 0.267 & 0.267 & 0.011 & 12.295 & 0.018 \\
        LNO        & 0.411 & 0.699 & 0.627 & 0.627 & 0.009 & 30.402 & 0.200 & 0.239 & 0.898 & 0.478 & 0.478 & 0.007 & 21.881 & 0.037 \\
        Transolver & 0.252 & 0.796 & 0.494 & 0.494 & 0.007 & 19.424 & 0.230 & 0.087 & 0.959 & 0.286 & 0.286 & 0.005 & 13.022 & 0.005 \\
        DeepONet   & 0.277 & 0.740 & 0.521 & 0.521 & 0.018 & 16.467 & 0.228 & 0.084 & 0.936 & 0.285 & 0.285 & 0.003 & 12.155 & 0.007 \\
        LaMO       & 0.248 & 0.809 & 0.478 & 0.478 & 0.012 & 23.242 & 0.201 & 0.093 & 0.955 & 0.291 & 0.291 & 0.005 & 14.129 & 0.005 \\
        \bottomrule
    \end{tabular}
\end{table}

\begin{table}[p]
    \scriptsize
    \setlength{\tabcolsep}{3pt}
    \renewcommand{\arraystretch}{1.4}
    \centering
    \caption{Temperature-to-Source reconstruction under geometric OOD with the Type I multi-layer defect (type 5) held out. Models are trained on the other five defect types and tested on the held-out type, using irregular and regular grids.}
    \label{tab:t2q_fsplit_test_5}
    \begin{tabular}{l | ccccccc | ccccccc}
        \toprule
        \multirow{2}{*}[-0.6ex]{Model}
        & \multicolumn{7}{c|}{Irregular Data} & \multicolumn{7}{c}{Regular Data} \\
        \cmidrule{2-15}
        & MSE & SSIM & RMSE & nRMSE & cRMSE & Max & bRMSE
        & MSE & SSIM & RMSE & nRMSE & cRMSE & Max & bRMSE \\
        \midrule
        MLP        & 0.155 & 0.811 & 0.388 & 0.388 & 0.016 & 13.142 & 0.122 & 0.041 & 0.959 & 0.197 & 0.197 & 0.010 & 8.328 & 0.010 \\
        FNO        & 0.323 & 0.664 & 0.564 & 0.564 & 0.009 & 17.689 & 0.191 & 0.049 & 0.926 & 0.212 & 0.212 & 0.005 & 8.778 & 0.012 \\
        GeoFNO     & 0.250 & 0.230 & 0.482 & 0.482 & 0.008 & 14.693 & 0.170 & 0.070 & 0.335 & 0.250 & 0.250 & 0.005 & 10.187 & 0.074 \\
        FFNO       & 0.209 & 0.556 & 0.453 & 0.453 & 0.021 & 16.310 & 0.152 & 0.043 & 0.929 & 0.197 & 0.197 & 0.013 & 7.994 & 0.017 \\
        FCNO       & 0.198 & 0.523 & 0.440 & 0.440 & 0.019 & 15.628 & 0.138 & 0.042 & 0.904 & 0.197 & 0.197 & 0.011 & 7.920 & 0.019 \\
        LNO        & 0.321 & 0.777 & 0.553 & 0.553 & 0.014 & 23.280 & 0.187 & 0.304 & 0.899 & 0.529 & 0.529 & 0.006 & 19.361 & 0.065 \\
        Transolver & 0.112 & 0.862 & 0.327 & 0.327 & 0.006 & 12.811 & 0.127 & 0.056 & 0.974 & 0.217 & 0.217 & 0.003 & 11.116 & 0.006 \\
        DeepONet   & 0.164 & 0.774 & 0.399 & 0.399 & 0.007 & 14.397 & 0.120 & 0.035 & 0.951 & 0.179 & 0.179 & 0.002 & 7.603 & 0.007 \\
        LaMO       & 0.116 & 0.861 & 0.330 & 0.330 & 0.005 & 13.486 & 0.133 & 0.049 & 0.969 & 0.205 & 0.205 & 0.004 & 10.108 & 0.009 \\
        \bottomrule
    \end{tabular}
\end{table}

\begin{table}[p]
    \scriptsize
    \setlength{\tabcolsep}{3pt}
    \renewcommand{\arraystretch}{1.4}
    \centering
    \caption{Temperature-to-Source reconstruction under geometric OOD with the Type II multi-layer defect (type 6) held out. Models are trained on the other five defect types and tested on the held-out type, using irregular and regular grids.}
    \label{tab:t2q_fsplit_test_6}
    \begin{tabular}{l | ccccccc | ccccccc}
        \toprule
        \multirow{2}{*}[-0.6ex]{Model}
        & \multicolumn{7}{c|}{Irregular Data} & \multicolumn{7}{c}{Regular Data} \\
        \cmidrule{2-15}
        & MSE & SSIM & RMSE & nRMSE & cRMSE & Max & bRMSE
        & MSE & SSIM & RMSE & nRMSE & cRMSE & Max & bRMSE \\
        \midrule
        MLP        & 0.213 & 0.796 & 0.460 & 0.460 & 0.009 & 12.320 & 0.163 & 0.126 & 0.966 & 0.299 & 0.299 & 0.004 & 20.133 & 0.014 \\
        FNO        & 0.503 & 0.555 & 0.707 & 0.707 & 0.008 & 19.070 & 0.180 & 0.020 & 0.948 & 0.141 & 0.141 & 0.004 & 4.767 & 0.009\\
        GeoFNO     & 0.175 & 0.211 & 0.413 & 0.413 & 0.006 & 15.362 & 0.240 & 0.076 & 0.649 & 0.248 & 0.248 & 0.003 & 14.481 & 0.041 \\
        FFNO       & 0.276 & 0.503 & 0.523 & 0.523 & 0.011 & 15.061 & 0.166 & 0.140 & 0.933 & 0.324 & 0.324 & 0.016 & 21.179 & 0.038 \\
        FCNO       & 0.256 & 0.462 & 0.504 & 0.504 & 0.010 & 13.762 & 0.155 & 0.144 & 0.904 & 0.338 & 0.338 & 0.015 & 21.425 & 0.025 \\
        LNO        & 0.408 & 0.781 & 0.633 & 0.633 & 0.011 & 25.231 & 0.218 & 0.341 & 0.890 & 0.574 & 0.574 & 0.004 & 28.309 & 0.007 \\
        Transolver & 0.128 & 0.866 & 0.342 & 0.342 & 0.007 & 17.324 & 0.070 & 0.032 & 0.959 & 0.174 & 0.174 & 0.004 & 7.249 & 0.009\\
        DeepONet   & 0.218 & 0.762 & 0.465 & 0.465 & 0.022 & 12.275 & 0.157 & 0.147 & 0.969 & 0.324 & 0.324 & 0.004 & 22.058 & 0.016  \\
        LaMO       & 0.113 & 0.859 & 0.325 & 0.325 & 0.005 & 15.699 & 0.071 & 0.033 & 0.941 & 0.173 & 0.173 & 0.003 & 8.837 & 0.009\\
        \bottomrule
    \end{tabular}
\end{table}

\begin{table}[p]
    \scriptsize
    \setlength{\tabcolsep}{2.5pt}
    \renewcommand{\arraystretch}{1.4}
    \centering
    \caption{Temperature-to-Source reconstruction performance (MSE / RMSE) for the all six defect types defect under normal training and three frequency-band OOD regimes, on irregular and regular grids.}
    \label{tab:t2q_multi_type_123456_ood}
    \begin{tabular}{l | cc | cc | cc | cc | cc | cc | cc | cc}
        \toprule
         &
        \multicolumn{8}{c|}{Irregular Data} &
        \multicolumn{8}{c}{Regular Data} \\
        \cmidrule{2-17}
        \multicolumn{1}{c|}{Model}  &
        \multicolumn{2}{c|}{Normal} &
        \multicolumn{2}{c|}{High-OOD} &
        \multicolumn{2}{c|}{Mid-OOD} &
        \multicolumn{2}{c|}{Low-OOD} &
        \multicolumn{2}{c|}{Normal} &
        \multicolumn{2}{c|}{High-OOD} &
        \multicolumn{2}{c|}{Mid-OOD} &
        \multicolumn{2}{c}{Low-OOD} \\
        \cmidrule{2-17}
        & MSE & RMSE & MSE & RMSE & MSE & RMSE & MSE & RMSE & MSE & RMSE & MSE & RMSE & MSE & RMSE & MSE & RMSE \\
        \midrule
        MLP        & 0.218 & 0.451 & 0.230 & 0.465 & 0.248 & 0.480 & 0.265 & 0.505 & 0.058 & 0.210 & 0.071 & 0.230 & 0.064 & 0.212 & 0.065 & 0.210 \\
        FNO        & 0.175 & 0.372 & 0.254 & 0.478 & 0.196 & 0.394 & 0.245 & 0.471 & 0.029 & 0.145 & 0.034 & 0.160 & 0.040 & 0.171 & 0.040 & 0.189 \\
        GeoFNO     & 0.217 & 0.450 & 0.237 & 0.471 & 0.262 & 0.492 & 0.272 & 0.513 & 0.061 & 0.214 & 0.083 & 0.240 & 0.069 & 0.219 & 0.073 & 0.256 \\
        FFNO       & 0.158 & 0.377 & 0.212 & 0.440 & 0.192 & 0.415 & 0.214 & 0.450 & 0.032 & 0.152 & 0.038 & 0.166 & 0.043 & 0.178 & 0.040 & 0.189 \\
        FCNO       & 0.178 & 0.406 & 0.212 & 0.442 & 0.208 & 0.439 & 0.233 & 0.472 & 0.035 & 0.160 & 0.039 & 0.171 & 0.048 & 0.188 & 0.044 & 0.199 \\
\bottomrule
    \end{tabular}
\end{table}

\begin{table}[p]
    \scriptsize
    \setlength{\tabcolsep}{2.2pt}
    \renewcommand{\arraystretch}{1.4}
    \centering
    \caption{Temperature-to-Source reconstruction performance (MSE / RMSE) for the all six defect types defect under single-frequency OOD (SFO). Models are trained on all frequencies (Normal) or on a single excitation frequency, evaluated on irregular and regular grids.}
    \label{tab:t2q_multi_type_123456_sfo}
    \begin{tabular}{l | cc | cc | cc | cc | cc | cc | cc | cc}
        \toprule
         &
        \multicolumn{8}{c|}{Irregular Data} &
        \multicolumn{8}{c}{Regular Data} \\
        \cmidrule{2-17}
        \multicolumn{1}{c|}{Model}  &
        \multicolumn{2}{c|}{Normal} &
        \multicolumn{2}{c|}{SFO-9kHz} &
        \multicolumn{2}{c|}{SFO-36kHz} &
        \multicolumn{2}{c|}{SFO-81kHz} &
        \multicolumn{2}{c|}{Normal} &
        \multicolumn{2}{c|}{SFO-9kHz} &
        \multicolumn{2}{c|}{SFO-36kHz} &
        \multicolumn{2}{c}{SFO-81kHz} \\
        \cmidrule{2-17}
        & MSE & RMSE & MSE & RMSE & MSE & RMSE & MSE & RMSE & MSE & RMSE & MSE & RMSE & MSE & RMSE & MSE & RMSE \\
        \midrule
        MLP        & 0.218 & 0.451 & 0.275 & 0.514 & 0.256 & 0.496 & 0.259 & 0.498 & 0.058 & 0.210 & 0.123 & 0.305 & 0.119 & 0.295 & 0.115 & 0.291 \\
        FNO        & 0.175 & 0.372 & 0.254 & 0.478 & 0.246 & 0.478 & 0.265 & 0.489 & 0.029 & 0.145 & 0.152 & 0.358 & 0.094 & 0.284 & 0.112 & 0.306 \\
        GeoFNO     & 0.217 & 0.450 & 0.306 & 0.541 & 0.270 & 0.508 & 0.285 & 0.521 & 0.061 & 0.214 & 0.153 & 0.326 & 0.192 & 0.357 & 0.147 & 0.341 \\
        FFNO       & 0.158 & 0.377 & 0.198 & 0.381 & 0.192 & 0.358 & 0.199 & 0.389 & 0.032 & 0.152 & 0.098 & 0.281 & 0.092 & 0.258 & 0.099 & 0.279 \\
        FCNO       & 0.178 & 0.406 & 0.225 & 0.466 & 0.217 & 0.459 & 0.231 & 0.473 & 0.035 & 0.160 & 0.106 & 0.286 & 0.099 & 0.272 & 0.107 & 0.295 \\
\bottomrule
    \end{tabular}
\end{table}

\begin{table}[p]
    \scriptsize
    \setlength{\tabcolsep}{2.5pt}
    \renewcommand{\arraystretch}{1.4}
    \centering
    \caption{Temperature-to-Source reconstruction performance (MSE / RMSE) for the single layer defect under normal training and three frequency-band OOD regimes, on irregular and regular grids.}
    \label{tab:t2q_single_ood}
    \begin{tabular}{l | cc | cc | cc | cc | cc | cc | cc | cc}
        \toprule
         &
        \multicolumn{8}{c|}{Irregular Data} &
        \multicolumn{8}{c}{Regular Data} \\
        \cmidrule{2-17}
        \multicolumn{1}{c|}{Model}  &
        \multicolumn{2}{c|}{Normal} &
        \multicolumn{2}{c|}{High-OOD} &
        \multicolumn{2}{c|}{Mid-OOD} &
        \multicolumn{2}{c|}{Low-OOD} &
        \multicolumn{2}{c|}{Normal} &
        \multicolumn{2}{c|}{High-OOD} &
        \multicolumn{2}{c|}{Mid-OOD} &
        \multicolumn{2}{c}{Low-OOD} \\
        \cmidrule{2-17}
        MSE & RMSE & MSE & RMSE & MSE & RMSE & MSE & RMSE & MSE & RMSE & MSE & RMSE & MSE & RMSE & MSE & RMSE \\
        \midrule
        MLP        & 0.127 & 0.352 & 0.146 & 0.371 & 0.154 & 0.378 & 0.273 & 0.505 & 0.020 & 0.139 & 0.027 & 0.160 & 0.030 & 0.163 & 0.049 & 0.213 \\
        FNO        & 0.042 & 0.193 & 0.061 & 0.232 & 0.046 & 0.202 & 0.053 & 0.224 & 0.005 & 0.061 & 0.012 & 0.088 & 0.004 & 0.061 & 0.014 & 0.112 \\
        GeoFNO     & 0.044 & 0.198 & 0.060 & 0.233 & 0.047 & 0.207 & 0.051 & 0.220 & 0.007 & 0.077 & 0.016 & 0.110 & 0.006 & 0.075 & 0.025 & 0.147 \\
        FFNO       & 0.042 & 0.195 & 0.059 & 0.233 & 0.043 & 0.199 & 0.065 & 0.246 & 0.005 & 0.066 & 0.011 & 0.091 & 0.005 & 0.065 & 0.021 & 0.136 \\
        FCNO       & 0.048 & 0.211 & 0.068 & 0.251 & 0.049 & 0.214 & 0.078 & 0.272 & 0.006 & 0.070 & 0.012 & 0.094 & 0.005 & 0.065 & 0.025 & 0.150 \\
\bottomrule
    \end{tabular}
\end{table}

\begin{table}[p]
    \scriptsize
    \setlength{\tabcolsep}{2.5pt}
    \renewcommand{\arraystretch}{1.4}
    \centering
    \caption{Temperature-to-Source reconstruction performance (MSE / RMSE) for the Type I double-layer defect under normal training and three frequency-band OOD regimes, on irregular and regular grids.}
    \label{tab:t2q_I_double_ood}
    \begin{tabular}{l | cc | cc | cc | cc | cc | cc | cc | cc}
        \toprule
         &
        \multicolumn{8}{c|}{Irregular Data} &
        \multicolumn{8}{c}{Regular Data} \\
        \cmidrule{2-17}
        \multicolumn{1}{c|}{Model}  &
        \multicolumn{2}{c|}{Normal} &
        \multicolumn{2}{c|}{High-OOD} &
        \multicolumn{2}{c|}{Mid-OOD} &
        \multicolumn{2}{c|}{Low-OOD} &
        \multicolumn{2}{c|}{Normal} &
        \multicolumn{2}{c|}{High-OOD} &
        \multicolumn{2}{c|}{Mid-OOD} &
        \multicolumn{2}{c}{Low-OOD} \\
        \cmidrule{2-17}
        & MSE & RMSE & MSE & RMSE & MSE & RMSE & MSE & RMSE & MSE & RMSE & MSE & RMSE & MSE & RMSE & MSE & RMSE \\
        \midrule
        MLP        & 0.286 & 0.533 & 0.327 & 0.567 & 0.402 & 0.628 & 0.347 & 0.583 & 0.115 & 0.331 & 0.135 & 0.344 & 0.235 & 0.459 & 0.151 & 0.353 \\
        FNO        & 0.164 & 0.385 & 0.261 & 0.492 & 0.199 & 0.426 & 0.189 & 0.430 & 0.077 & 0.255 & 0.131 & 0.325 & 0.092 & 0.277 & 0.109 & 0.321 \\
        GeoFNO     & 0.163 & 0.390 & 0.258 & 0.490 & 0.200 & 0.431 & 0.164 & 0.401 & 0.088 & 0.275 & 0.142 & 0.338 & 0.098 & 0.288 & 0.157 & 0.380 \\
        FFNO       & 0.156 & 0.393 & 0.166 & 0.398 & 0.263 & 0.501 & 0.202 & 0.438 & 0.074 & 0.252 & 0.156 & 0.341 & 0.095 & 0.277 & 0.106 & 0.316 \\
        FCNO       & 0.176 & 0.413 & 0.279 & 0.519 & 0.210 & 0.450 & 0.178 & 0.421 & 0.083 & 0.264 & 0.166 & 0.357 & 0.104 & 0.290 & 0.106 & 0.318 \\
\bottomrule
    \end{tabular}
\end{table}

\begin{table}[p]
    \scriptsize
    \setlength{\tabcolsep}{2.2pt}
    \renewcommand{\arraystretch}{1.4}
    \centering
    \caption{Temperature-to-Source reconstruction performance (MSE / RMSE) for the Type I double-layer defect under single-frequency OOD (SFO). Models are trained on all frequencies (Normal) or on a single excitation frequency, evaluated on irregular and regular grids.}
    \label{tab:t2q_I_double_sfo}
    \begin{tabular}{l | cc | cc | cc | cc | cc | cc | cc | cc}
        \toprule
         &
        \multicolumn{8}{c|}{Irregular Data} &
        \multicolumn{8}{c}{Regular Data} \\
        \cmidrule{2-17}
        \multicolumn{1}{c|}{Model}  &
        \multicolumn{2}{c|}{Normal} &
        \multicolumn{2}{c|}{SFO-9kHz} &
        \multicolumn{2}{c|}{SFO-36kHz} &
        \multicolumn{2}{c|}{SFO-81kHz} &
        \multicolumn{2}{c|}{Normal} &
        \multicolumn{2}{c|}{SFO-9kHz} &
        \multicolumn{2}{c|}{SFO-36kHz} &
        \multicolumn{2}{c}{SFO-81kHz} \\
        \cmidrule{2-17}
        & MSE & RMSE & MSE & RMSE & MSE & RMSE & MSE & RMSE & MSE & RMSE & MSE & RMSE & MSE & RMSE & MSE & RMSE \\
        \midrule
        MLP        & 0.286 & 0.533 & 0.327 & 0.567 & 0.345 & 0.583 & 0.331 & 0.572 & 0.115 & 0.331 & 0.168 & 0.388 & 0.156 & 0.372 & 0.149 & 0.360 \\
        FNO        & 0.164 & 0.385 & 0.292 & 0.520 & 0.270 & 0.504 & 0.284 & 0.519 & 0.077 & 0.255 & 0.232 & 0.442 & 0.149 & 0.366 & 0.207 & 0.426 \\
        GeoFNO     & 0.163 & 0.390 & 0.282 & 0.518 & 0.232 & 0.472 & 0.248 & 0.488 & 0.088 & 0.275 & 0.262 & 0.487 & 0.210 & 0.419 & 0.291 & 0.482 \\
        FFNO       & 0.156 & 0.393 & 0.215 & 0.453 & 0.193 & 0.434 & 0.209 & 0.451 & 0.074 & 0.252 & 0.125 & 0.330 & 0.146 & 0.344 & 0.145 & 0.347 \\
        FCNO       & 0.176 & 0.413 & 0.219 & 0.459 & 0.216 & 0.459 & 0.234 & 0.479 & 0.083 & 0.264 & 0.135 & 0.341 & 0.165 & 0.368 & 0.194 & 0.404 \\
\bottomrule
    \end{tabular}
\end{table}

\begin{table}[p]
    \scriptsize
    \setlength{\tabcolsep}{2.5pt}
    \renewcommand{\arraystretch}{1.4}
    \centering
    \caption{Temperature-to-Source reconstruction performance (MSE / RMSE) for the Type II double-layer defect under normal training and three frequency-band OOD regimes, on irregular and regular grids.}
    \label{tab:t2q_II_double_ood}
    \begin{tabular}{l | cc | cc | cc | cc | cc | cc | cc | cc}
        \toprule
         &
        \multicolumn{8}{c|}{Irregular Data} &
        \multicolumn{8}{c}{Regular Data} \\
        \cmidrule{2-17}
        \multicolumn{1}{c|}{Model}  &
        \multicolumn{2}{c|}{Normal} &
        \multicolumn{2}{c|}{High-OOD} &
        \multicolumn{2}{c|}{Mid-OOD} &
        \multicolumn{2}{c|}{Low-OOD} &
        \multicolumn{2}{c|}{Normal} &
        \multicolumn{2}{c|}{High-OOD} &
        \multicolumn{2}{c|}{Mid-OOD} &
        \multicolumn{2}{c}{Low-OOD} \\
        \cmidrule{2-17}
        & MSE & RMSE & MSE & RMSE & MSE & RMSE & MSE & RMSE & MSE & RMSE & MSE & RMSE & MSE & RMSE & MSE & RMSE \\
        \midrule
        MLP        & 0.163 & 0.397 & 0.172 & 0.406 & 0.194 & 0.424 & 0.256 & 0.493 & 0.017 & 0.127 & 0.022 & 0.142 & 0.024 & 0.150 & 0.051 & 0.213 \\
        FNO        & 0.318 & 0.559 & 0.321 & 0.554 & 0.320 & 0.562 & 0.320 & 0.565 & 0.004 & 0.058 & 0.006 & 0.075 & 0.004 & 0.059 & 0.016 & 0.117 \\
        GeoFNO     & 0.123 & 0.343 & 0.165 & 0.397 & 0.133 & 0.361 & 0.112 & 0.333 & 0.007 & 0.078 & 0.010 & 0.100 & 0.007 & 0.085 & 0.022 & 0.140 \\
        FFNO       & 0.218 & 0.463 & 0.224 & 0.468 & 0.250 & 0.482 & 0.278 & 0.520 & 0.004 & 0.065 & 0.005 & 0.067 & 0.007 & 0.083 & 0.029 & 0.161 \\
        FCNO       & 0.202 & 0.444 & 0.217 & 0.458 & 0.234 & 0.473 & 0.292 & 0.532 & 0.005 & 0.069 & 0.006 & 0.072 & 0.008 & 0.087 & 0.034 & 0.175 \\
\bottomrule
    \end{tabular}
\end{table}

\begin{table}[p]
    \scriptsize
    \setlength{\tabcolsep}{2.5pt}
    \renewcommand{\arraystretch}{1.4}
    \centering
    \caption{Temperature-to-Source reconstruction performance (MSE / RMSE) for the Type III double-layer defect under normal training and three frequency-band OOD regimes, on irregular and regular grids.}
    \label{tab:t2q_III_double_ood}
    \begin{tabular}{l | cc | cc | cc | cc | cc | cc | cc | cc}
        \toprule
         &
        \multicolumn{8}{c|}{Irregular Data} &
        \multicolumn{8}{c}{Regular Data} \\
        \cmidrule{2-17}
        \multicolumn{1}{c|}{Model}  &
        \multicolumn{2}{c|}{Normal} &
        \multicolumn{2}{c|}{High-OOD} &
        \multicolumn{2}{c|}{Mid-OOD} &
        \multicolumn{2}{c|}{Low-OOD} &
        \multicolumn{2}{c|}{Normal} &
        \multicolumn{2}{c|}{High-OOD} &
        \multicolumn{2}{c|}{Mid-OOD} &
        \multicolumn{2}{c}{Low-OOD} \\
        \cmidrule{2-17}
        & MSE & RMSE & MSE & RMSE & MSE & RMSE & MSE & RMSE & MSE & RMSE & MSE & RMSE & MSE & RMSE & MSE & RMSE \\
        \midrule
        MLP        & 0.219 & 0.464 & 0.222 & 0.465 & 0.245 & 0.485 & 0.232 & 0.481 & 0.081 & 0.282 & 0.091 & 0.297 & 0.081 & 0.280 & 0.097 & 0.310 \\
        FNO        & 0.244 & 0.479 & 0.250 & 0.488 & 0.251 & 0.496 & 0.295 & 0.542 & 0.052 & 0.225 & 0.063 & 0.243 & 0.056 & 0.232 & 0.081 & 0.281 \\
        GeoFNO     & 0.247 & 0.493 & 0.265 & 0.509 & 0.324 & 0.559 & 0.320 & 0.560 & 0.076 & 0.273 & 0.092 & 0.296 & 0.076 & 0.271 & 0.122 & 0.339 \\
        FFNO       & 0.227 & 0.471 & 0.242 & 0.480 & 0.229 & 0.474 & 0.259 & 0.507 & 0.047 & 0.212 & 0.059 & 0.234 & 0.052 & 0.223 & 0.059 & 0.241 \\
        FCNO       & 0.218 & 0.463 & 0.221 & 0.464 & 0.230 & 0.469 & 0.254 & 0.503 & 0.047 & 0.213 & 0.058 & 0.233 & 0.052 & 0.223 & 0.070 & 0.262 \\
\bottomrule
    \end{tabular}
\end{table}

\begin{table}[p]
    \scriptsize
    \setlength{\tabcolsep}{2.2pt}
    \renewcommand{\arraystretch}{1.4}
    \centering
    \caption{Temperature-to-Source reconstruction performance (MSE / RMSE) for the Type III double-layer defect under single-frequency OOD (SFO). Models are trained on all frequencies (Normal) or on a single excitation frequency, evaluated on irregular and regular grids.}
    \label{tab:t2q_III_double_sfo}
    \begin{tabular}{l | cc | cc | cc | cc | cc | cc | cc | cc}
        \toprule
         &
        \multicolumn{8}{c|}{Irregular Data} &
        \multicolumn{8}{c}{Regular Data} \\
        \cmidrule{2-17}
        \multicolumn{1}{c|}{Model}  &
        \multicolumn{2}{c|}{Normal} &
        \multicolumn{2}{c|}{SFO-9kHz} &
        \multicolumn{2}{c|}{SFO-36kHz} &
        \multicolumn{2}{c|}{SFO-81kHz} &
        \multicolumn{2}{c|}{Normal} &
        \multicolumn{2}{c|}{SFO-9kHz} &
        \multicolumn{2}{c|}{SFO-36kHz} &
        \multicolumn{2}{c}{SFO-81kHz} \\
        \cmidrule{2-17}
        & MSE & RMSE & MSE & RMSE & MSE & RMSE & MSE & RMSE & MSE & RMSE & MSE & RMSE & MSE & RMSE & MSE & RMSE \\
        \midrule
        MLP        & 0.219 & 0.464 & 0.251 & 0.497 & 0.224 & 0.469 & 0.231 & 0.477 & 0.081 & 0.282 & 0.087 & 0.292 & 0.083 & 0.286 & 0.086 & 0.286 \\
        FNO        & 0.244 & 0.479 & 0.306 & 0.549 & 0.276 & 0.522 & 0.280 & 0.526 & 0.052 & 0.225 & 0.076 & 0.270 & 0.068 & 0.254 & 0.072 & 0.263 \\
        GeoFNO     & 0.247 & 0.493 & 0.343 & 0.577 & 0.288 & 0.528 & 0.455 & 0.628 & 0.076 & 0.273 & 0.252 & 0.421 & 0.266 & 0.430 & 0.145 & 0.348 \\
        FFNO       & 0.227 & 0.471 & 0.273 & 0.517 & 0.235 & 0.481 & 0.245 & 0.491 & 0.047 & 0.212 & 0.060 & 0.240 & 0.056 & 0.232 & 0.059 & 0.238 \\
        FCNO       & 0.218 & 0.463 & 0.259 & 0.505 & 0.230 & 0.476 & 0.243 & 0.489 & 0.047 & 0.213 & 0.062 & 0.243 & 0.059 & 0.238 & 0.061 & 0.239 \\
\bottomrule
    \end{tabular}
\end{table}

\begin{table}[p]
    \scriptsize
    \setlength{\tabcolsep}{2.5pt}
    \renewcommand{\arraystretch}{1.4}
    \centering
    \caption{Temperature-to-Source reconstruction performance (MSE / RMSE) for the Type I multi-layer defect under normal training and three frequency-band OOD regimes, on irregular and regular grids.}
    \label{tab:t2q_I_multi_ood}
    \begin{tabular}{l | cc | cc | cc | cc | cc | cc | cc | cc}
        \toprule
         &
        \multicolumn{8}{c|}{Irregular Data} &
        \multicolumn{8}{c}{Regular Data} \\
        \cmidrule{2-17}
        \multicolumn{1}{c|}{Model}  &
        \multicolumn{2}{c|}{Normal} &
        \multicolumn{2}{c|}{High-OOD} &
        \multicolumn{2}{c|}{Mid-OOD} &
        \multicolumn{2}{c|}{Low-OOD} &
        \multicolumn{2}{c|}{Normal} &
        \multicolumn{2}{c|}{High-OOD} &
        \multicolumn{2}{c|}{Mid-OOD} &
        \multicolumn{2}{c}{Low-OOD} \\
        \cmidrule{2-17}
        & MSE & RMSE & MSE & RMSE & MSE & RMSE & MSE & RMSE & MSE & RMSE & MSE & RMSE & MSE & RMSE & MSE & RMSE \\
        \midrule
        MLP        & 0.142 & 0.373 & 0.153 & 0.385 & 0.164 & 0.395 & 0.211 & 0.453 & 0.039 & 0.193 & 0.044 & 0.204 & 0.055 & 0.225 & 0.043 & 0.205 \\
        FNO        & 0.272 & 0.517 & 0.288 & 0.534 & 0.282 & 0.529 & 0.291 & 0.538 & 0.032 & 0.174 & 0.037 & 0.174 & 0.058 & 0.224 & 0.036 & 0.179 \\
        GeoFNO     & 0.181 & 0.421 & 0.201 & 0.440 & 0.211 & 0.455 & 0.209 & 0.453 & 0.042 & 0.186 & 0.070 & 0.245 & 0.044 & 0.197 & 0.044 & 0.182 \\
        FFNO       & 0.198 & 0.441 & 0.202 & 0.442 & 0.195 & 0.438 & 0.242 & 0.488 & 0.028 & 0.154 & 0.046 & 0.198 & 0.026 & 0.153 & 0.033 & 0.177 \\
        FCNO       & 0.194 & 0.437 & 0.196 & 0.436 & 0.184 & 0.427 & 0.259 & 0.506 & 0.028 & 0.155 & 0.043 & 0.189 & 0.026 & 0.152 & 0.033 & 0.177 \\
\bottomrule
    \end{tabular}
\end{table}

\begin{table}[p]
    \scriptsize
    \setlength{\tabcolsep}{2.5pt}
    \renewcommand{\arraystretch}{1.4}
    \centering
    \caption{Temperature-to-Source reconstruction performance (MSE / RMSE) for the Type II multi-layer defect under normal training and three frequency-band OOD regimes, on irregular and regular grids.}
    \label{tab:t2q_II_multi_ood}
    %
\end{table}

\begin{table}[p]
    \scriptsize
    \setlength{\tabcolsep}{2.2pt}
    \renewcommand{\arraystretch}{1.4}
    \centering
    \caption{Temperature-to-Source reconstruction performance (MSE / RMSE) for the Type II multi-layer defect under single-frequency OOD (SFO). Models are trained on all frequencies (Normal) or on a single excitation frequency, evaluated on irregular and regular grids.}
    \label{tab:t2q_II_multi_sfo}
    %
\end{table}

\clearpage


\begin{table}[p]
    \scriptsize
    \setlength{\tabcolsep}{3pt}
    \renewcommand{\arraystretch}{1.4}
    \centering
    \caption{Source-to-Temperature reconstruction under geometric OOD with the single-layer defect (type 1) held out. Models are trained on the other five defect types and tested on the held-out type, using irregular and regular grids.}
    \label{tab:q2t_fsplit_test_1}
    %
\end{table}

\begin{table}[p]
    \scriptsize
    \setlength{\tabcolsep}{3pt}
    \renewcommand{\arraystretch}{1.4}
    \centering
    \caption{Source-to-Temperature reconstruction under geometric OOD with the Type I double-layer defect (type 2) held out. Models are trained on the other five defect types and tested on the held-out type, using irregular and regular grids.}
    \label{tab:q2t_fsplit_test_2}
    %
\end{table}

\begin{table}[p]
    \scriptsize
    \setlength{\tabcolsep}{3pt}
    \renewcommand{\arraystretch}{1.4}
    \centering
    \caption{Source-to-Temperature reconstruction under geometric OOD with the Type II double-layer defect (type 3) held out. Models are trained on the other five defect types and tested on the held-out type, using irregular and regular grids.}
    \label{tab:q2t_fsplit_test_3}
    %
\end{table}

\begin{table}[p]
    \scriptsize
    \setlength{\tabcolsep}{3pt}
    \renewcommand{\arraystretch}{1.4}
    \centering
    \caption{Source-to-Temperature reconstruction under geometric OOD with the Type III double-layer defect (type 4) held out. Models are trained on the other five defect types and tested on the held-out type, using irregular and regular grids.}
    \label{tab:q2t_fsplit_test_4}
    %
\end{table}

\begin{table}[p]
    \scriptsize
    \setlength{\tabcolsep}{3pt}
    \renewcommand{\arraystretch}{1.4}
    \centering
    \caption{Source-to-Temperature reconstruction under geometric OOD with the Type I multi-layer defect (type 5) held out. Models are trained on the other five defect types and tested on the held-out type, using irregular and regular grids.}
    \label{tab:q2t_fsplit_test_5}
    %
\end{table}

\begin{table}[p]
    \scriptsize
    \setlength{\tabcolsep}{3pt}
    \renewcommand{\arraystretch}{1.4}
    \centering
    \caption{Source-to-Temperature reconstruction under geometric OOD with the Type II multi-layer defect (type 6) held out. Models are trained on the other five defect types and tested on the held-out type, using irregular and regular grids.}
    \label{tab:q2t_fsplit_test_6}
    %
\end{table}

\begin{table}[p]
    \scriptsize
    \setlength{\tabcolsep}{2.5pt}
    \renewcommand{\arraystretch}{1.4}
    \centering
    \caption{Source-to-Temperature reconstruction performance (MSE / RMSE) for the all six defect types defect under normal training and three frequency-band OOD regimes, on irregular and regular grids.}
    \label{tab:q2t_multi_type_123456_ood}
    %
\end{table}

\begin{table}[p]
    \scriptsize  
    \setlength{\tabcolsep}{2.2pt}
    \renewcommand{\arraystretch}{1.4}
    \centering
    \caption{Source-to-Temperature reconstruction performance (MSE / RMSE) for the all six defect types defect under single-frequency OOD (SFO). Models are trained on all frequencies (Normal) or on a single excitation frequency, evaluated on irregular and regular grids.}
    \label{tab:q2t_multi_type_123456_sfo}
    %
\end{table}

\begin{table}[p]
    \scriptsize
    \setlength{\tabcolsep}{2.5pt}
    \renewcommand{\arraystretch}{1.4}
    \centering
    \caption{Source-to-Temperature reconstruction performance (MSE / RMSE) for the single layer defect under normal training and three frequency-band OOD regimes, on irregular and regular grids.}
    \label{tab:q2t_single_ood}
    %
\end{table}

\begin{table}[p]
    \scriptsize
    \setlength{\tabcolsep}{2.5pt}
    \renewcommand{\arraystretch}{1.4}
    \centering
    \caption{Source-to-Temperature reconstruction performance (MSE / RMSE) for the Type I double-layer defect under normal training and three frequency-band OOD regimes, on irregular and regular grids.}
    \label{tab:q2t_I_double_ood}
    %
\end{table}

\begin{table}[p]
    \scriptsize
    \setlength{\tabcolsep}{2.2pt}
    \renewcommand{\arraystretch}{1.4}
    \centering
    \caption{Source-to-Temperature reconstruction performance (MSE / RMSE) for the Type I double-layer defect under single-frequency OOD (SFO). Models are trained on all frequencies (Normal) or on a single excitation frequency, evaluated on irregular and regular grids.}
    \label{tab:q2t_I_double_sfo}
    %
\end{table}

\begin{table}[p]
    \scriptsize
    \setlength{\tabcolsep}{2.5pt}
    \renewcommand{\arraystretch}{1.4}
    \centering
    \caption{Source-to-Temperature reconstruction performance (MSE / RMSE) for the Type II double-layer defect under normal training and three frequency-band OOD regimes, on irregular and regular grids.}
    \label{tab:q2t_II_double_ood}
    %
\end{table}

\begin{table}[p]
    \scriptsize
    \setlength{\tabcolsep}{2.5pt}
    \renewcommand{\arraystretch}{1.4}
    \centering
    \caption{Source-to-Temperature reconstruction performance (MSE / RMSE) for the Type III double-layer defect under normal training and three frequency-band OOD regimes, on irregular and regular grids.}
    \label{tab:q2t_III_double_ood}
    %
\end{table}

\begin{table}[p]
    \scriptsize
    \setlength{\tabcolsep}{2.2pt}
    \renewcommand{\arraystretch}{1.4}
    \centering
    \caption{Source-to-Temperature reconstruction performance (MSE / RMSE) for the Type III double-layer defect under single-frequency OOD (SFO). Models are trained on all frequencies (Normal) or on a single excitation frequency, evaluated on irregular and regular grids.}
    \label{tab:q2t_III_double_sfo}
    %
\end{table}

\begin{table}[p]
    \scriptsize
    \setlength{\tabcolsep}{2.5pt}
    \renewcommand{\arraystretch}{1.4}
    \centering
    \caption{Source-to-Temperature reconstruction performance (MSE / RMSE) for the Type I multi-layer defect under normal training and three frequency-band OOD regimes, on irregular and regular grids.}
    \label{tab:q2t_I_multi_ood}
    %
\end{table}

\clearpage


\begin{table}[p]
    \scriptsize
    \setlength{\tabcolsep}{3pt}
    \renewcommand{\arraystretch}{1.4}
    \centering
    \caption{Temperature-to-Temperature prediction under geometric OOD with the single-layer defect (type 1) held out. Models are trained on the other five defect types and tested on the held-out type, using irregular and regular grids. MSE and RMSE values are multiplied by $10^3$.}
    \label{tab:t2t_fsplit_test_1}

\end{table}

\begin{table}[p]
    \scriptsize
    \setlength{\tabcolsep}{3pt}
    \renewcommand{\arraystretch}{1.4}
    \centering
    \caption{Temperature-to-Temperature prediction under geometric OOD with the Type I double-layer defect (type 2) held out. Models are trained on the other five defect types and tested on the held-out type, using irregular and regular grids. MSE and RMSE values are multiplied by $10^3$.}
    \label{tab:t2t_fsplit_test_2}
    %
\end{table}

\begin{table}[p]
    \scriptsize
    \setlength{\tabcolsep}{3pt}
    \renewcommand{\arraystretch}{1.4}
    \centering
    \caption{Temperature-to-Temperature prediction under geometric OOD with the Type II double-layer defect (type 3) held out. Models are trained on the other five defect types and tested on the held-out type, using irregular and regular grids. MSE and RMSE values are multiplied by $10^3$.}
    \label{tab:t2t_fsplit_test_3}
    %
\end{table}

\begin{table}[p]
    \scriptsize
    \setlength{\tabcolsep}{3pt}
    \renewcommand{\arraystretch}{1.4}
    \centering
    \caption{Temperature-to-Temperature prediction under geometric OOD with the Type III double-layer defect (type 4) held out. Models are trained on the other five defect types and tested on the held-out type, using irregular and regular grids. MSE and RMSE values are multiplied by $10^3$.}
    \label{tab:t2t_fsplit_test_4}
    %
\end{table}

\begin{table}[p]
    \scriptsize
    \setlength{\tabcolsep}{3pt}
    \renewcommand{\arraystretch}{1.4}
    \centering
    \caption{Temperature-to-Temperature prediction under geometric OOD with the Type I multi-layer defect (type 5) held out. Models are trained on the other five defect types and tested on the held-out type, using irregular and regular grids. MSE and RMSE values are multiplied by $10^3$.}
    \label{tab:t2t_fsplit_test_5}
    %
\end{table}

\begin{table}[p]
    \scriptsize
    \setlength{\tabcolsep}{3pt}
    \renewcommand{\arraystretch}{1.4}
    \centering
    \caption{Temperature-to-Temperature prediction under geometric OOD with the Type II multi-layer defect (type 6) held out. Models are trained on the other five defect types and tested on the held-out type, using irregular and regular grids. MSE and RMSE values are multiplied by $10^3$.}
    \label{tab:t2t_fsplit_test_6}
    %
\end{table}

\begin{table}[p]
    \tiny
    \setlength{\tabcolsep}{2.5pt}
    \renewcommand{\arraystretch}{1.4}
    \centering
    \caption{Temperature-to-Temperature reconstruction performance (MSE / RMSE, scaled by $10^{-3}$) for the all six defect types defect under normal training and three frequency-band OOD regimes, on irregular and regular grids.}
    \label{tab:t2t_multi_type_123456_ood}
    %
\end{table}

\begin{table}[p]
    \tiny
    \setlength{\tabcolsep}{2.5pt}
    \renewcommand{\arraystretch}{1.4}
    \centering
    \caption{Temperature-to-Temperature reconstruction performance (MSE / RMSE, scaled by $10^{-3}$) for the all six defect types defect under single-frequency OOD (SFO). Models are trained on all frequencies (Normal) or on a single excitation frequency, evaluated on irregular and regular grids.}
    \label{tab:t2t_multi_type_123456_sfo}
    %
\end{table}

\begin{table}[p]
    \tiny
    \setlength{\tabcolsep}{2.5pt}
    \renewcommand{\arraystretch}{1.4}
    \centering
    \caption{Temperature-to-Temperature reconstruction performance (MSE / RMSE, scaled by $10^{-3}$) for the single layer defect under normal training and three frequency-band OOD regimes, on irregular and regular grids.}
    \label{tab:t2t_single_ood}
    %
\end{table}

\begin{table}[p]
    \tiny
    \setlength{\tabcolsep}{2.2pt}
    \renewcommand{\arraystretch}{1.4}
    \centering
    \caption{Temperature-to-Temperature reconstruction performance (MSE / RMSE, scaled by $10^{-3}$) for the Type I double-layer defect under normal training and three frequency-band OOD regimes, on irregular and regular grids.}
    \label{tab:t2t_I_double_ood}
    %
\end{table}

\begin{table}[p]
    \tiny
    \setlength{\tabcolsep}{2pt}
    \renewcommand{\arraystretch}{1.4}
    \centering
    \caption{Temperature-to-Temperature reconstruction performance (MSE / RMSE, scaled by $10^{-3}$) for the Type I double-layer defect under single-frequency OOD (SFO). Models are trained on all frequencies (Normal) or on a single excitation frequency, evaluated on irregular and regular grids.}
    \label{tab:t2t_I_double_sfo}
    %
\end{table}

\begin{table}[p]
    \tiny
    \setlength{\tabcolsep}{2.5pt}
    \renewcommand{\arraystretch}{1.4}
    \centering
    \caption{Temperature-to-Temperature reconstruction performance (MSE / RMSE, scaled by $10^{-3}$) for the Type II double-layer defect under normal training and three frequency-band OOD regimes, on irregular and regular grids.}
    \label{tab:t2t_II_double_ood}
    %
\end{table}

\begin{table}[p]
    \tiny
    \setlength{\tabcolsep}{2pt}
    \renewcommand{\arraystretch}{1.4}
    \centering
    \caption{Temperature-to-Temperature reconstruction performance (MSE / RMSE, scaled by $10^{-3}$) for the Type III double-layer defect under normal training and three frequency-band OOD regimes, on irregular and regular grids.}
    \label{tab:t2t_III_double_ood}
    %
\end{table}

\begin{table}[p]
    \tiny
    \setlength{\tabcolsep}{1.5pt}
    \renewcommand{\arraystretch}{1.4}
    \centering
    \caption{Temperature-to-Temperature reconstruction performance (MSE / RMSE, scaled by $10^{-3}$) for the Type III double-layer defect under single-frequency OOD (SFO). Models are trained on all frequencies (Normal) or on a single excitation frequency, evaluated on irregular and regular grids.}
    \label{tab:t2t_III_double_sfo}
    %
\end{table}

\begin{table}[p]
    \tiny
    \setlength{\tabcolsep}{2pt}
    \renewcommand{\arraystretch}{1.4}
    \centering
    \caption{Temperature-to-Temperature reconstruction performance (MSE / RMSE, scaled by $10^{-3}$) for the Type I multi-layer defect under normal training and three frequency-band OOD regimes, on irregular and regular grids.}
    \label{tab:t2t_I_multi_ood}
    %
\end{table}

\begin{table}[p]
    \tiny
    \setlength{\tabcolsep}{2.5pt}
    \renewcommand{\arraystretch}{1.4}
    \centering
    \caption{Temperature-to-Temperature reconstruction performance (MSE / RMSE, scaled by $10^{-3}$) for the Type II multi-layer defect under normal training and three frequency-band OOD regimes, on irregular and regular grids.}
    \label{tab:t2t_II_multi_ood}
    %
\end{table}

\begin{table}[p]
    \tiny
    \setlength{\tabcolsep}{2pt}
    \renewcommand{\arraystretch}{1.4}
    \centering
    \caption{Temperature-to-Temperature reconstruction performance (MSE / RMSE, scaled by $10^{-3}$) for the Type II multi-layer defect under single-frequency OOD (SFO). Models are trained on all frequencies (Normal) or on a single excitation frequency, evaluated on irregular and regular grids.}
    \label{tab:t2t_II_multi_sfo}
    %
\end{table}

\clearpage

\section{More Visualizations}
\label{appendix: visualization_all}
The following presents visualizations of all experimental results across all tasks, including Forward Modeling, Inverse Source Reconstruction, Temporal Evolution Prediction and Surface-to-Source Reconstruction, involving regular/irregular grids and full-frequency/OOD environments. We selected OOD-high (high-frequency hold-out) regime for visualization. We have selected experimental results from FNO, Transolver, GeoFNO, and LNO models trained jointly on all six defect types for display, with visualizations provided for inputs and outputs under each configuration.

\begin{figure}[htbp]
    \centering
    \includegraphics[width=1\textwidth]{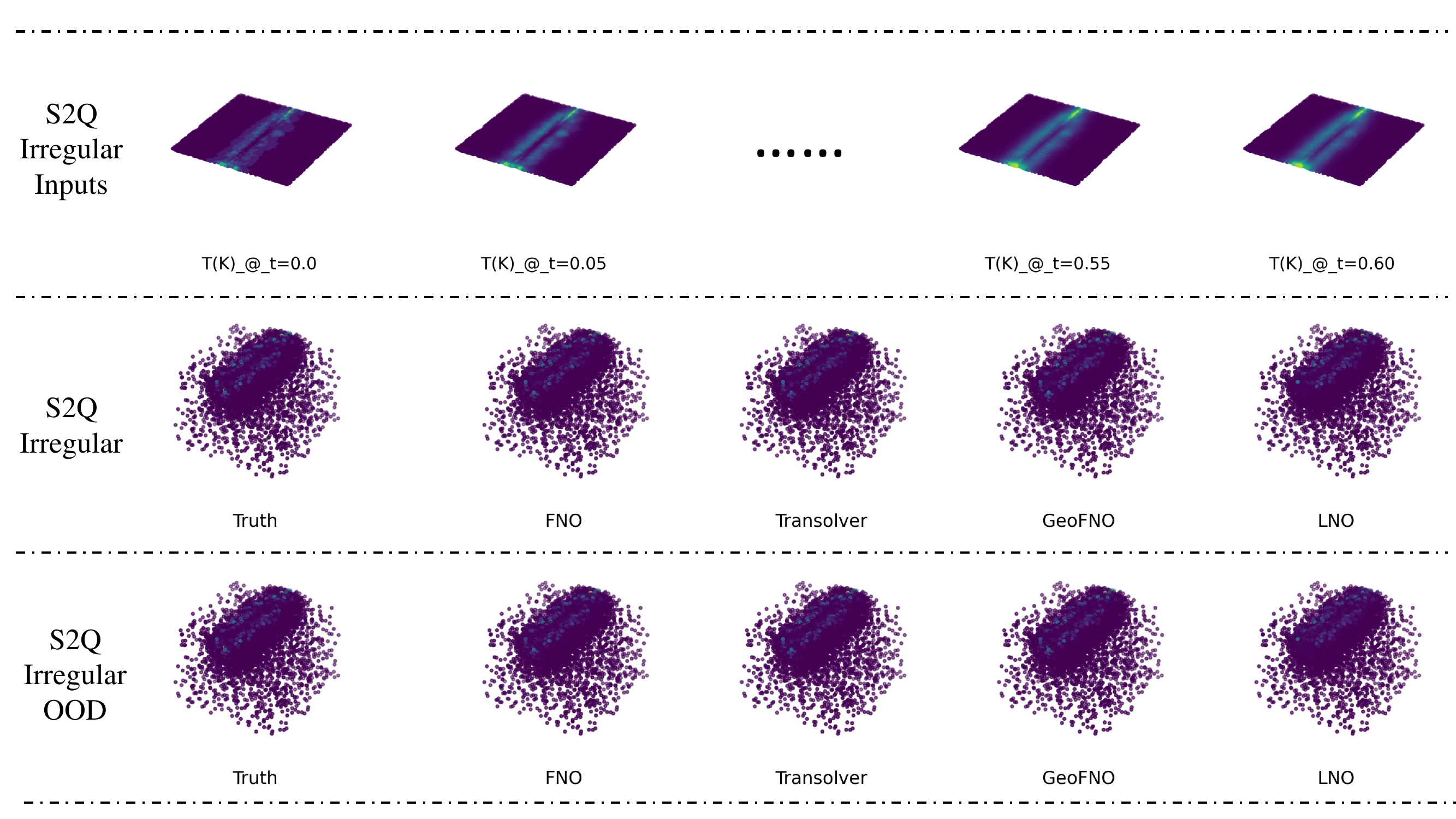}
    \caption{Visualization of experimental results under the surface temperature to heat (S2Q) inverse task on irregular grids, using models trained jointly on all six defect types of the Aletheia dataset.}
    \label{fig:vis-S2Q-uns}
\end{figure}

\begin{figure}[htbp]
    \centering
    \includegraphics[width=1\textwidth]{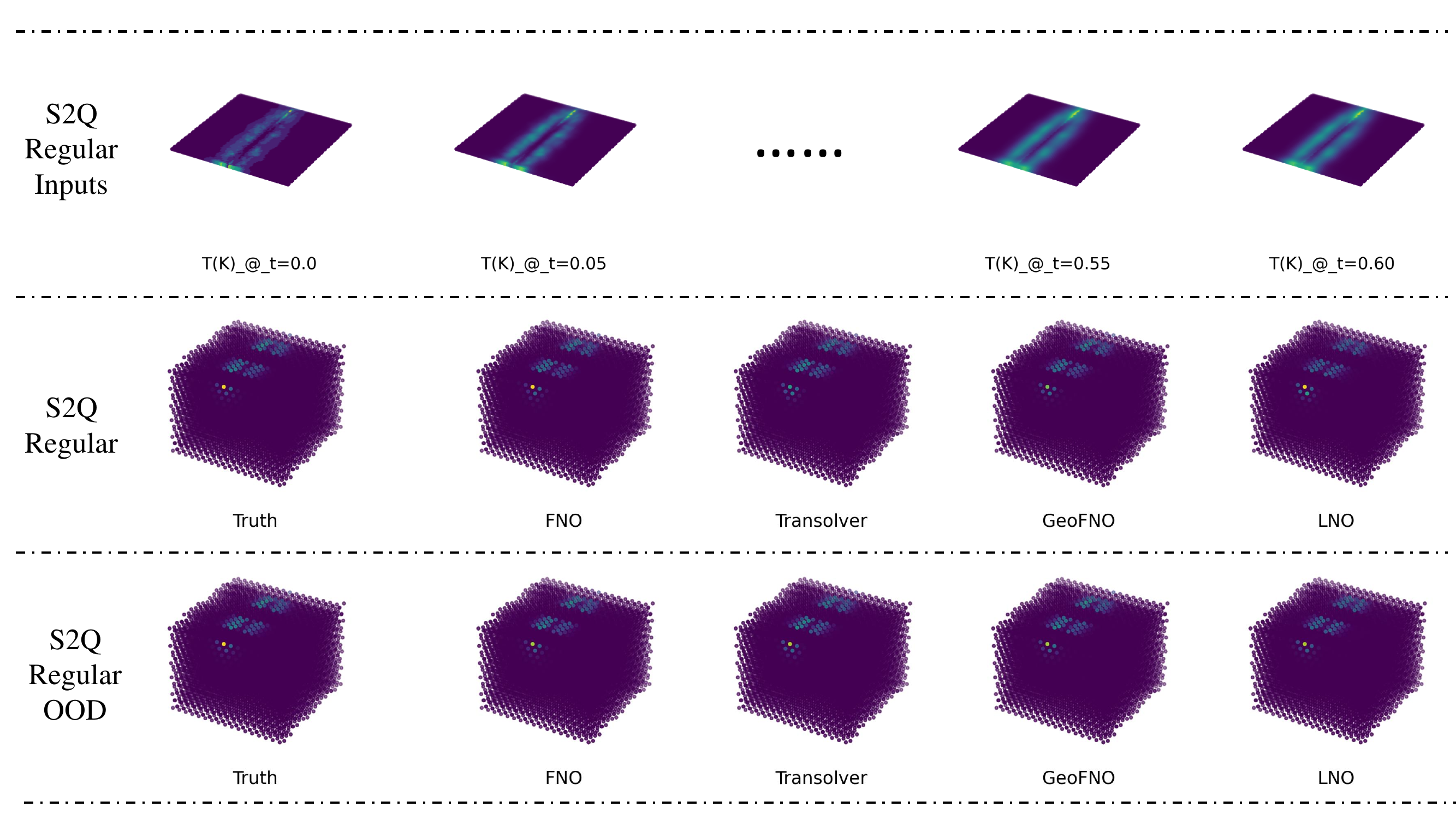}
    \caption{Visualization of experimental results under the surface temperature to heat (S2Q) inverse task on regular grids, using models trained jointly on all six defect types of the Aletheia dataset.}
    \label{fig:vis-S2Q-s}
\end{figure}

\begin{figure}[htbp]
    \centering
    \includegraphics[width=1\textwidth]{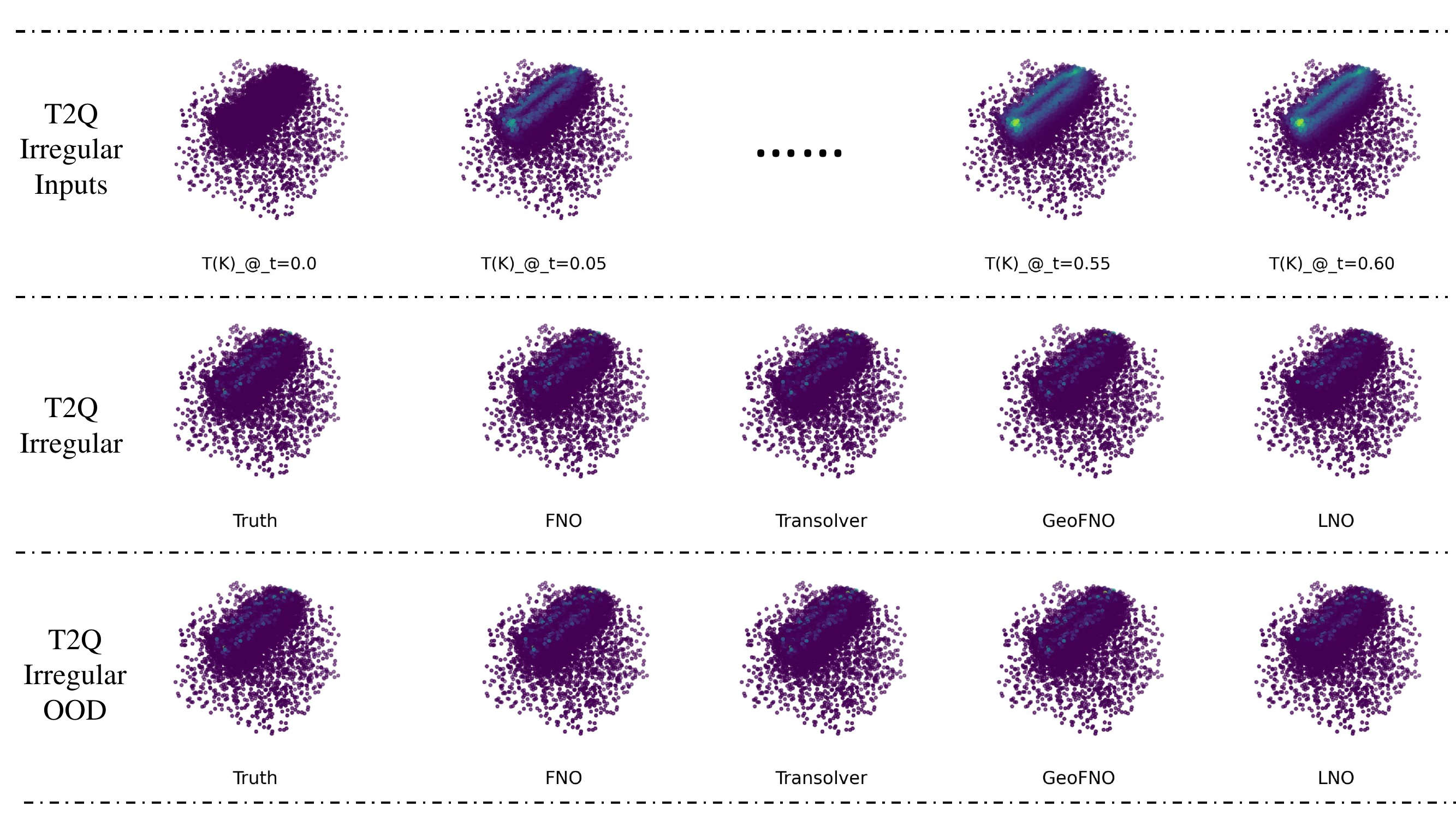}
    \caption{Visualization of experimental results under the temperature to heat (T2Q) inverse task on irregular grids, using models trained jointly on all six defect types of the Aletheia dataset.}
    \label{fig:vis-T2Q-uns}
\end{figure}

\begin{figure}[htbp]
    \centering
    \includegraphics[width=1\textwidth]{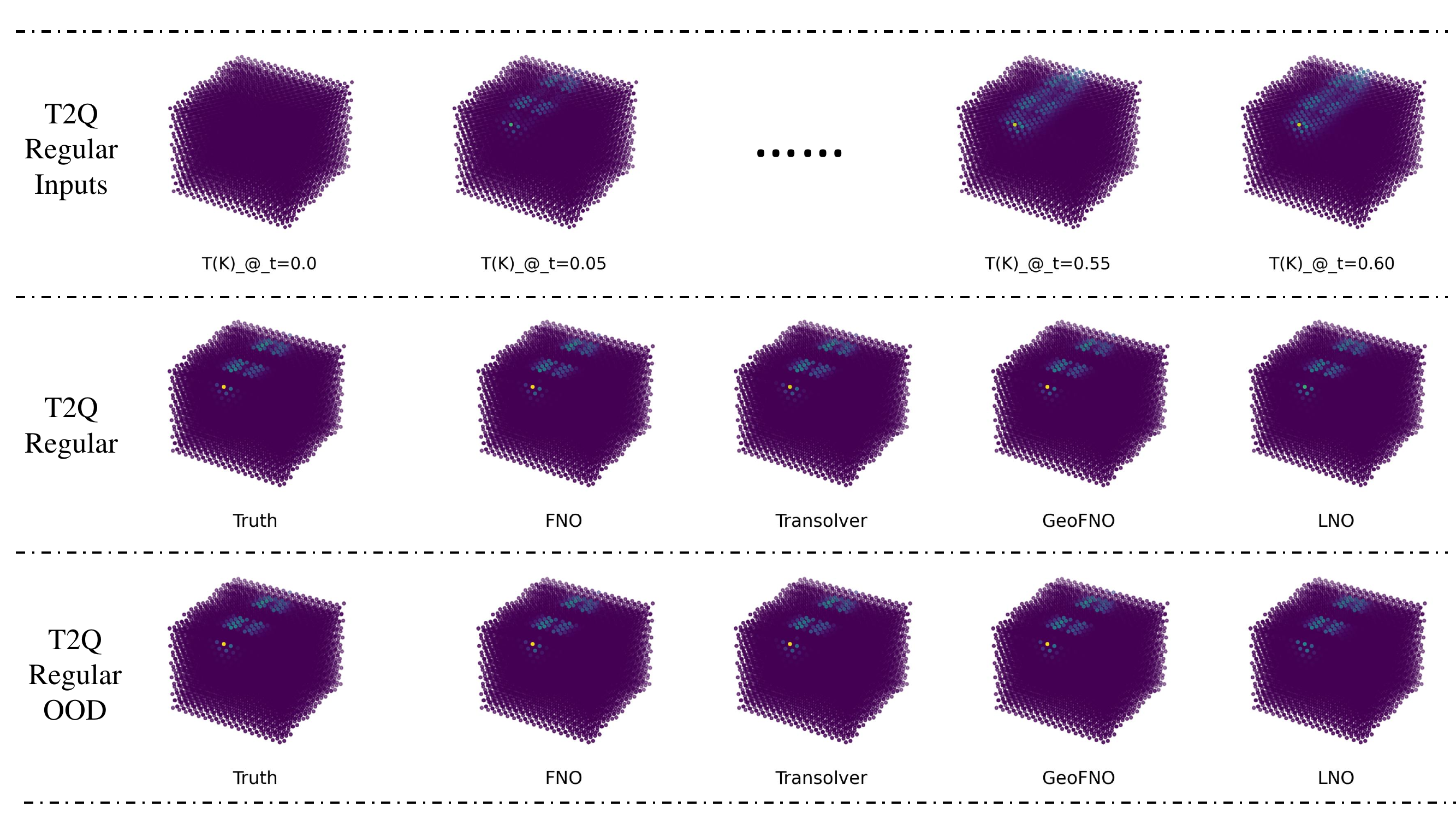}
    \caption{Visualization of experimental results under the temperature to heat (T2Q) inverse task on regular grids, using models trained jointly on all six defect types of the Aletheia dataset.}
    \label{fig:vis-T2Q-s}
\end{figure}

\begin{figure}[htbp]
    \centering
    \includegraphics[width=1\textwidth]{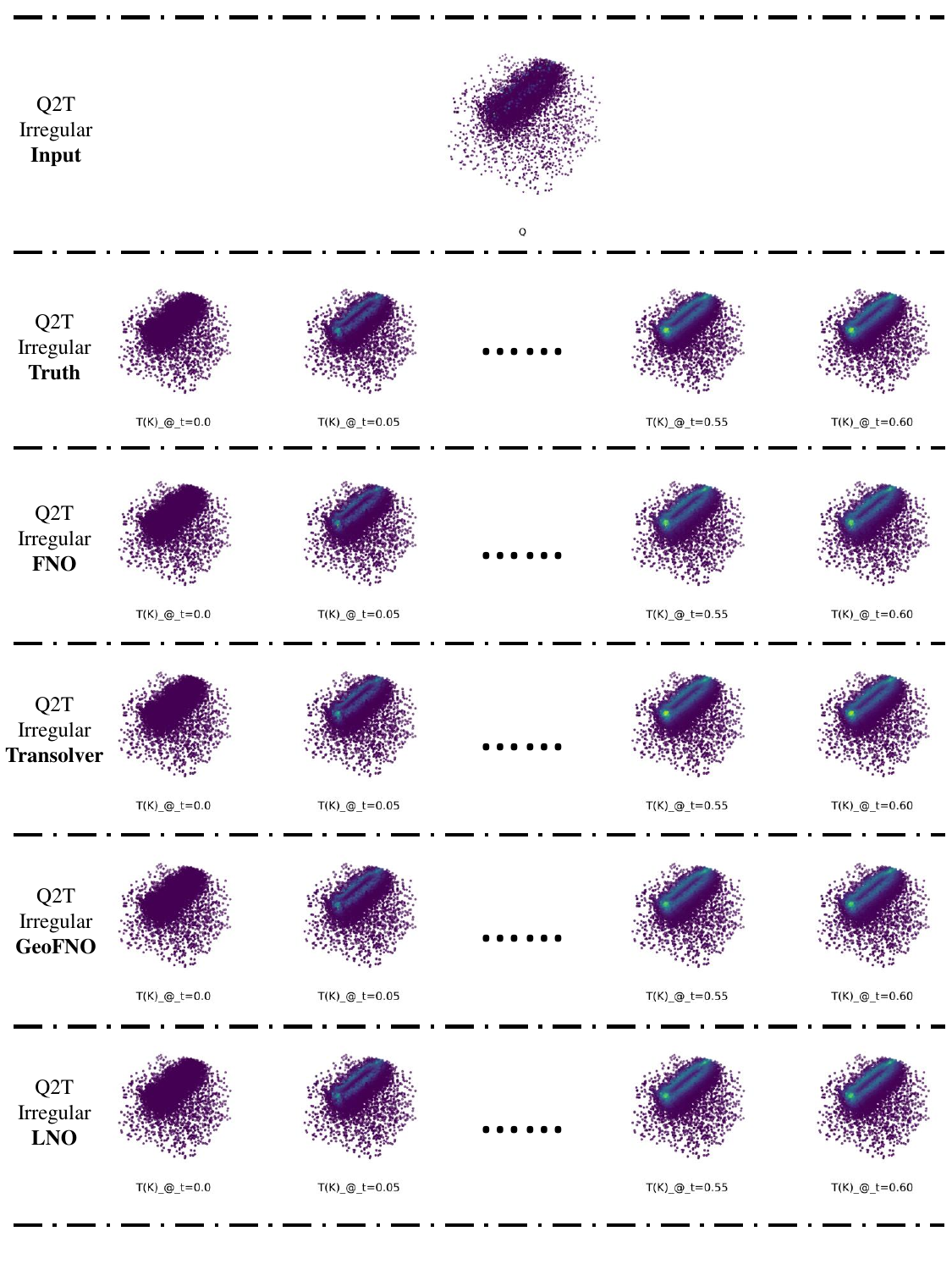}
    \caption{Visualization of experimental results under the heat-to-temperature (Q2T) forward task on irregular grids under full-frequency setting, using models trained jointly on all six defect types of the Aletheia dataset.}
    \label{fig:vis-Q2T-uns}
\end{figure}

\begin{figure}[htbp]
    \centering
    \includegraphics[width=1\textwidth]{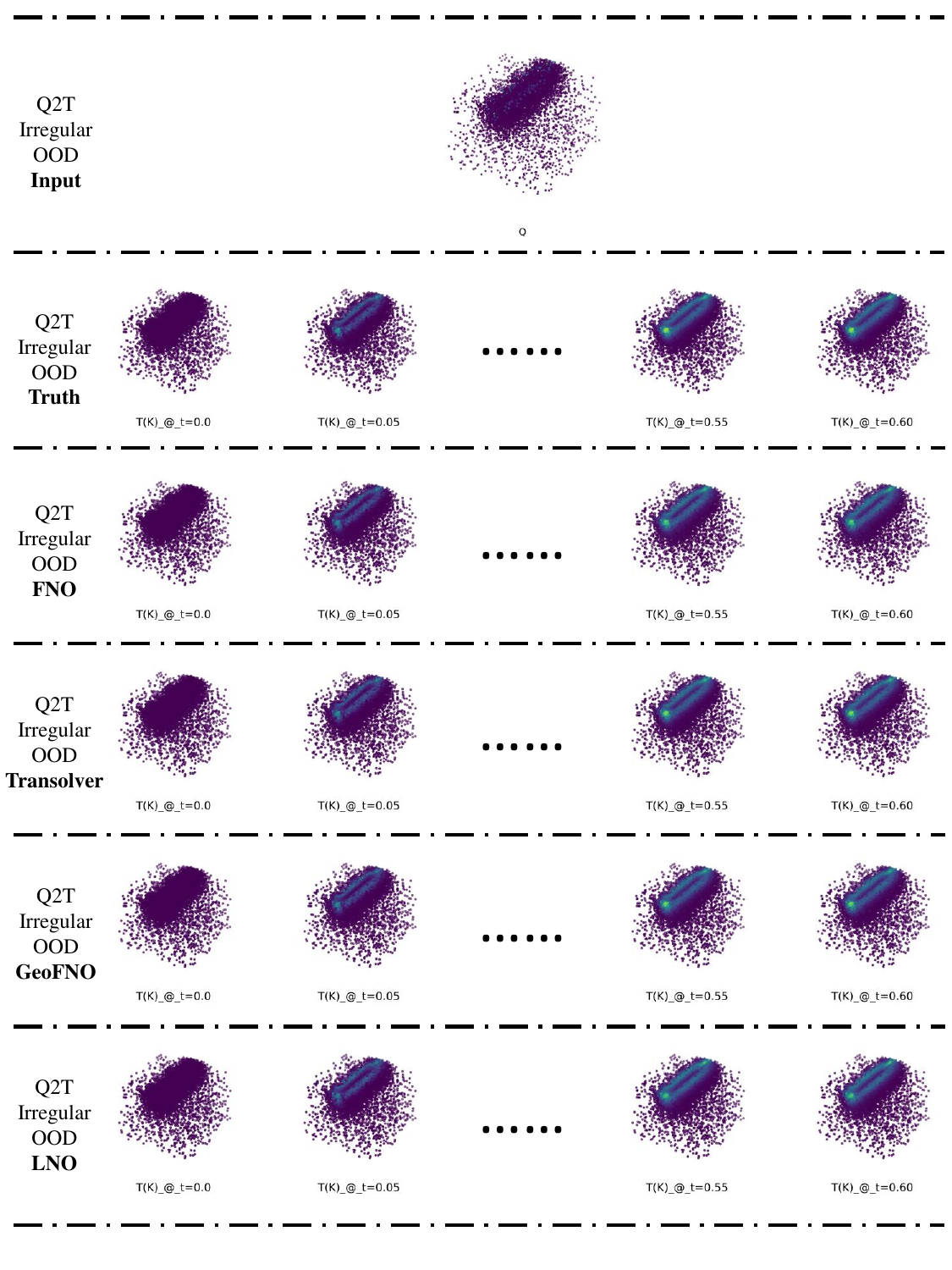}
    \caption{Visualization of experimental results under the heat-to-temperature (Q2T) forward task on irregular grids under OOD setting, using models trained jointly on all six defect types of the Aletheia dataset.}
    \label{fig:vis-Q2T-uns-OOD}
\end{figure}

\begin{figure}[htbp]
    \centering
    \includegraphics[width=1\textwidth]{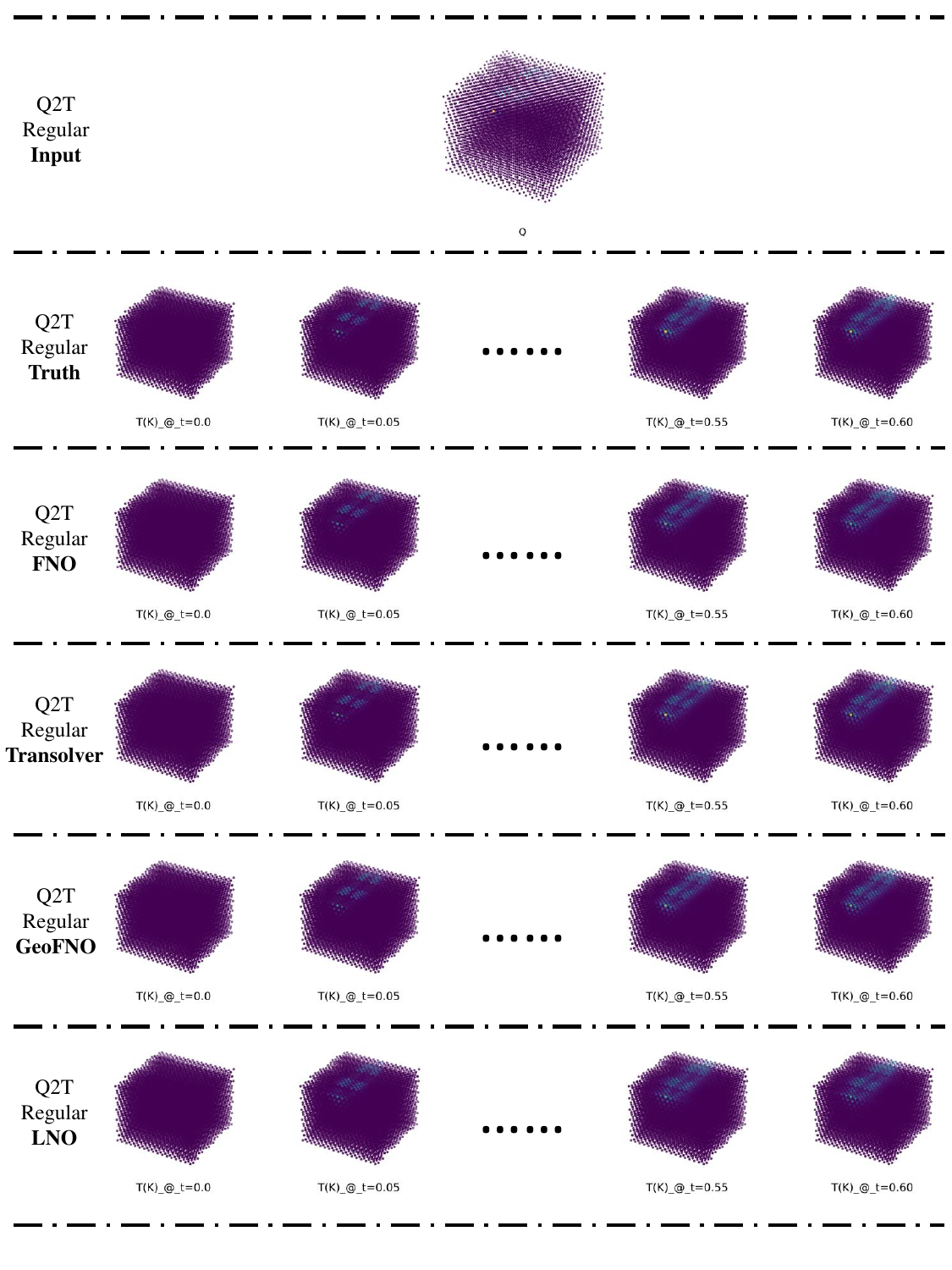}
    \caption{Visualization of experimental results under the heat-to-temperature (Q2T) forward task on regular grids under full-frequency setting, using models trained jointly on all six defect types of the Aletheia dataset.}
    \label{fig:vis-Q2T-s}
\end{figure}

\begin{figure}[htbp]
    \centering
    \includegraphics[width=1\textwidth]{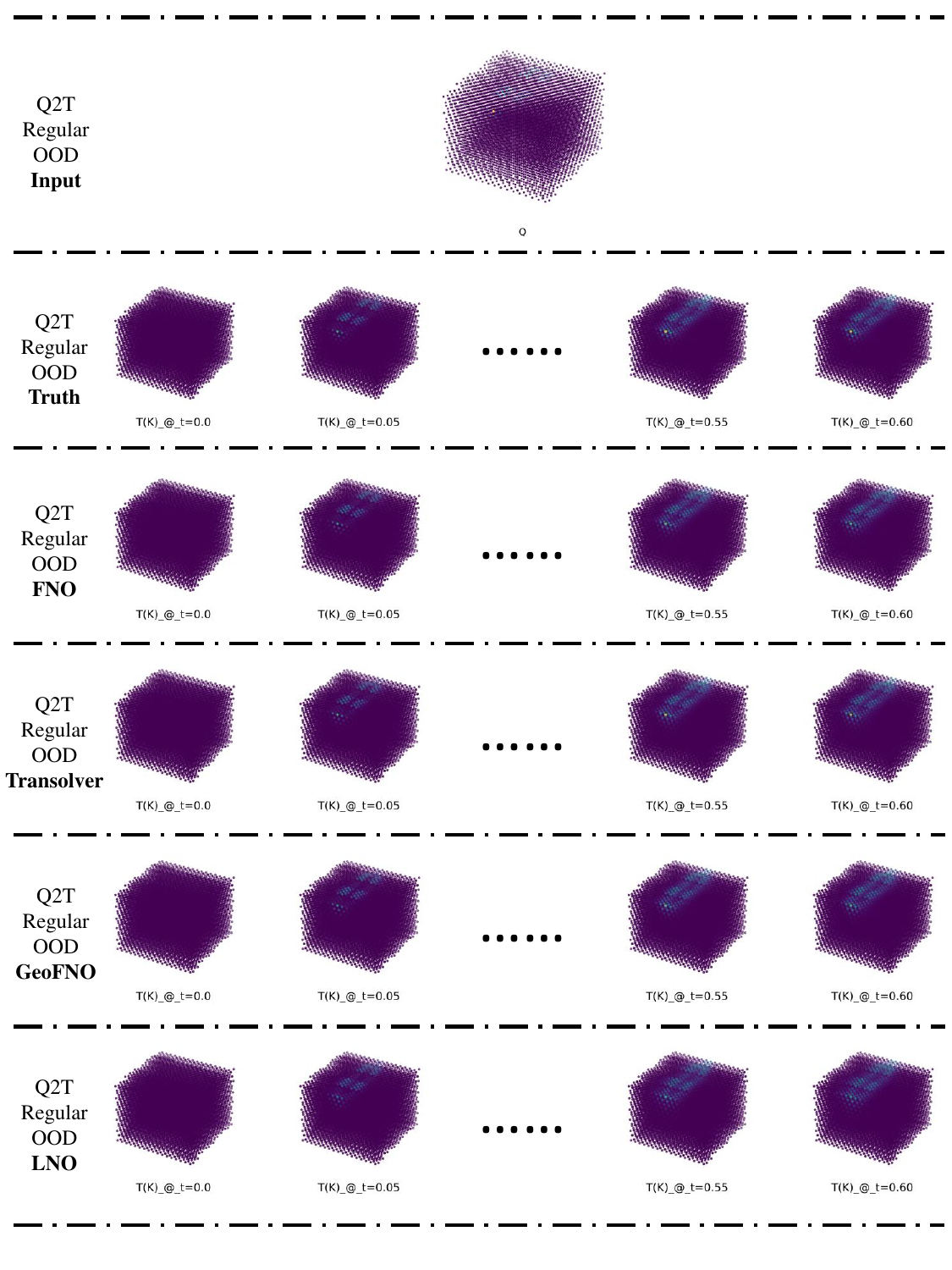}
    \caption{Visualization of experimental results under the heat-to-temperature (Q2T) forward task on regular grids under OOD setting, using models trained jointly on all six defect types of the Aletheia dataset.}
    \label{fig:vis-Q2T-s-OOD}
\end{figure}

\begin{figure}[htbp]
    \centering
    \includegraphics[width=0.9\textwidth]{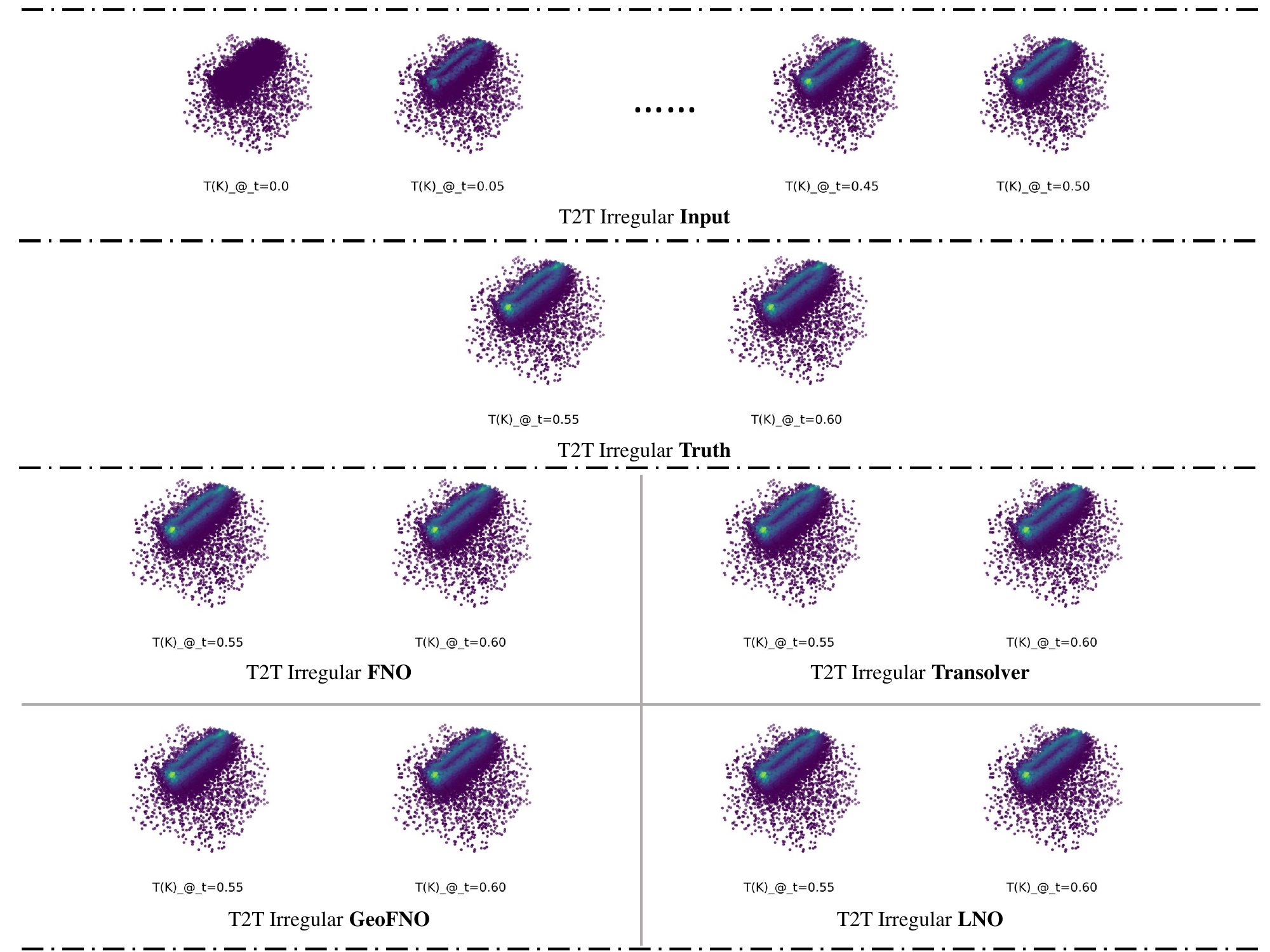}
    \caption{Visualization of experimental results under the temperature-to-temperature (T2T) prediction task on irregular grids under full-frequency setting, using models trained jointly on all six defect types of the Aletheia dataset.}
    \label{fig:vis-T2T-uns}
\end{figure}

\begin{figure}[htbp]
    \centering
    \includegraphics[width=0.9\textwidth]{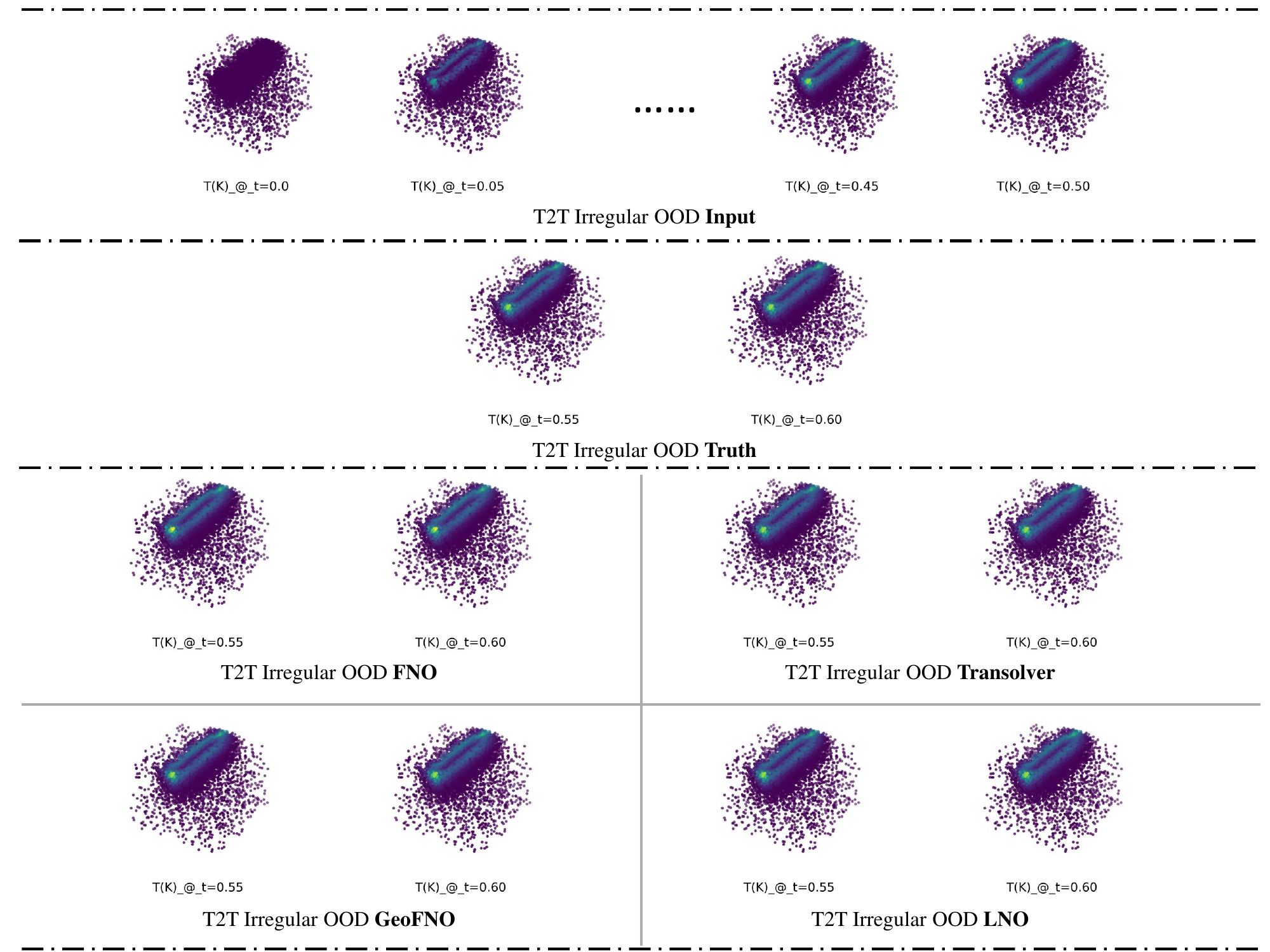}
    \caption{Visualization of experimental results under the temperature-to-temperature (T2T) prediction task on irregular grids under OOD setting, using models trained jointly on all six defect types of the Aletheia dataset.}
    \label{fig:vis-T2T-uns-OOD}
\end{figure}

\begin{figure}[htbp]
    \centering
    \includegraphics[width=0.9\textwidth]{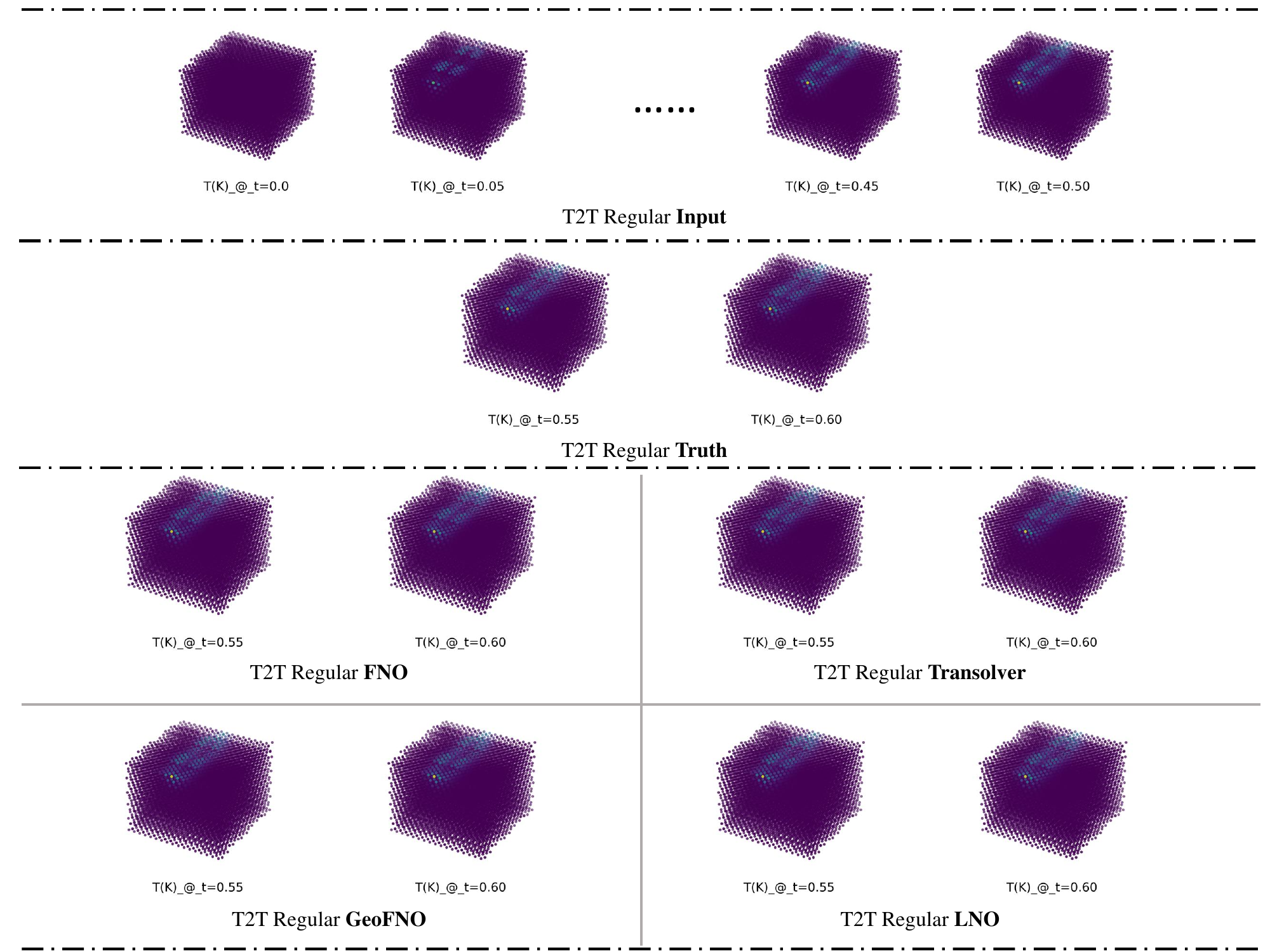}
    \caption{Visualization of experimental results under the temperature-to-temperature (T2T) prediction task on regular grids under full-frequency setting, using models trained jointly on all six defect types of the Aletheia dataset.}
    \label{fig:vis-T2T-s}
\end{figure}

\begin{figure}[htbp]
    \centering
    \includegraphics[width=0.9\textwidth]{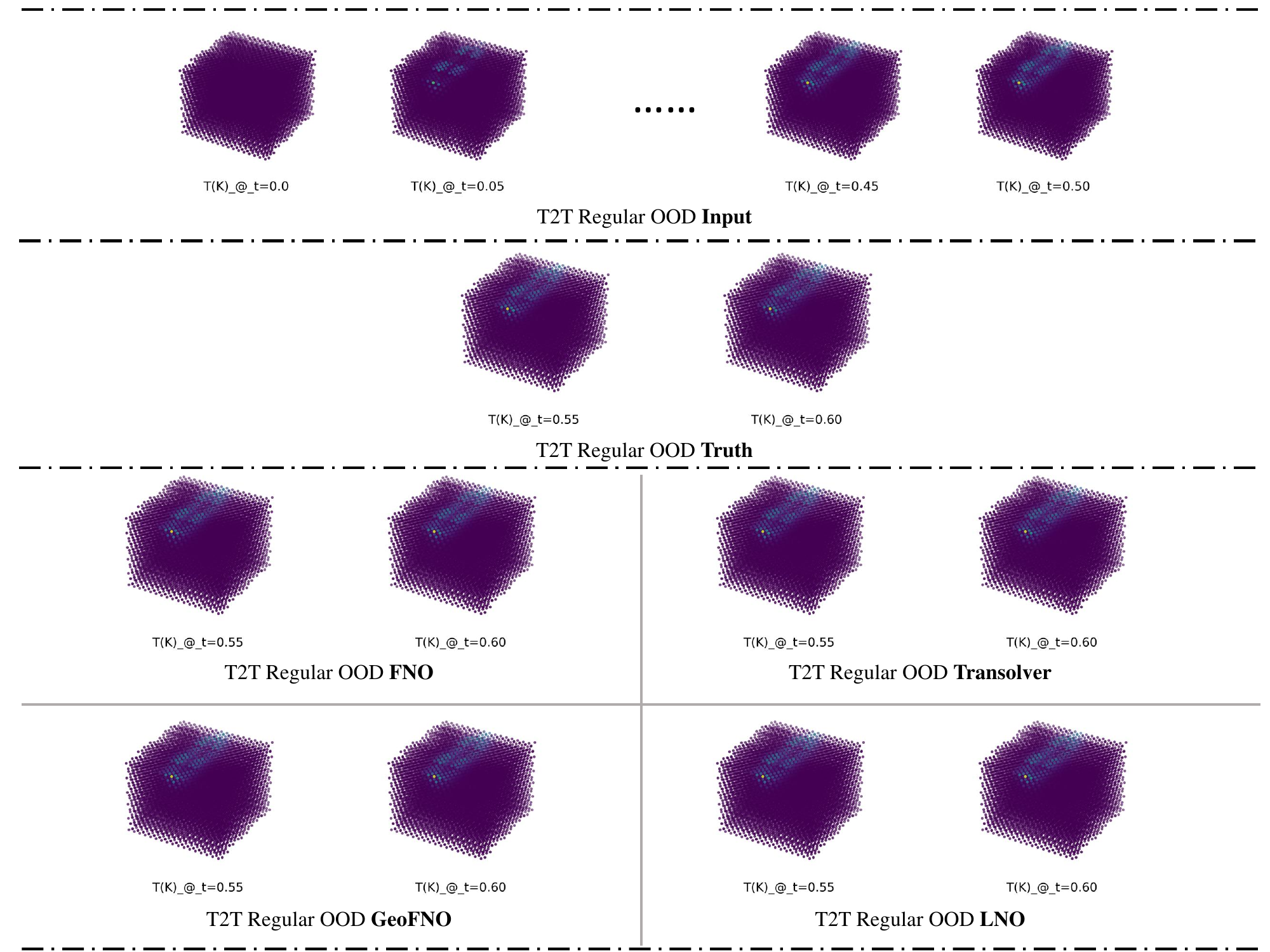}
    \caption{Visualization of experimental results under the temperature-to-temperature (T2T) prediction task on regular grids under OOD setting, using models trained jointly on all six defect types of the Aletheia dataset.}
    \label{fig:vis-T2T-s-OOD}
\end{figure}


\newpage
\section*{NeurIPS Paper Checklist}

\begin{enumerate}

\item {\bf Claims}
    \item[] Question: Do the main claims made in the abstract and introduction accurately reflect the paper's contributions and scope?
    \item[] Answer: \answerYes{} 
    \item[] Justification: The theory of the claims we make matches the experimental results and can exemplify the extent to which these results can be generalized to other settings, see \Cref{section: Introduction} and \Cref{section: conclusion}.
    \item[] Guidelines:
    \begin{itemize}
        \item The answer \answerNA{} means that the abstract and introduction do not include the claims made in the paper.
        \item The abstract and/or introduction should clearly state the claims made, including the contributions made in the paper and important assumptions and limitations. A \answerNo{} or \answerNA{} answer to this question will not be perceived well by the reviewers. 
        \item The claims made should match theoretical and experimental results, and reflect how much the results can be expected to generalize to other settings. 
        \item It is fine to include aspirational goals as motivation as long as it is clear that these goals are not attained by the paper. 
    \end{itemize}

\item {\bf Limitations}
    \item[] Question: Does the paper discuss the limitations of the work performed by the authors?
    \item[] Answer: \answerYes{} 
    \item[] Justification: Limitations are discussed in the conclusion, particularly with regard to defect complexity and reconstruction precision, see \Cref{section: conclusion}
    \item[] Guidelines:
    \begin{itemize}
        \item The answer \answerNA{} means that the paper has no limitation while the answer \answerNo{} means that the paper has limitations, but those are not discussed in the paper. 
        \item The authors are encouraged to create a separate ``Limitations'' section in their paper.
        \item The paper should point out any strong assumptions and how robust the results are to violations of these assumptions (e.g., independence assumptions, noiseless settings, model well-specification, asymptotic approximations only holding locally). The authors should reflect on how these assumptions might be violated in practice and what the implications would be.
        \item The authors should reflect on the scope of the claims made, e.g., if the approach was only tested on a few datasets or with a few runs. In general, empirical results often depend on implicit assumptions, which should be articulated.
        \item The authors should reflect on the factors that influence the performance of the approach. For example, a facial recognition algorithm may perform poorly when image resolution is low or images are taken in low lighting. Or a speech-to-text system might not be used reliably to provide closed captions for online lectures because it fails to handle technical jargon.
        \item The authors should discuss the computational efficiency of the proposed algorithms and how they scale with dataset size.
        \item If applicable, the authors should discuss possible limitations of their approach to address problems of privacy and fairness.
        \item While the authors might fear that complete honesty about limitations might be used by reviewers as grounds for rejection, a worse outcome might be that reviewers discover limitations that aren't acknowledged in the paper. The authors should use their best judgment and recognize that individual actions in favor of transparency play an important role in developing norms that preserve the integrity of the community. Reviewers will be specifically instructed to not penalize honesty concerning limitations.
    \end{itemize}

\item {\bf Theory assumptions and proofs}
    \item[] Question: For each theoretical result, does the paper provide the full set of assumptions and a complete (and correct) proof?
    \item[] Answer: \answerNA{} 
    \item[] Justification: The work is primarily empirical and dataset-driven, focusing on benchmarking neural operator models rather than developing new mathematical theory.
    \item[] Guidelines:
    \begin{itemize}
        \item The answer \answerNA{} means that the paper does not include theoretical results. 
        \item All the theorems, formulas, and proofs in the paper should be numbered and cross-referenced.
        \item All assumptions should be clearly stated or referenced in the statement of any theorems.
        \item The proofs can either appear in the main paper or the supplemental material, but if they appear in the supplemental material, the authors are encouraged to provide a short proof sketch to provide intuition. 
        \item Inversely, any informal proof provided in the core of the paper should be complemented by formal proofs provided in appendix or supplemental material.
        \item Theorems and Lemmas that the proof relies upon should be properly referenced. 
    \end{itemize}

    \item {\bf Experimental result reproducibility}
    \item[] Question: Does the paper fully disclose all the information needed to reproduce the main experimental results of the paper to the extent that it affects the main claims and/or conclusions of the paper (regardless of whether the code and data are provided or not)?
    \item[] Answer: \answerYes{} 
    \item[] Justification: The paper discloses all details needed to reproduce the main experiments. It clearly describes the data splitting strategies and provides the training details applied consistently across models, which should allow one to replicate the benchmark results.
    \item[] Guidelines:
    \begin{itemize}
        \item The answer \answerNA{} means that the paper does not include experiments.
        \item If the paper includes experiments, a \answerNo{} answer to this question will not be perceived well by the reviewers: Making the paper reproducible is important, regardless of whether the code and data are provided or not.
        \item If the contribution is a dataset and\slash or model, the authors should describe the steps taken to make their results reproducible or verifiable. 
        \item Depending on the contribution, reproducibility can be accomplished in various ways. For example, if the contribution is a novel architecture, describing the architecture fully might suffice, or if the contribution is a specific model and empirical evaluation, it may be necessary to either make it possible for others to replicate the model with the same dataset, or provide access to the model. In general. releasing code and data is often one good way to accomplish this, but reproducibility can also be provided via detailed instructions for how to replicate the results, access to a hosted model (e.g., in the case of a large language model), releasing of a model checkpoint, or other means that are appropriate to the research performed.
        \item While NeurIPS does not require releasing code, the conference does require all submissions to provide some reasonable avenue for reproducibility, which may depend on the nature of the contribution. For example
        \begin{enumerate}
            \item If the contribution is primarily a new algorithm, the paper should make it clear how to reproduce that algorithm.
            \item If the contribution is primarily a new model architecture, the paper should describe the architecture clearly and fully.
            \item If the contribution is a new model (e.g., a large language model), then there should either be a way to access this model for reproducing the results or a way to reproduce the model (e.g., with an open-source dataset or instructions for how to construct the dataset).
            \item We recognize that reproducibility may be tricky in some cases, in which case authors are welcome to describe the particular way they provide for reproducibility. In the case of closed-source models, it may be that access to the model is limited in some way (e.g., to registered users), but it should be possible for other researchers to have some path to reproducing or verifying the results.
        \end{enumerate}
    \end{itemize}

\item {\bf Open access to data and code}
    \item[] Question: Does the paper provide open access to the data and code, with sufficient instructions to faithfully reproduce the main experimental results, as described in supplemental material?
    \item[] Answer: \answerYes{} 
    \item[] Justification: We make the data and code publicly available, and we provide links to the Aletheia dataset, as well as the evaluation code, which is also available at \url{https://github.com/sherlock-0327/ALETHEIA}.
    \item[] Guidelines:
    \begin{itemize}
        \item The answer \answerNA{} means that paper does not include experiments requiring code.
        \item Please see the NeurIPS code and data submission guidelines (\url{https://neurips.cc/public/guides/CodeSubmissionPolicy}) for more details.
        \item While we encourage the release of code and data, we understand that this might not be possible, so \answerNo{} is an acceptable answer. Papers cannot be rejected simply for not including code, unless this is central to the contribution (e.g., for a new open-source benchmark).
        \item The instructions should contain the exact command and environment needed to run to reproduce the results. See the NeurIPS code and data submission guidelines (\url{https://neurips.cc/public/guides/CodeSubmissionPolicy}) for more details.
        \item The authors should provide instructions on data access and preparation, including how to access the raw data, preprocessed data, intermediate data, and generated data, etc.
        \item The authors should provide scripts to reproduce all experimental results for the new proposed method and baselines. If only a subset of experiments are reproducible, they should state which ones are omitted from the script and why.
        \item At submission time, to preserve anonymity, the authors should release anonymized versions (if applicable).
        \item Providing as much information as possible in supplemental material (appended to the paper) is recommended, but including URLs to data and code is permitted.
    \end{itemize}

\item {\bf Experimental setting/details}
    \item[] Question: Does the paper specify all the training and test details (e.g., data splits, hyperparameters, how they were chosen, type of optimizer) necessary to understand the results?
    \item[] Answer: \answerYes{} 
    \item[] Justification: We have specified the key training and testing details required. We describe how the training/testing dataset is partitioned, list all the important training parameters. For more details, see \Cref{section: Benchmark}.
    \item[] Guidelines:
    \begin{itemize}
        \item The answer \answerNA{} means that the paper does not include experiments.
        \item The experimental setting should be presented in the core of the paper to a level of detail that is necessary to appreciate the results and make sense of them.
        \item The full details can be provided either with the code, in appendix, or as supplemental material.
    \end{itemize}

\item {\bf Experiment statistical significance}
    \item[] Question: Does the paper report error bars suitably and correctly defined or other appropriate information about the statistical significance of the experiments?
    \item[] Answer: \answerYes{} 
    \item[] Justification: We report multiple metrics across models and settings, including mean and sample standard deviation over three training runs for T2Q and S2Q on both regular and irregular grids in the representative geometric-OOD split (Table~\ref{tab:test6_mean_std}).
    \item[] Guidelines:
    \begin{itemize}
        \item The answer \answerNA{} means that the paper does not include experiments.
        \item The authors should answer \answerYes{} if the results are accompanied by error bars, confidence intervals, or statistical significance tests, at least for the experiments that support the main claims of the paper.
        \item The factors of variability that the error bars are capturing should be clearly stated (for example, train/test split, initialization, random drawing of some parameter, or overall run with given experimental conditions).
        \item The method for calculating the error bars should be explained (closed form formula, call to a library function, bootstrap, etc.)
        \item The assumptions made should be given (e.g., Normally distributed errors).
        \item It should be clear whether the error bar is the standard deviation or the standard error of the mean.
        \item It is OK to report 1-sigma error bars, but one should state it. The authors should preferably report a 2-sigma error bar than state that they have a 96\% CI, if the hypothesis of Normality of errors is not verified.
        \item For asymmetric distributions, the authors should be careful not to show in tables or figures symmetric error bars that would yield results that are out of range (e.g., negative error rates).
        \item If error bars are reported in tables or plots, the authors should explain in the text how they were calculated and reference the corresponding figures or tables in the text.
    \end{itemize}

\item {\bf Experiments compute resources}
    \item[] Question: For each experiment, does the paper provide sufficient information on the computer resources (type of compute workers, memory, time of execution) needed to reproduce the experiments?
    \item[] Answer: \answerYes{} 
    \item[] Justification: Compute details such as training time, and memory usage are explicitly stated in the current version.
    \item[] Guidelines:
    \begin{itemize}
        \item The answer \answerNA{} means that the paper does not include experiments.
        \item The paper should indicate the type of compute workers CPU or GPU, internal cluster, or cloud provider, including relevant memory and storage.
        \item The paper should provide the amount of compute required for each of the individual experimental runs as well as estimate the total compute. 
        \item The paper should disclose whether the full research project required more compute than the experiments reported in the paper (e.g., preliminary or failed experiments that didn't make it into the paper). 
    \end{itemize}
    
\item {\bf Code of ethics}
    \item[] Question: Does the research conducted in the paper conform, in every respect, with the NeurIPS Code of Ethics \url{https://neurips.cc/public/EthicsGuidelines}?
    \item[] Answer: \answerYes{} 
    \item[] Justification: The research complies with ethical guidelines and poses no ethical concerns. All data are synthetic or anonymized.
    \item[] Guidelines:
    \begin{itemize}
        \item The answer \answerNA{} means that the authors have not reviewed the NeurIPS Code of Ethics.
        \item If the authors answer \answerNo, they should explain the special circumstances that require a deviation from the Code of Ethics.
        \item The authors should make sure to preserve anonymity (e.g., if there is a special consideration due to laws or regulations in their jurisdiction).
    \end{itemize}

\item {\bf Broader impacts}
    \item[] Question: Does the paper discuss both potential positive societal impacts and negative societal impacts of the work performed?
    \item[] Answer: \answerYes{} 
    \item[] Justification: Positive societal impacts are discussed in terms of improving industrial inspection and safety; potential misuse or bias is limited due to the nature of synthetic data, see \Cref{section: conclusion} and \Cref{appendix: Broader Impacts}.
    \item[] Guidelines:
    \begin{itemize}
        \item The answer \answerNA{} means that there is no societal impact of the work performed.
        \item If the authors answer \answerNA{} or \answerNo, they should explain why their work has no societal impact or why the paper does not address societal impact.
        \item Examples of negative societal impacts include potential malicious or unintended uses (e.g., disinformation, generating fake profiles, surveillance), fairness considerations (e.g., deployment of technologies that could make decisions that unfairly impact specific groups), privacy considerations, and security considerations.
        \item The conference expects that many papers will be foundational research and not tied to particular applications, let alone deployments. However, if there is a direct path to any negative applications, the authors should point it out. For example, it is legitimate to point out that an improvement in the quality of generative models could be used to generate Deepfakes for disinformation. On the other hand, it is not needed to point out that a generic algorithm for optimizing neural networks could enable people to train models that generate Deepfakes faster.
        \item The authors should consider possible harms that could arise when the technology is being used as intended and functioning correctly, harms that could arise when the technology is being used as intended but gives incorrect results, and harms following from (intentional or unintentional) misuse of the technology.
        \item If there are negative societal impacts, the authors could also discuss possible mitigation strategies (e.g., gated release of models, providing defenses in addition to attacks, mechanisms for monitoring misuse, mechanisms to monitor how a system learns from feedback over time, improving the efficiency and accessibility of ML).
    \end{itemize}
    
\item {\bf Safeguards}
    \item[] Question: Does the paper describe safeguards that have been put in place for responsible release of data or models that have a high risk for misuse (e.g., pre-trained language models, image generators, or scraped datasets)?
    \item[] Answer: \answerNA{} 
    \item[] Justification: The dataset and models do not pose significant risks of misuse and are not sensitive in nature.
    \item[] Guidelines:
    \begin{itemize}
        \item The answer \answerNA{} means that the paper poses no such risks.
        \item Released models that have a high risk for misuse or dual-use should be released with necessary safeguards to allow for controlled use of the model, for example by requiring that users adhere to usage guidelines or restrictions to access the model or implementing safety filters. 
        \item Datasets that have been scraped from the Internet could pose safety risks. The authors should describe how they avoided releasing unsafe images.
        \item We recognize that providing effective safeguards is challenging, and many papers do not require this, but we encourage authors to take this into account and make a best faith effort.
    \end{itemize}

\item {\bf Licenses for existing assets}
    \item[] Question: Are the creators or original owners of assets (e.g., code, data, models), used in the paper, properly credited and are the license and terms of use explicitly mentioned and properly respected?
    \item[] Answer: \answerYes{} 
    \item[] Justification:  External models and datasets are properly cited in the references (e.g., FNO ~\cite{FNO}, GeoFNO ~\cite{Geo-FNO}).
    \item[] Guidelines:
    \begin{itemize}
        \item The answer \answerNA{} means that the paper does not use existing assets.
        \item The authors should cite the original paper that produced the code package or dataset.
        \item The authors should state which version of the asset is used and, if possible, include a URL.
        \item The name of the license (e.g., CC-BY 4.0) should be included for each asset.
        \item For scraped data from a particular source (e.g., website), the copyright and terms of service of that source should be provided.
        \item If assets are released, the license, copyright information, and terms of use in the package should be provided. For popular datasets, \url{paperswithcode.com/datasets} has curated licenses for some datasets. Their licensing guide can help determine the license of a dataset.
        \item For existing datasets that are re-packaged, both the original license and the license of the derived asset (if it has changed) should be provided.
        \item If this information is not available online, the authors are encouraged to reach out to the asset's creators.
    \end{itemize}

\item {\bf New assets}
    \item[] Question: Are new assets introduced in the paper well documented and is the documentation provided alongside the assets?
    \item[] Answer: \answerYes{} 
    \item[] Justification: The Aletheia dataset is documented, includes structured metadata, and is publicly available.
    \item[] Guidelines:
    \begin{itemize}
        \item The answer \answerNA{} means that the paper does not release new assets.
        \item Researchers should communicate the details of the dataset\slash code\slash model as part of their submissions via structured templates. This includes details about training, license, limitations, etc. 
        \item The paper should discuss whether and how consent was obtained from people whose asset is used.
        \item At submission time, remember to anonymize your assets (if applicable). You can either create an anonymized URL or include an anonymized zip file.
    \end{itemize}

\item {\bf Crowdsourcing and research with human subjects}
    \item[] Question: For crowdsourcing experiments and research with human subjects, does the paper include the full text of instructions given to participants and screenshots, if applicable, as well as details about compensation (if any)? 
    \item[] Answer: \answerNA{} 
    \item[] Justification: No human subjects or crowdsourcing are involved in this research.
    \item[] Guidelines:
    \begin{itemize}
        \item The answer \answerNA{} means that the paper does not involve crowdsourcing nor research with human subjects.
        \item Including this information in the supplemental material is fine, but if the main contribution of the paper involves human subjects, then as much detail as possible should be included in the main paper. 
        \item According to the NeurIPS Code of Ethics, workers involved in data collection, curation, or other labor should be paid at least the minimum wage in the country of the data collector. 
    \end{itemize}

\item {\bf Institutional review board (IRB) approvals or equivalent for research with human subjects}
    \item[] Question: Does the paper describe potential risks incurred by study participants, whether such risks were disclosed to the subjects, and whether Institutional Review Board (IRB) approvals (or an equivalent approval/review based on the requirements of your country or institution) were obtained?
    \item[] Answer: \answerNA{} 
    \item[] Justification: The research does not involve human subjects or crowdsourcing. All data are either synthetically generated through simulations or acquired from controlled laboratory experiments on inanimate rail specimens. There are no ethical, privacy, or consent concerns associated with the data collection process.
    \item[] Guidelines:
    \begin{itemize}
        \item The answer \answerNA{} means that the paper does not involve crowdsourcing nor research with human subjects.
        \item Depending on the country in which research is conducted, IRB approval (or equivalent) may be required for any human subjects research. If you obtained IRB approval, you should clearly state this in the paper. 
        \item We recognize that the procedures for this may vary significantly between institutions and locations, and we expect authors to adhere to the NeurIPS Code of Ethics and the guidelines for their institution. 
        \item For initial submissions, do not include any information that would break anonymity (if applicable), such as the institution conducting the review.
    \end{itemize}

\item {\bf Declaration of LLM usage}
    \item[] Question: Does the paper describe the usage of LLMs if it is an important, original, or non-standard component of the core methods in this research? Note that if the LLM is used only for writing, editing, or formatting purposes and does \emph{not} impact the core methodology, scientific rigor, or originality of the research, declaration is not required.
    \item[] Answer: \answerNA{} 
    \item[] Justification: LLMs are not used as a component of the core methodology.
    \item[] Guidelines:
    \begin{itemize}
        \item The answer \answerNA{} means that the core method development in this research does not involve LLMs as any important, original, or non-standard components.
        \item Please refer to our LLM policy in the NeurIPS handbook for what should or should not be described.
    \end{itemize}

\end{enumerate}

\end{document}